%% file: main.tex
\documentclass[10pt,twocolumn,letterpaper]{article}

\usepackage[pagenumbers]{cvpr} % To force page numbers, e.g. for an arXiv version

\input{preamble}

\definecolor{cvprblue}{rgb}{0.21,0.49,0.74}
\usepackage[pagebackref,breaklinks,colorlinks,allcolors=cvprblue]{hyperref}

\def\paperID{10379} % *** Enter the Paper ID here
\def\confName{CVPR}
\def\confYear{2025}

\title{Forensic Self-Descriptions Are All You Need for Zero-Shot Detection, Open-Set Source Attribution, and Clustering of AI-generated Images}

\author{
	Tai D. Nguyen, Aref Azizpour, Matthew C. Stamm\\
	Drexel University\\
	Philadelphia, PA, USA\\
	{\tt\small {tdn47,aa4639,mcs382}@drexel.edu}
}

\begin{document}
\makeatletter
\let\@oldmaketitle\@maketitle % Store the original \@maketitle
\renewcommand{\@maketitle}{%
	\@oldmaketitle% Insert the original title block
	\begin{center} % Center the figure and caption
		\vspace{-1.5em}
		\includegraphics[width=0.95\linewidth,height=7.2\baselineskip]{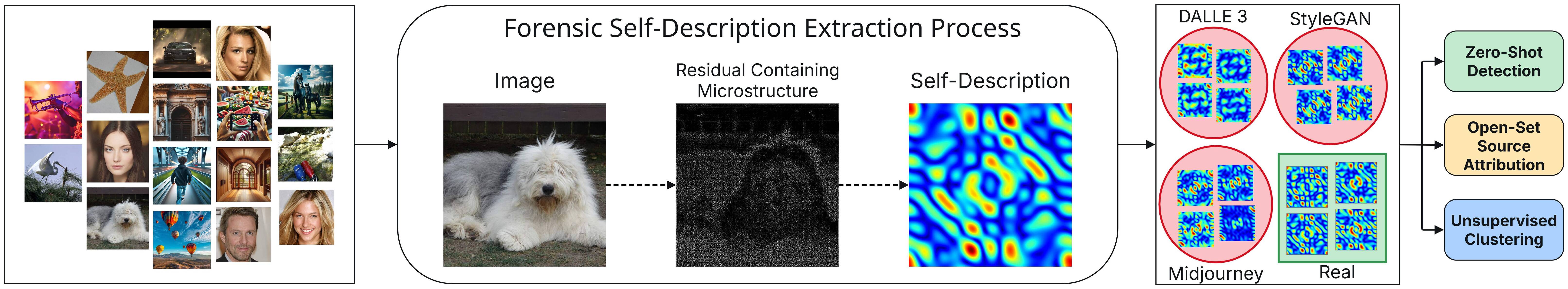}
%		\vspace{-3.5em} % Adjust space between image and caption if needed
		\small % Set caption font size
		Figure 1. We extract the self-description of the forensic microstructures in each image and use them to accurately perform a variety of challenging tasks, including: zero-shot detection, open-set source attribution, and unsupervised clustering of image sources.
	\end{center}%
	\vspace{0.1em} % Adjust space after the figure and caption
}
\makeatother

\maketitle

\setcounter{figure}{1}

\input{sections/Abstract_v2}

\input{figures/DataViz}

\input{sections/Intro_v1}

\input{figures/SystemDiagram}

\pullupp
\input{sections/Background_v2}

\input{figures/SelfDescViz}

\pullup
\input{sections/ProposedMethod_v0}

\input{sections/Applications_v0}

\input{sections/Results_v0}

\pullupp
\input{sections/Ablation_v0}

\pullup
\input{sections/Discussion_v0}

\pullup
\input{sections/Conclusion_v0}

\pullup
\section{Acknowledgement}
This material is based upon work supported by the National Science Foundation under Grant No. 2320600. Any opinions, findings, and conclusions or recommendations expressed in this material are those of the authors and do not necessarily reflect the views of the National Science Foundation.

{
    \small
    \bibliographystyle{ieeenat_fullname}
    \bibliography{main}
}

\input{sections/X_suppl}

\end{document}

%% file: preamble.tex
\usepackage[dvipsnames]{xcolor}
\usepackage{amsmath}
\usepackage{verbatim}
\usepackage{tabularray}
\usepackage{array}
\usepackage{rotating}
\usepackage{multirow}
\usepackage{ragged2e}
\usepackage{scalerel}
\usepackage{bbm}
\usepackage{graphicx}
\usepackage{wrapfig}
\usepackage{booktabs}
\usepackage{xspace}
\usepackage{fontawesome}

\newcommand{\subheader}[1]{\vspace{0.25\baselineskip} \noindent \textbf{#1.}}

\newcommand{\smaller}{\fontsize{8pt}{9pt}\selectfont}
\newcommand{\smallerr}{\fontsize{6.5pt}{7.6pt}\selectfont}
\newcommand{\pullup}{\vspace{-0.2\baselineskip}}
\newcommand{\pullupp}{\vspace{-0.4\baselineskip}}
\newcommand{\pulluppp}{\vspace{-0.8\baselineskip}}

\newcommand{\skipless}{
	\setlength{\abovedisplayskip}{3.75pt}
	\setlength{\belowdisplayskip}{3.75pt}
	\setlength{\abovedisplayshortskip}{3.75pt}
	\setlength{\belowdisplayshortskip}{3.75pt}
}

%% file: sections/Abstract_v2.tex
\begin{abstract}

	The emergence of advanced AI-based tools to generate realistic images poses significant challenges for forensic detection and source attribution, especially as new generative techniques appear rapidly. Traditional methods often fail to generalize to unseen generators due to reliance on features specific to known sources during training. To address this problem, we propose a novel approach that explicitly models forensic microstructures—subtle, pixel-level patterns unique to the image creation process.
	Using only real images in a self-supervised manner, we learn a set of diverse predictive filters to extract residuals that capture different aspects of these microstructures. By jointly modeling these residuals across multiple scales, we obtain a compact model whose parameters constitute a unique forensic self-description for each image. This self-description enables us to perform zero-shot detection of synthetic images, open-set source attribution of images, and clustering based on source without prior knowledge. Extensive experiments demonstrate that our method achieves superior accuracy and adaptability compared to competing techniques, advancing the state of the art in synthetic media forensics.

\end{abstract}
\vspace{-1.6em}

%% file: figures/DataViz.tex
\begin{figure*}[!t]
    \centering
    \setlength{\fboxsep}{0pt}
    \setlength{\fboxrule}{0.01pt}
    \begin{tabular}{
            *{10}{@{\hskip 2pt} p{0.090\textwidth} @{\hskip 2pt}}
        }
        \makebox[0.089\textwidth]{}
        & \makebox[0.089\textwidth]{}
        & \makebox[0.089\textwidth]{}
        & \makebox[0.089\textwidth]{\raisebox{1pt}{\smallerr COCO17}}
        & \makebox[0.089\textwidth]{\raisebox{1pt}{\smallerr ImageNet-1k}}
        & \makebox[0.089\textwidth]{\raisebox{1pt}{\smallerr ImageNet-22k}}
        & \makebox[0.089\textwidth]{\raisebox{1pt}{\smallerr MIDB}}
        & \makebox[0.089\textwidth]{}
        & \makebox[0.089\textwidth]{}
        & \makebox[0.089\textwidth]{} \\

        \makebox[0.089\textwidth]{}
        & \makebox[0.089\textwidth]{}
        & \makebox[0.089\textwidth]{}
        & \fbox{\includegraphics[width=0.089\textwidth]{{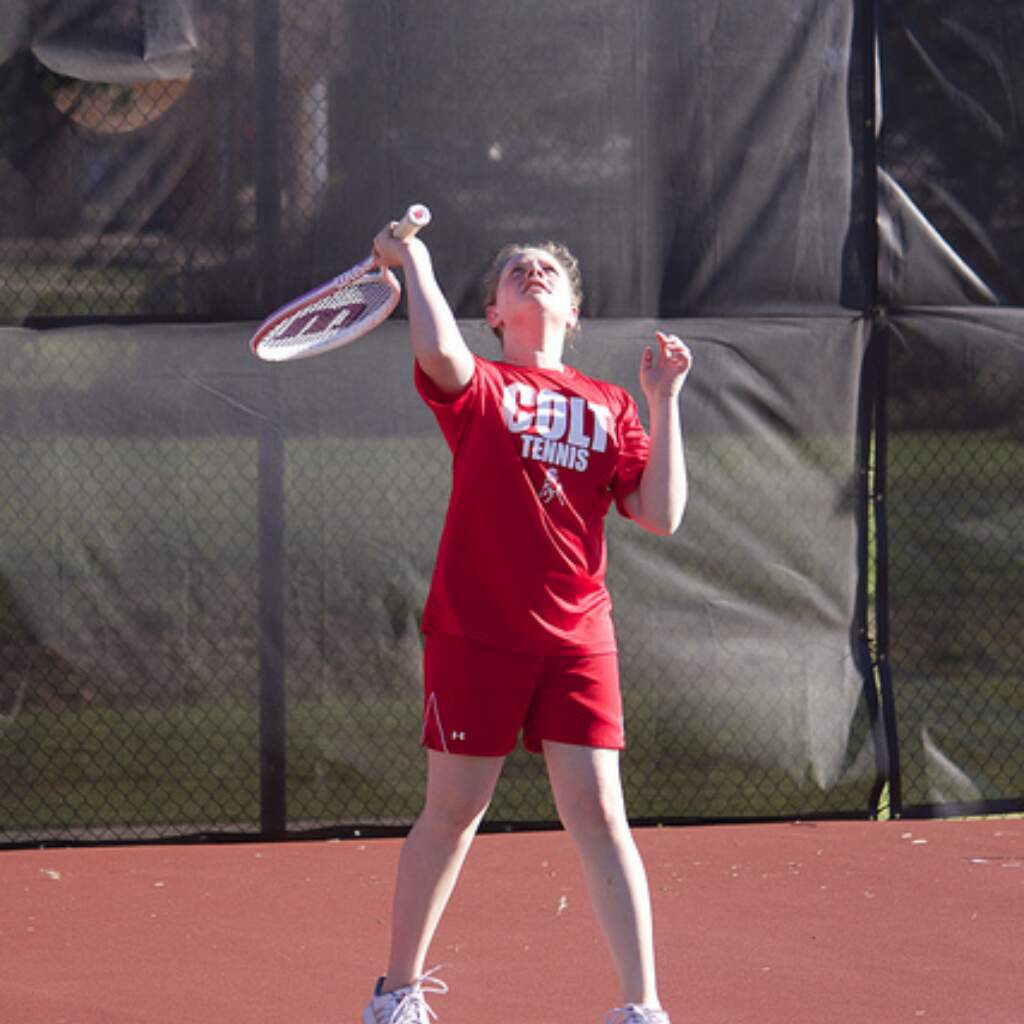}}}
        & \fbox{\includegraphics[width=0.089\textwidth]{{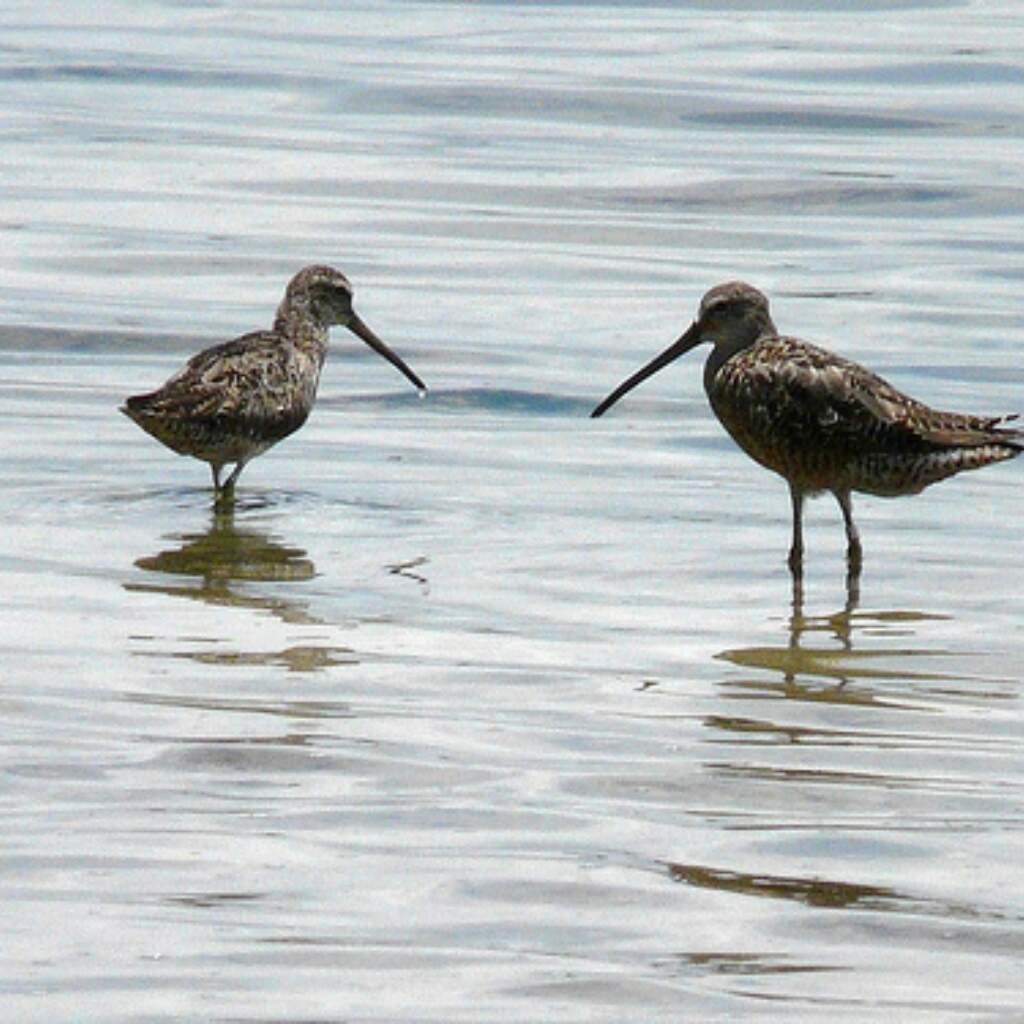}}}
        & \fbox{\includegraphics[width=0.089\textwidth]{{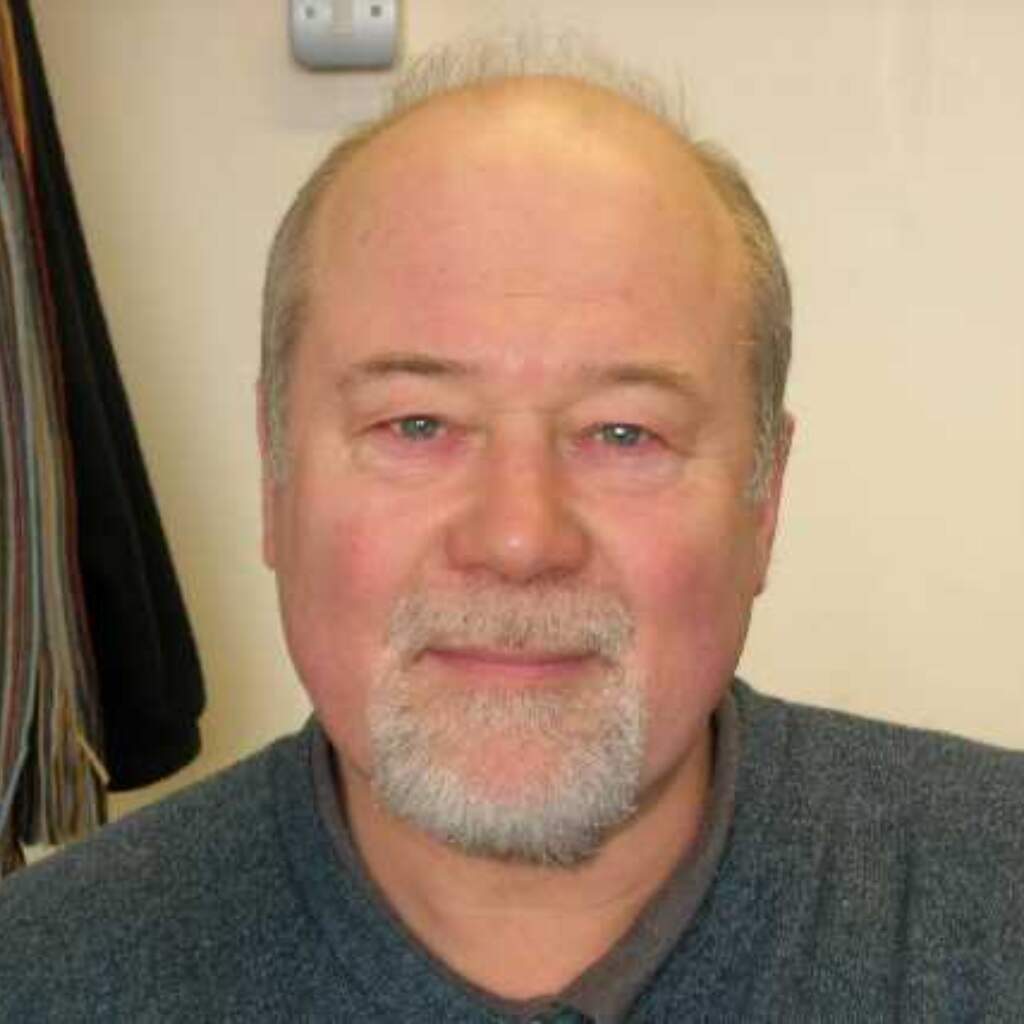}}}
        & \fbox{\includegraphics[width=0.089\textwidth]{{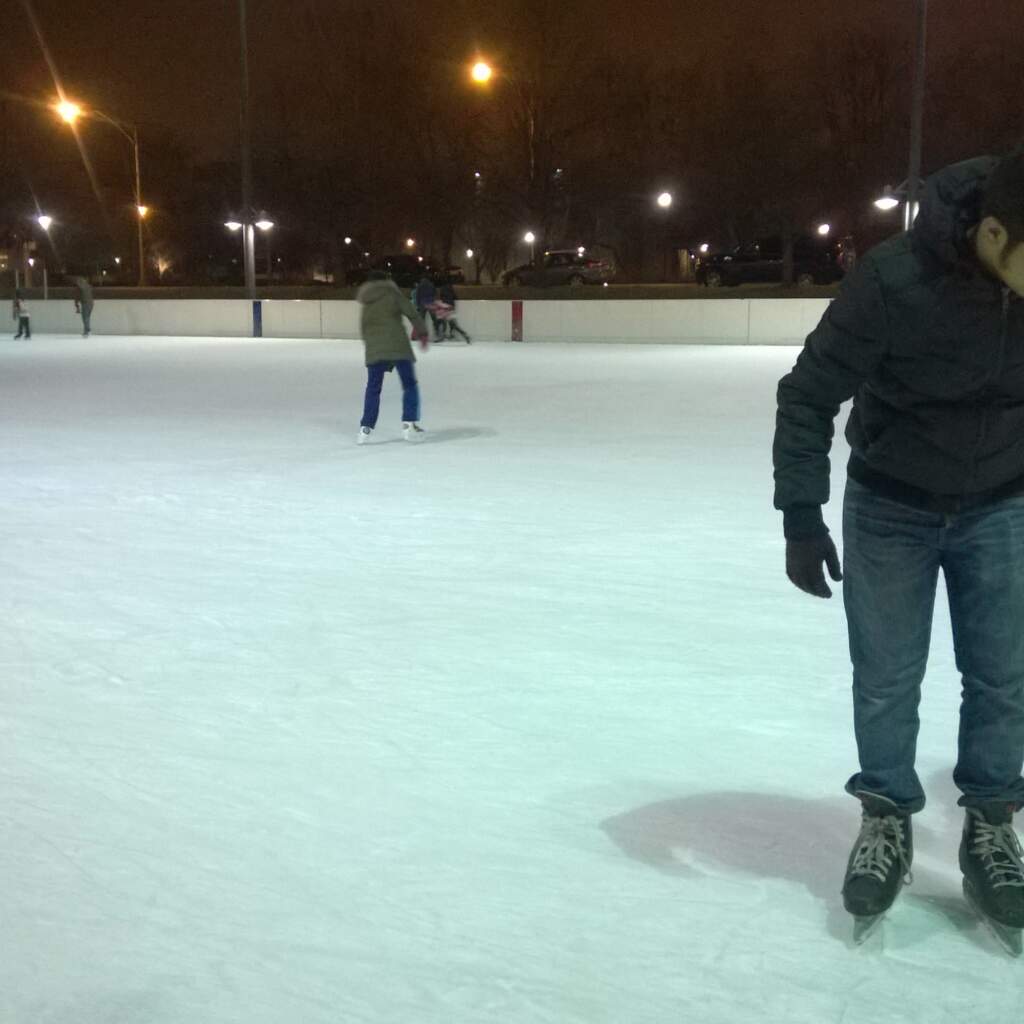}}}
        & \makebox[0.089\textwidth]{}
        & \makebox[0.089\textwidth]{}
        & \makebox[0.089\textwidth]{} \\

        \makebox[0.089\textwidth]{\raisebox{1pt}{\smallerr ProGAN}}
        & \makebox[0.089\textwidth]{\raisebox{1pt}{\smallerr Proj.GAN}}
        & \makebox[0.089\textwidth]{\raisebox{1pt}{\smallerr StyleGAN3}}
        & \makebox[0.089\textwidth]{\raisebox{1pt}{\smallerr GigaGAN}}
        & \makebox[0.089\textwidth]{\raisebox{1pt}{\smallerr SD 1.5}}
        & \makebox[0.089\textwidth]{\raisebox{1pt}{\smallerr SDXL}}
        & \makebox[0.089\textwidth]{\raisebox{1pt}{\smallerr SD 3}}
        & \makebox[0.089\textwidth]{\raisebox{1pt}{\smallerr DALLE 3}}
        & \makebox[0.089\textwidth]{\raisebox{1pt}{\smallerr MJ v6}}
        & \makebox[0.089\textwidth]{\raisebox{1pt}{\smallerr Firefly}} \\

        \fbox{\includegraphics[width=0.089\textwidth]{{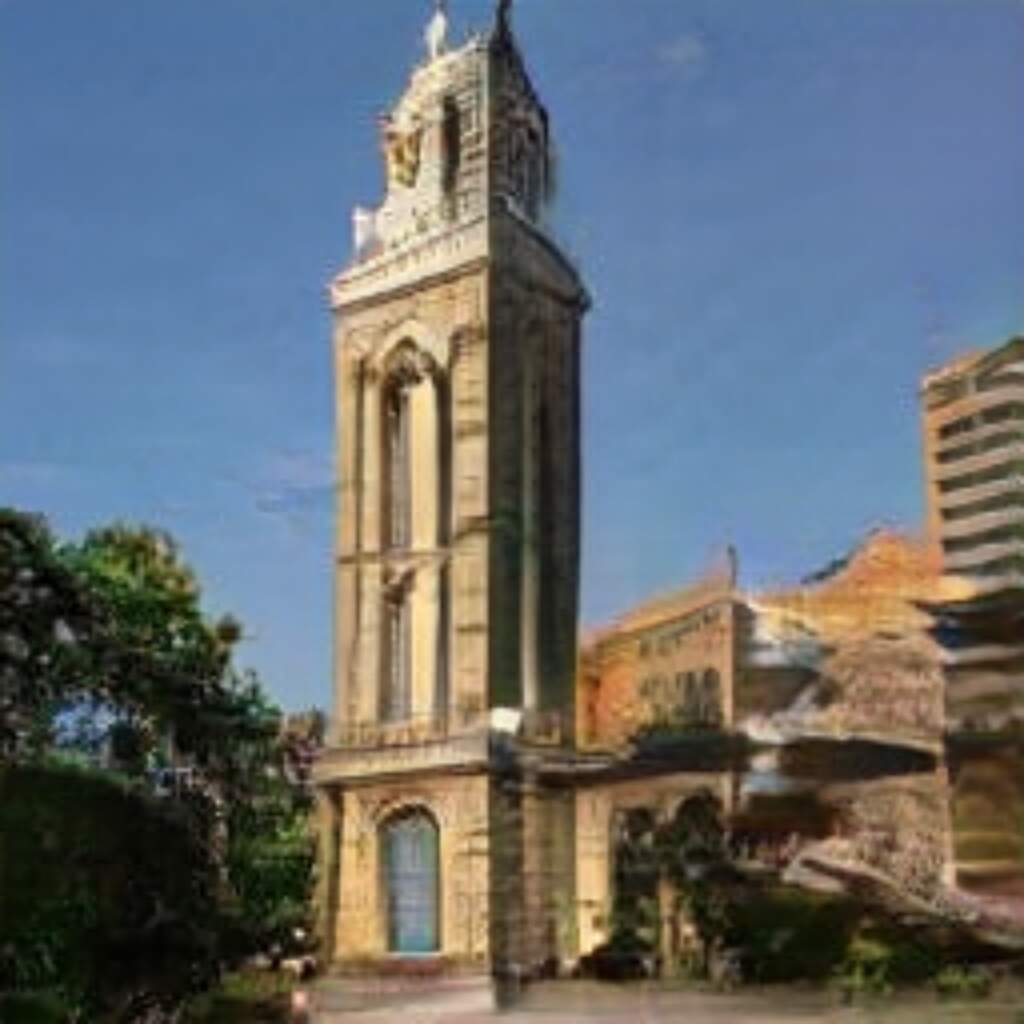}}}
        & \fbox{\includegraphics[width=0.089\textwidth]{{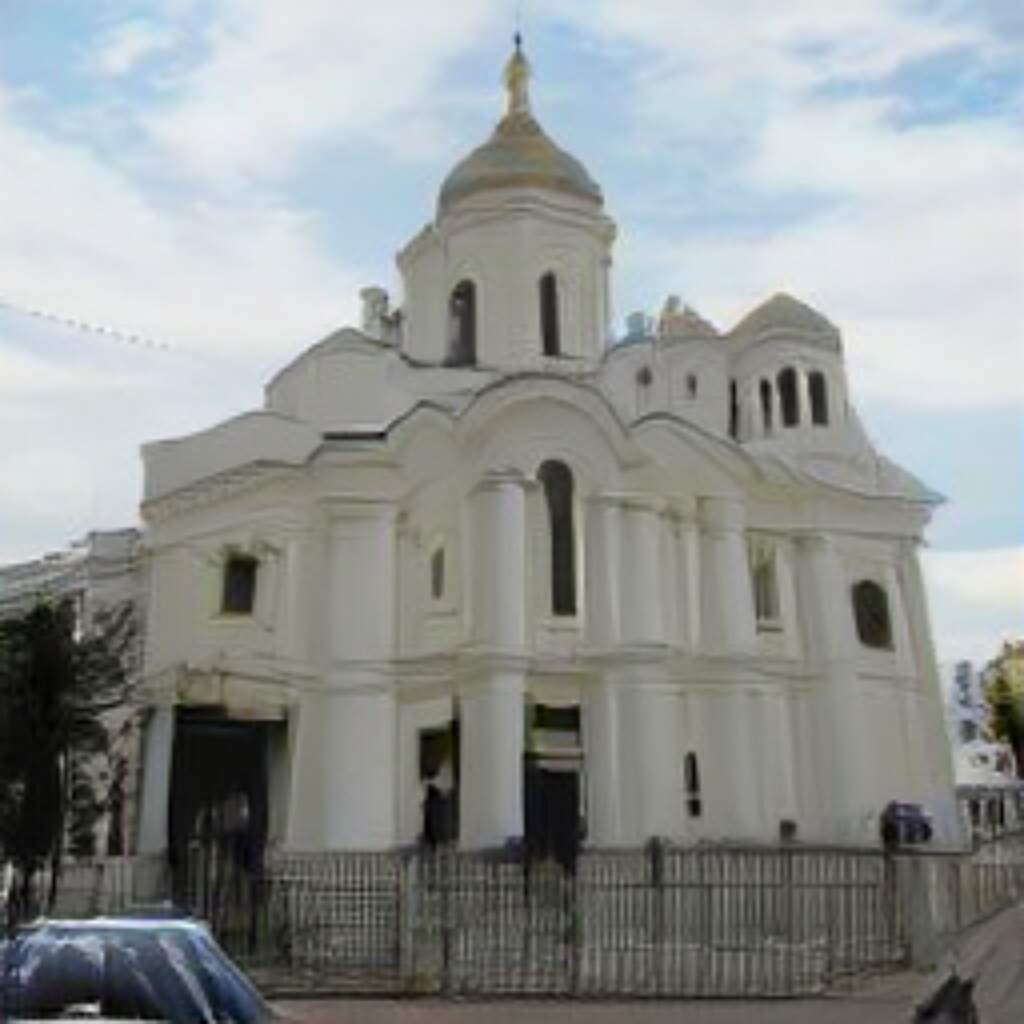}}}
        & \fbox{\includegraphics[width=0.089\textwidth]{{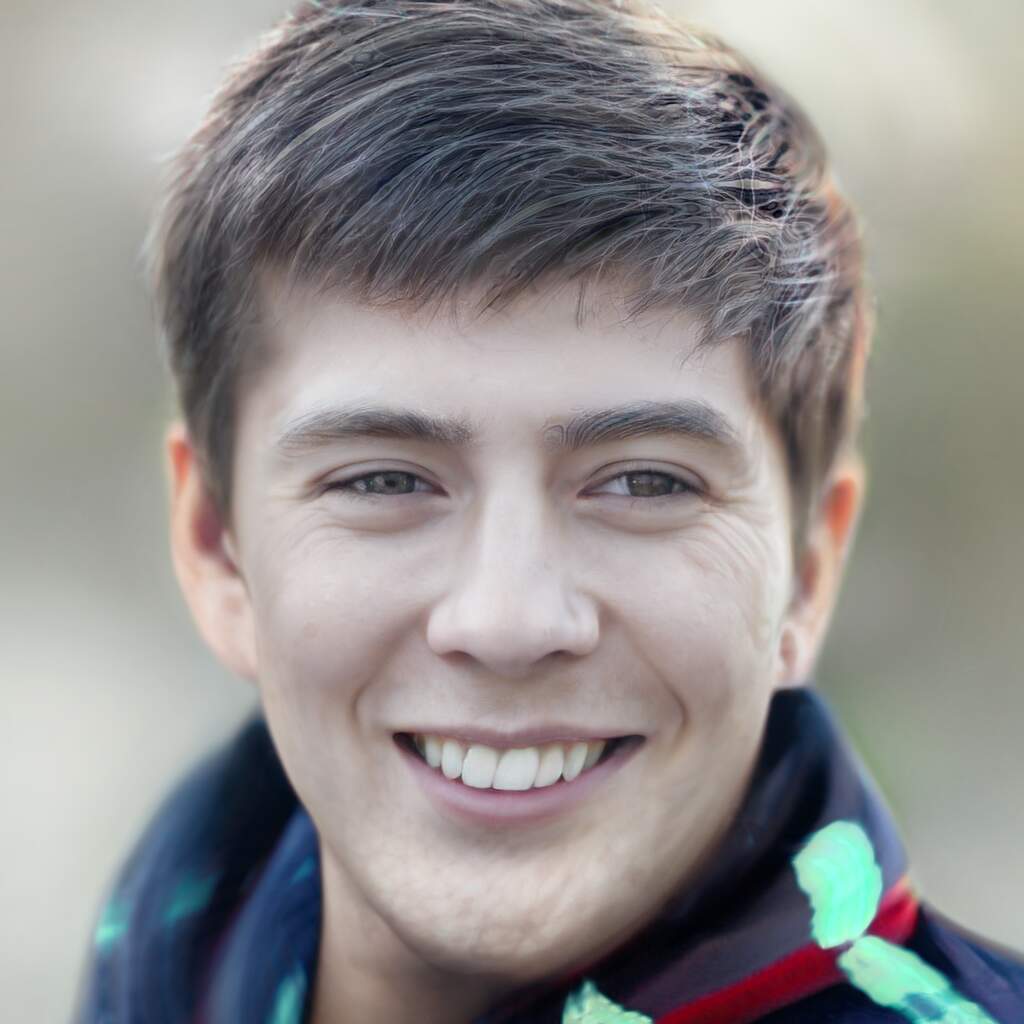}}}
        & \fbox{\includegraphics[width=0.089\textwidth]{{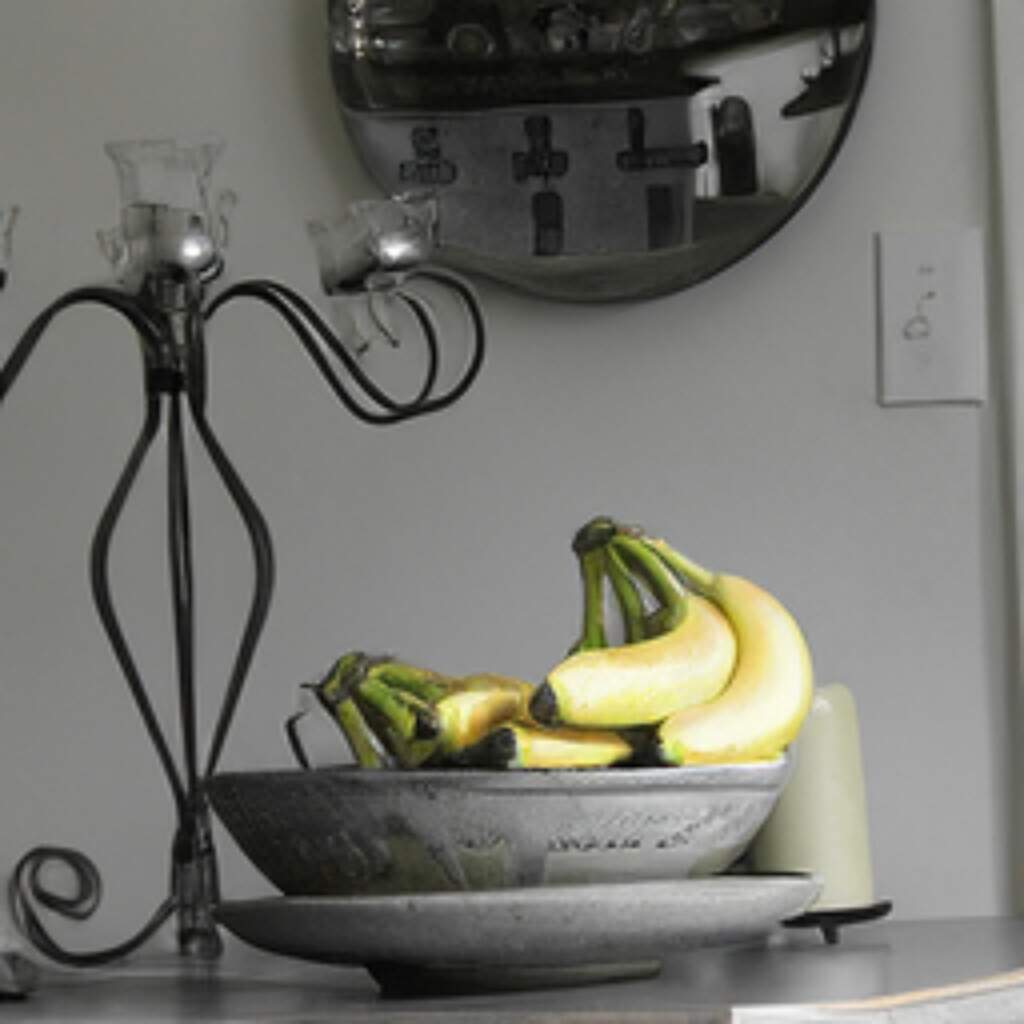}}}
        & \fbox{\includegraphics[width=0.089\textwidth]{{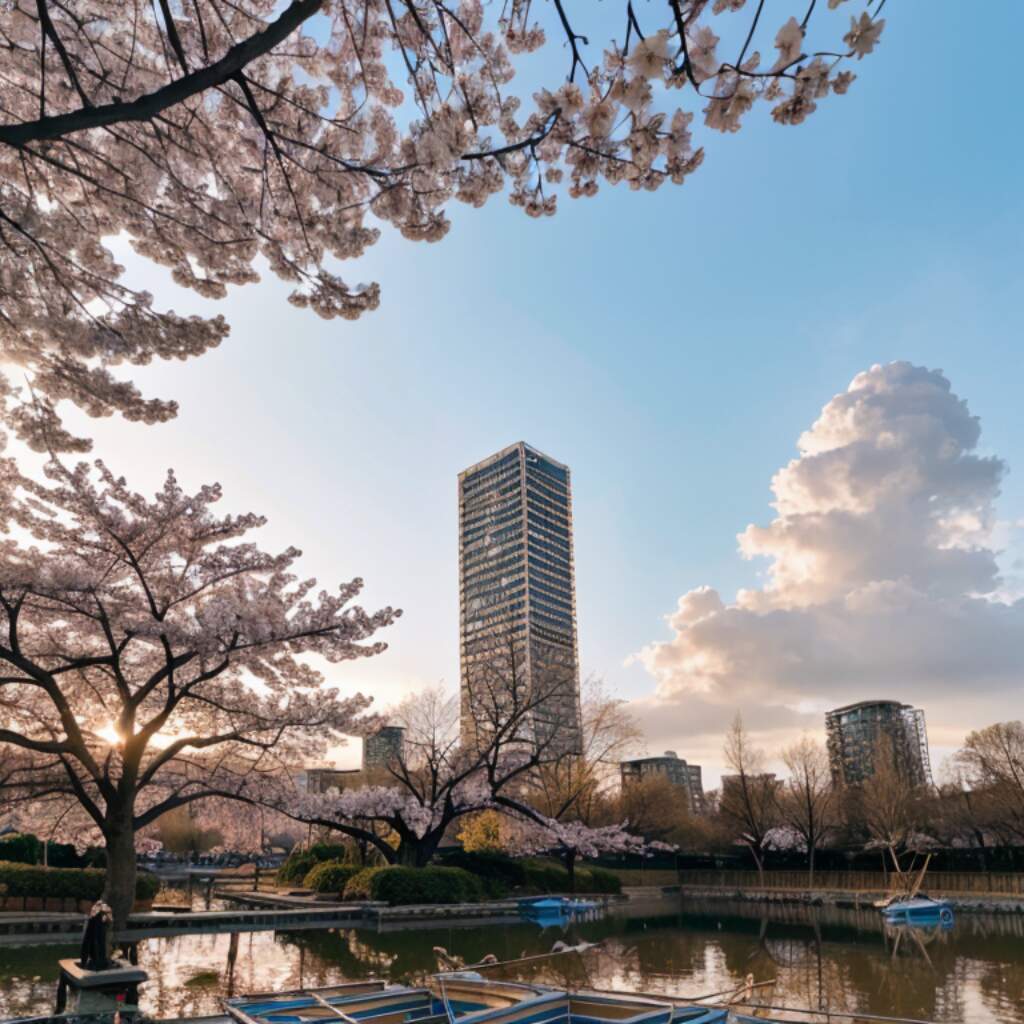}}}
        & \fbox{\includegraphics[width=0.089\textwidth]{{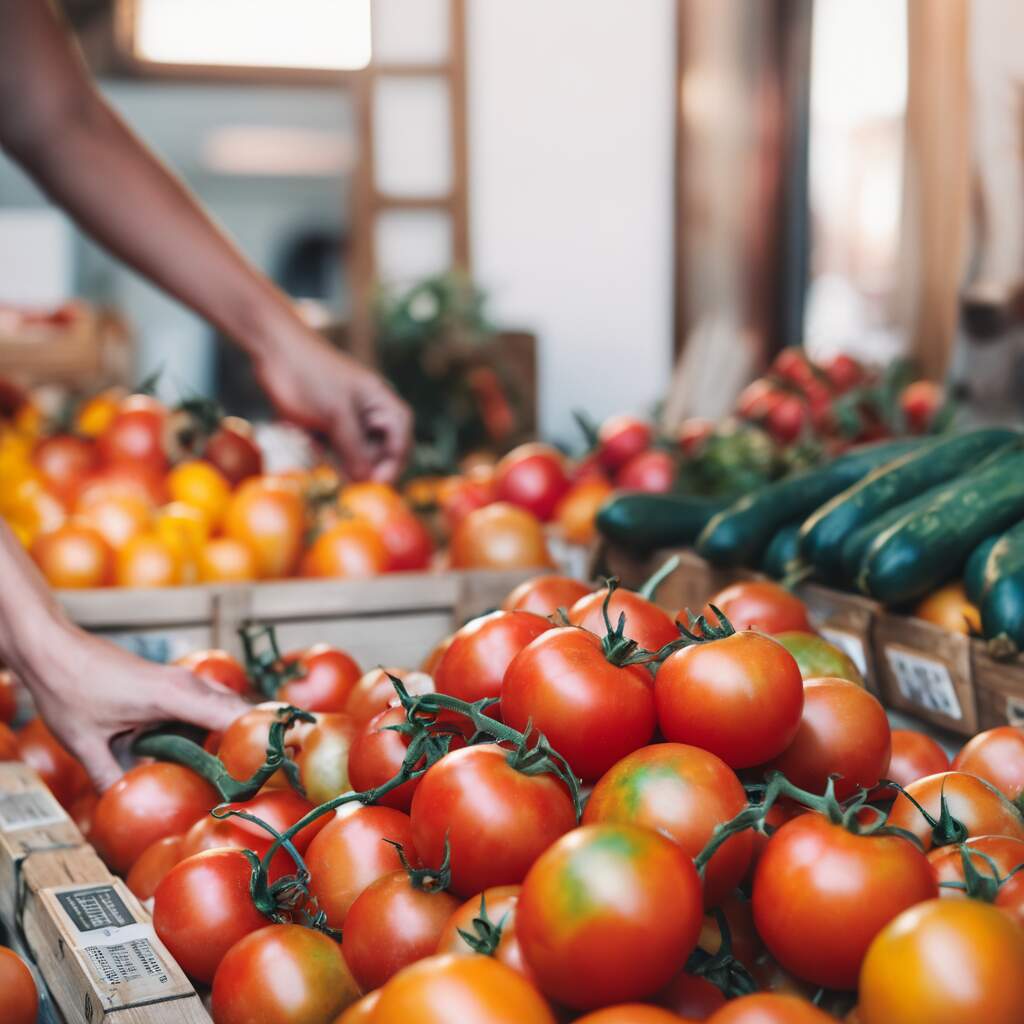}}}
        & \fbox{\includegraphics[width=0.089\textwidth]{{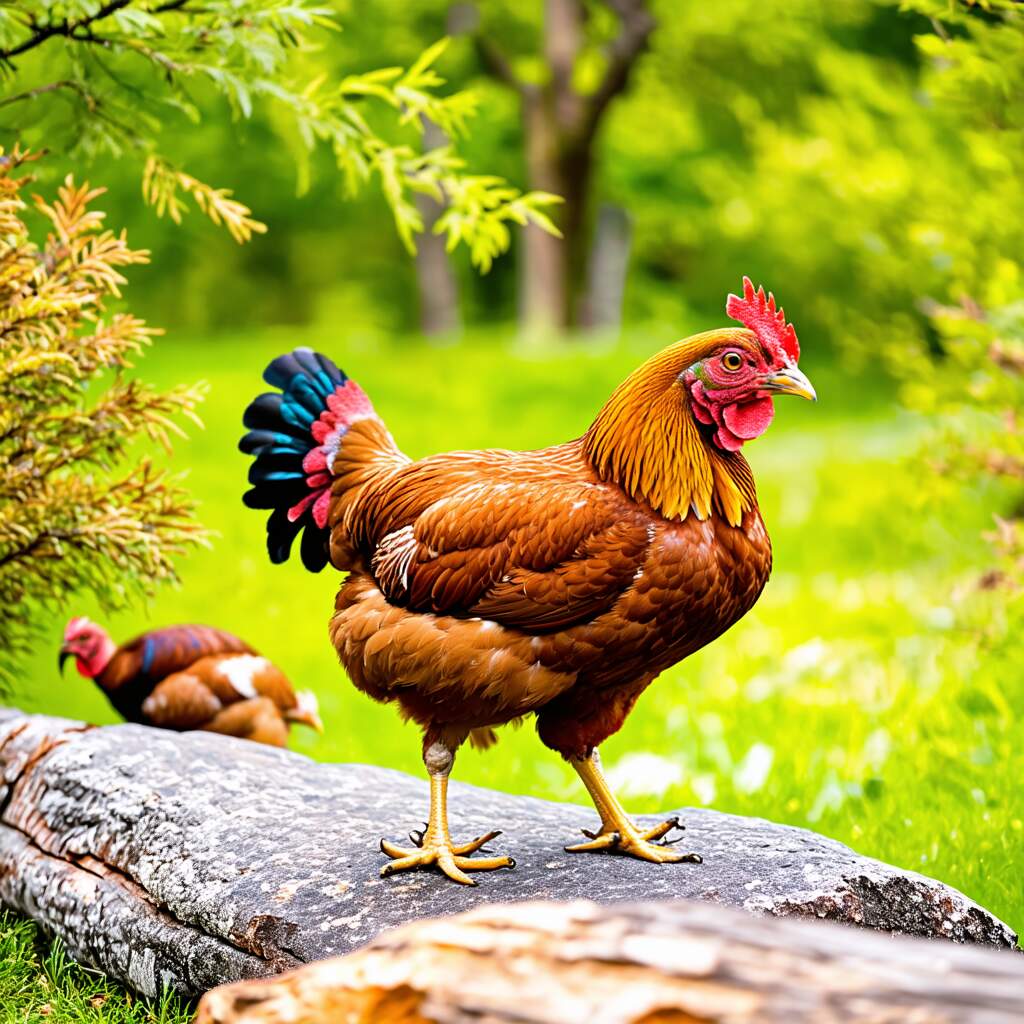}}}
        & \fbox{\includegraphics[width=0.089\textwidth]{{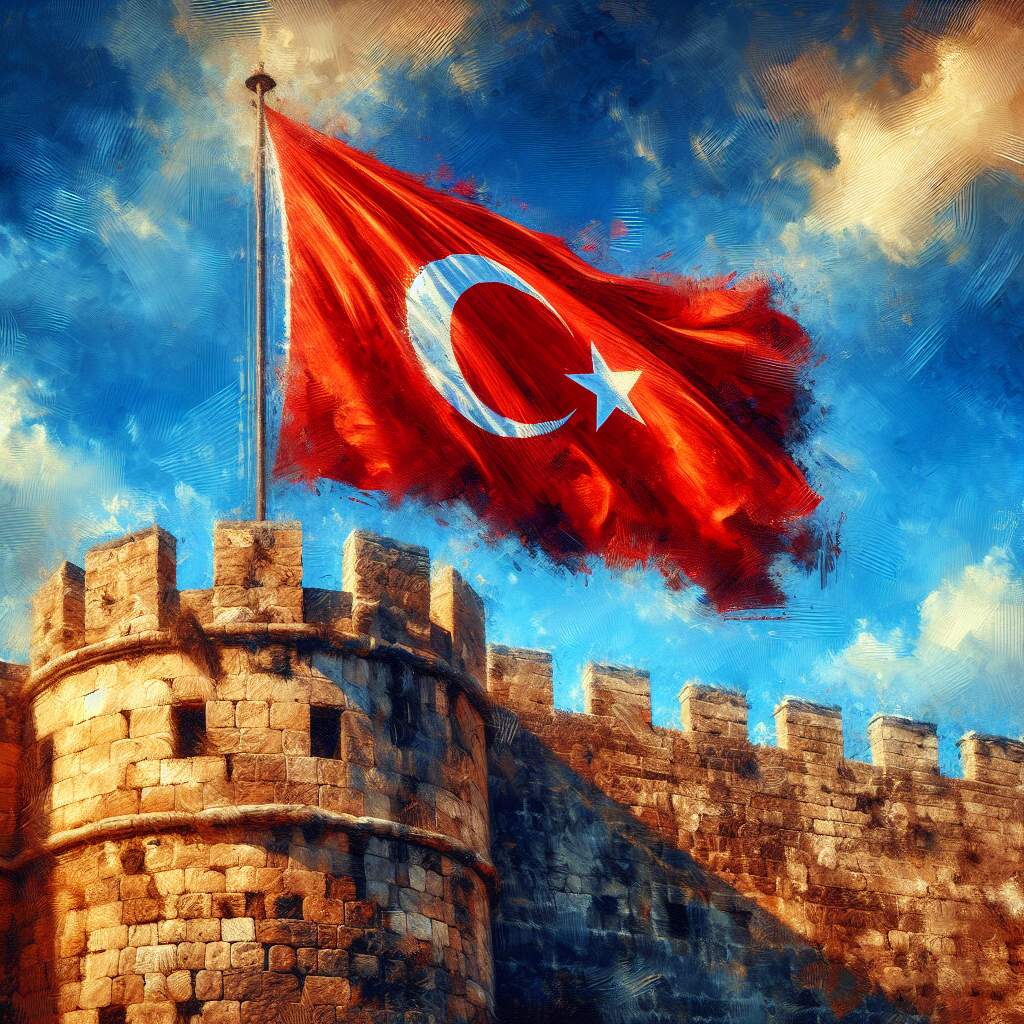}}}
        & \fbox{\includegraphics[width=0.089\textwidth]{{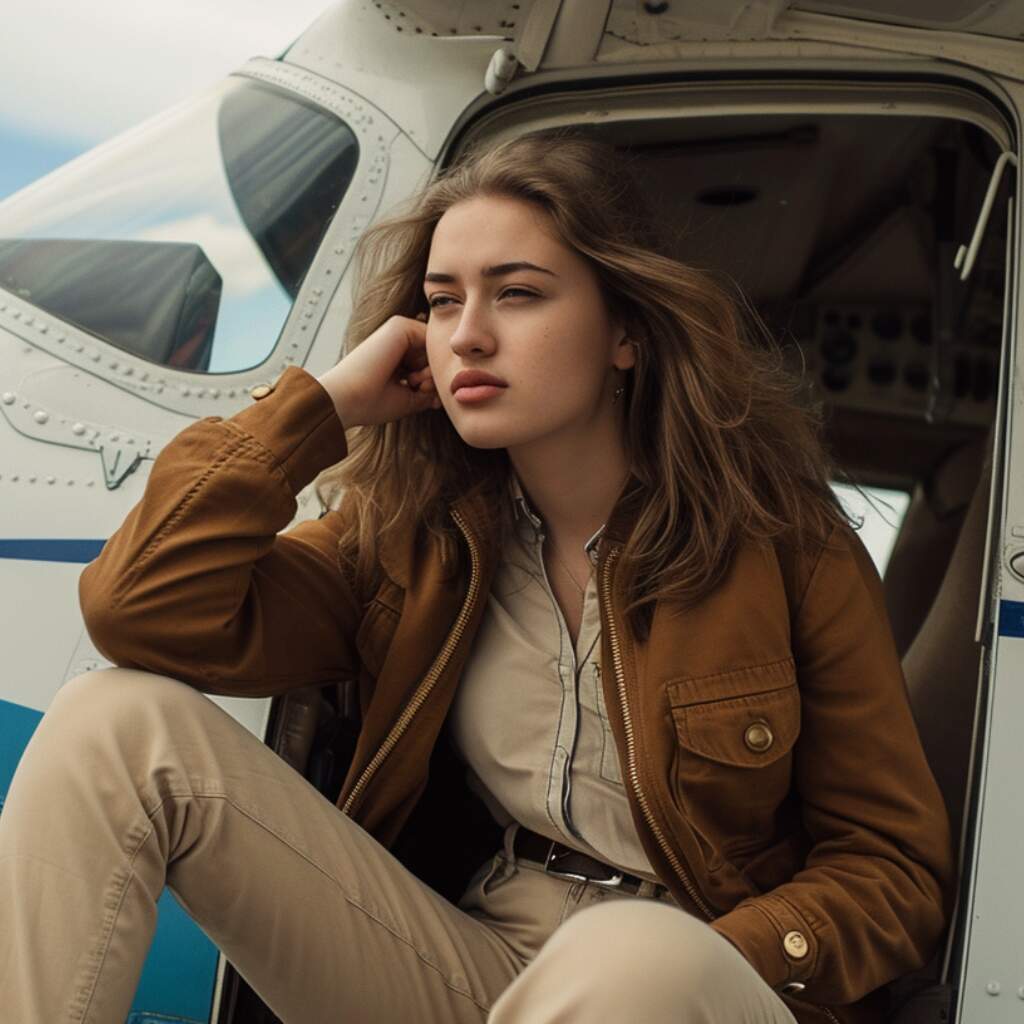}}}
        & \fbox{\includegraphics[width=0.089\textwidth]{{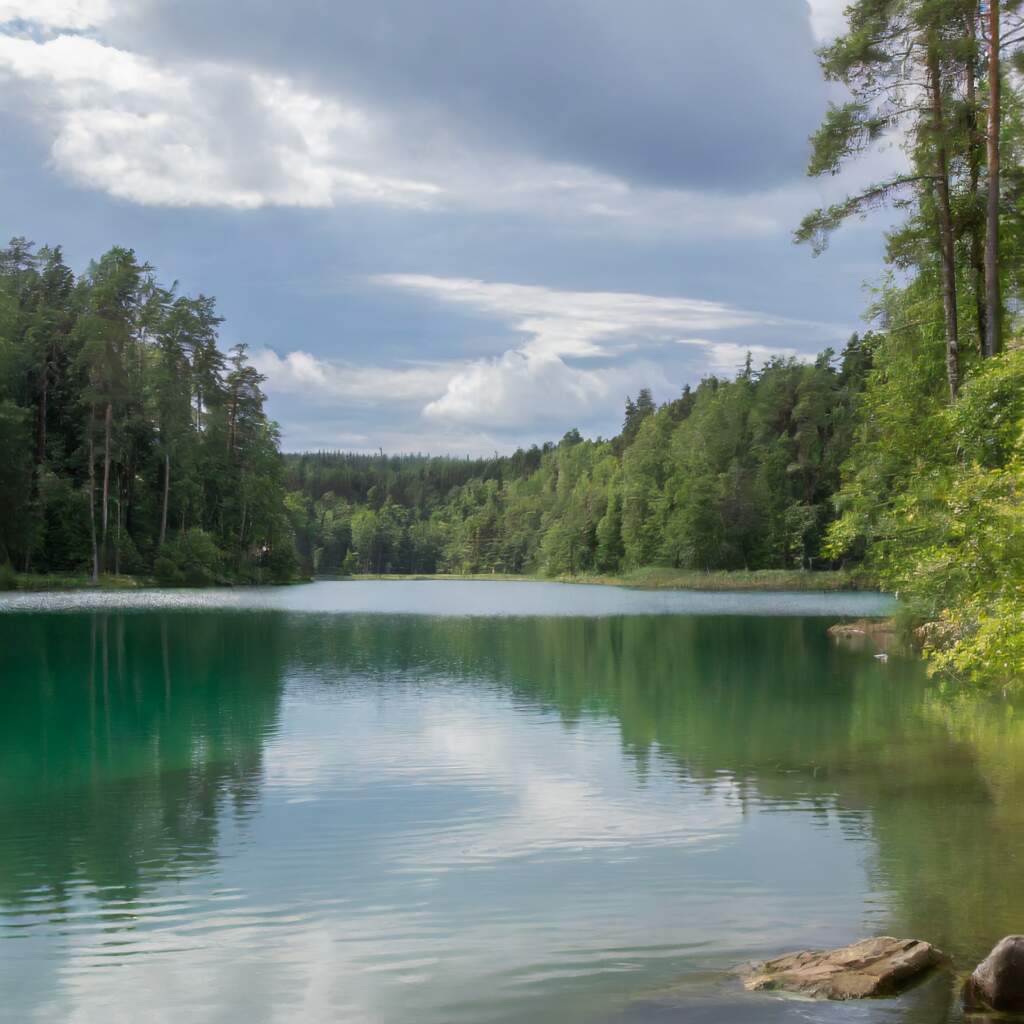}}} \\

    \end{tabular}
	\pullupp
    \caption{Visualization of real (top row) and synthetic (bottom row) images in the datasets used in this paper.}
    \label{fig:data_viz}
    \pulluppp\pullupp
\end{figure*}

%% file: sections/Intro_v1.tex
\section{Introduction}
\label{sec:intro}

The rapid improvement in AI-generated image quality has made synthetic images increasingly difficult to distinguish from real ones~\cite{review1, epstein2023online}. While traditional detection methods can be trained to identify these images, they struggle to generalize to content produced by new or unseen generators. As new generative models emerge at a rapid pace, there is an urgent need for detection methods that can reliably identify images from novel sources without prior exposure~\cite{review2, review3}.

Conventional approaches to synthetic image detection and source attribution typically rely on
learning embeddings that are discriminative between real and synthetic images, or between real and a number of specific synthetic sources~\cite{CNNDet, PatchFor, DCTCNN, marra2018detection, DIF}.
While these methods are effective for sources similar to those in training,
they often fail to adapt to new generative models~\cite{corvi2023detection, UFD}.
This occurs because their objective functions tend to make them learn features that are only useful to discriminate between known sources in the training data.
Consequently, these methods often overlook features that would be critical for identifying images from new, unseen generators.

To address this problem, we propose an alternative approach (as illustrated in Fig.~1 and detailed in Fig.~\ref{fig:system_diagram}) that is both more effective and general for detecting synthetic images and attributing them to their source.
%To address this problem, we propose an alternative approach (as illustrated in Fig.~1 and detailed in Fig.~\ref{fig:system_diagram}) that's more effective and general for detecting and attributing the source of synthetic images.
%
Instead of learning a discriminative embedding space, we focus on explicitly modeling the forensic microstructures embedded in images.
It is well-established that both cameras and synthetic image generators imprint unique forensic traces in the form of statistical microstructures—subtle, pixel-level relationships that can serve as identifying features~\cite{lukas2006digital, marra2019gans, zhao2016computationally, ChangWIFS2019}.
To isolate these microstructures from the image content, relying on only real images, we employ a self-supervised process that learns a set of diverse predictive filters to approximate the scene content. By applying these filters, we obtain multiple distinct residuals, each captures a different aspect of the forensic microstructures. We then jointly model these residuals across multiple scales using a compact parametric model, whose parameters constitute a unique \textbf{forensic self-description} for each image. This self-description effectively encapsulates the intrinsic forensic properties of an image, allowing us to perform several challenging tasks: \textbf{(1)} zero-shot detection of synthetic images, \textbf{(2)} attribute images to their source generators in an open-set manner, and \textbf{(3)} cluster images based on their sources without any prior knowledge of the generators involved.

Through extensive experiments and ablation studies, we demonstrate that our method achieves high accuracy in zero-shot detection, open-set source attribution, and clustering, consistently outperforming competing techniques in robustness and adaptability.
%Forensic self-descriptions enable our method to generalize effectively to unseen synthetic sources, advancing the state of synthetic media forensics.

Our main contributions are summarized as follows:
\begin{enumerate}
	\item We introduce forensic self-descriptions as a way to capture intrinsic properties of the forensic microstructures in an image. We then use these descriptions to accurately perform several critical tasks related to detecting and attributing the source of synthetic images.
	\item We demonstrate that these forensic self-descriptions enable accurate zero-shot detection of synthetic images without ever seeing them.
	\item We show that forensic self-descriptions are also well-suited to perform open-set attribution and clustering, allowing precise source identification and organization of images from unknown generators.
	\item We provide comprehensive experimental validation, highlighting the robustness and generalizability of our approach across a broad set of real and synthetic sources.
\end{enumerate}

%% file: figures/SystemDiagram.tex
\begin{figure*}[!ht]
    \centering
    \includegraphics[width=0.83\linewidth]{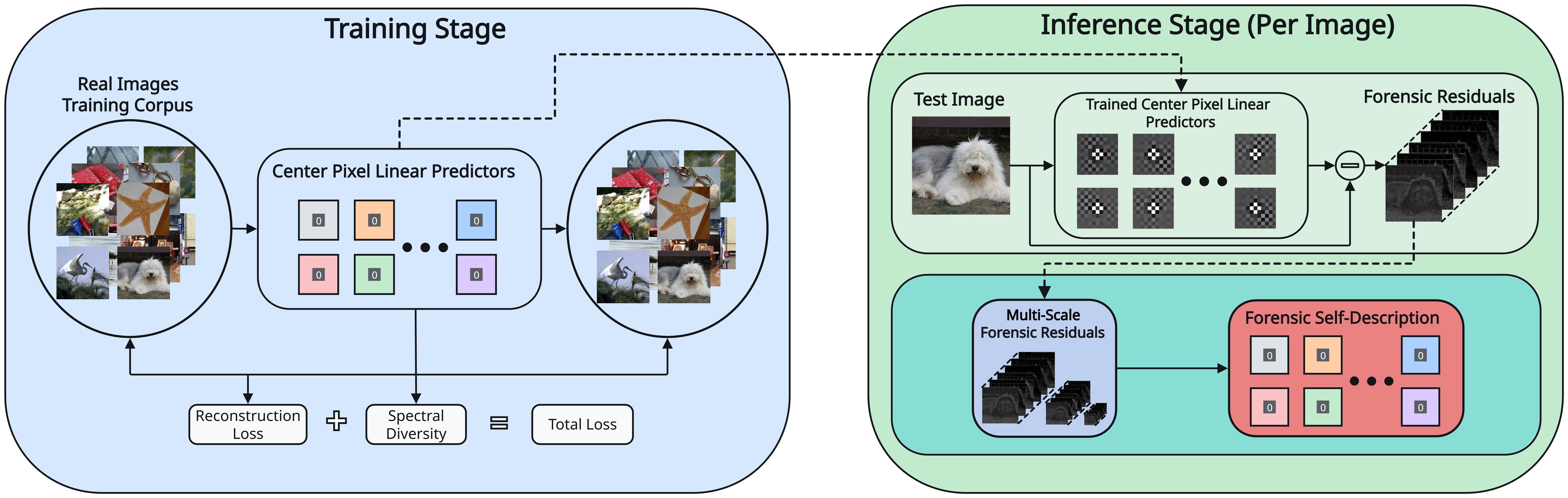}
    \caption{Our method can detect and attribute synthetic images without prior knowledge of the source. We do this by extracting residuals containing forensic microstructures from a single image and jointly modeling them across scales as a forensic self-description.}
    \label{fig:system_diagram}
    \pulluppp\pullupp
\end{figure*}

%% file: sections/Background_v2.tex
\section{Background and Related Work}

The rise of realistic AI-generated images has posed significant challenges for detection and source attribution, prompting the development of supervised, open-set, and zero-shot approaches.

\subheader{Forensic Microstructures}
It is well-established that different design choices in a generator's neural architecture induce specific statistical microstructures into AI generated images~\cite{ChangWIFS2019, review1, corvi2023detection}. Leveraging this, researchers initially built handcrafted filters or explicit mathematical models to extract these microstructures for detecting synthetic images~\cite{nataraj2019detect, FakeCatcher, barni2020crossband, LIU2023103876, CozzolinoICIP}. However, recent approaches often leverage CNNs to learn these models from data, enabling more generalized detection systems.

%Initially, handcrafted filters and statistical models were employed~\cite{khanna2009scanner, kirchner2008fast, popescu2005exposing, DCTCNN}, but recent methods leverage convolutional neural networks (CNNs) to automatically learn these forensic features, enabling more performant synthetic image detection systems.

% marra2018detection, DIF

\subheader{Supervised Methods}
Supervised methods to detect synthetic images~\cite{CNNDet, Repmix, DIF, Yu_2019_ICCV, marra2019gans, E3} often train their models on binary labeled datasets. While these methods perform well on data sources similar to those in the training set, prior work has shown that they struggle with images from unseen generative models~\cite{corvi2023detection, UFD}. This is because their learned features are specific to the training data, and may not capture the unique artifacts of new generators~\cite{review2, review3}.

% However, while , they struggle with images from unseen generative models, which often introduce unique artifacts that may not be captured by the learned features~\cite{corvi2023detection, UFD}.
%This limitation makes them ineffective in scenarios where new types of synthetic images continually emerge.

\subheader{Open-Set Source Attribution}
To address the limitations of supervised methods, researchers have recently explored adapting open-set recognition techniques developed from other computer vision areas~\cite{bendale2016towards, chen2020learning, neal2018open, Kong_2021_ICCV, zhou2021learning} to synthetic image source attribution. Notable works are POSE~\cite{POSE}, Fang et al.~\cite{Fang_2023_BMVC}, and Abady et al~\cite{Abady}. While these methods have better generalization than supervised ones, they still
heavily rely on feature representations learned from known sources, which may not generalize well to unseen ones.
%In addition, open-set methods require careful calibration to balance false positives and false negatives~\cite{POSE}, which can be challenging in practice.

\subheader{Zero-Shot Detection}
Recent work has developed approaches to detect synthetic images without requiring exposure to specific generative models. These methods typically rely on non-forensic features that differ between real and synthetic images. For instance, some methods~\cite{Aeroblade,Zed} use autoencoders (i.e., diffusion model, image compression network) for reconstruction error analysis, while others~\cite{DE-Fake, UFD} leverage CLIP embeddings to detect inconsistencies in general visual features. Few others~\cite{LGrad, NPR} use a limited set of forensic features for generalized detection.

While promising, as we show later, these zero-shot methods often yield inconsistent performance, which varies depending on the real-vs-synthetic dataset pairs used for benchmarking. This variability arises because non-forensic features may be influenced by the specific content characteristics of the datasets. Furthermore, there is no guarantee that these  features will remain effective as generative technologies continue to improve and evolve.

\subheader{Unsupervised Clustering}
Research to accomplish this task for synthetic images has been largely under-explored. Girish et al.~\cite{GanDiscovery} proposed a way to discover new GAN generators by over-clustering embeddings from a simple CNN. Yang et al.~\cite{POSE} proposed a new open-set method that can be leveraged to perform clustering. Overall, without any supervision, accurately clustering images based on their source remains very challenging.

%In summary, existing methods face challenges in generalizing to unseen generators. Supervised methods depend on labeled data, while open-set and zero-shot approaches may miss nuanced artifacts from novel generators. Unsupervised clustering is limited by the need for high-quality embeddings to capture class differences. Our approach overcomes these limitations by modeling forensic microstructures per image, enabling robust zero-shot detection, open-set attribution, and clustering without prior exposure to specific synthetic sources.

%% file: figures/SelfDescViz.tex
\begin{figure*}[!t]
    \centering
    \setlength{\fboxsep}{0pt}
    \setlength{\fboxrule}{0pt}
    \resizebox{0.80\linewidth}{!}{
    \begin{tabular}{
            p{0.095\textwidth}
            @{\hskip -2pt}p{0.090\textwidth}@{\hskip 4pt}
            !{\vrule width 1pt}
            @{\hskip 2pt}
            *{8}{@{\hskip 2pt} p{0.090\textwidth} @{\hskip 2pt}}
        } % Added a tabular environment with a vertical line
        & \makebox[0.089\textwidth]{\raisebox{1pt}{\fontsize{9pt}{10.5pt}\selectfont Real}}
        & \makebox[0.089\textwidth]{\raisebox{1pt}{\fontsize{9pt}{10.5pt}\selectfont ProGAN}}
        & \makebox[0.089\textwidth]{\raisebox{1pt}{\fontsize{9pt}{10.5pt}\selectfont GLIDE}}
        & \makebox[0.089\textwidth]{\raisebox{1pt}{\fontsize{9pt}{10.5pt}\selectfont StyleGAN3}}
        & \makebox[0.089\textwidth]{\raisebox{1pt}{\fontsize{9pt}{10.5pt}\selectfont SD 1.5}}
        & \makebox[0.089\textwidth]{\raisebox{1pt}{\fontsize{9pt}{10.5pt}\selectfont SDXL}}
        & \makebox[0.089\textwidth]{\raisebox{1pt}{\fontsize{9pt}{10.5pt}\selectfont SD 3.0}}
        & \makebox[0.089\textwidth]{\raisebox{1pt}{\fontsize{9pt}{10.5pt}\selectfont DALLE 3}}
        & \makebox[0.089\textwidth]{\raisebox{1pt}{\fontsize{9pt}{10.5pt}\selectfont MJ v6}} \\

        \raisebox{19.5pt}{\fontsize{9pt}{10.5pt}\selectfont $FFT(\phi_1)$\hspace{2pt}}
        & \fbox{\includegraphics[width=0.089\textwidth]{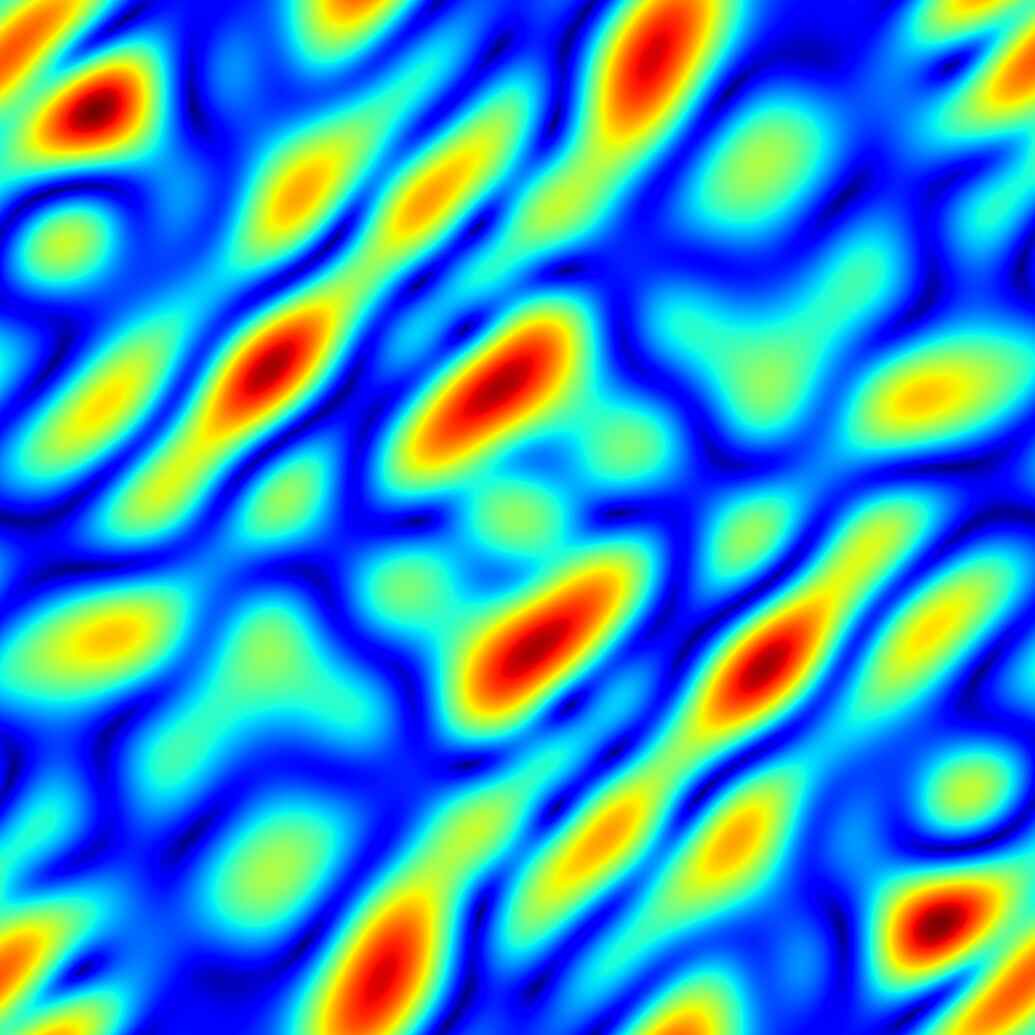}}
        & \fbox{\includegraphics[width=0.089\textwidth]{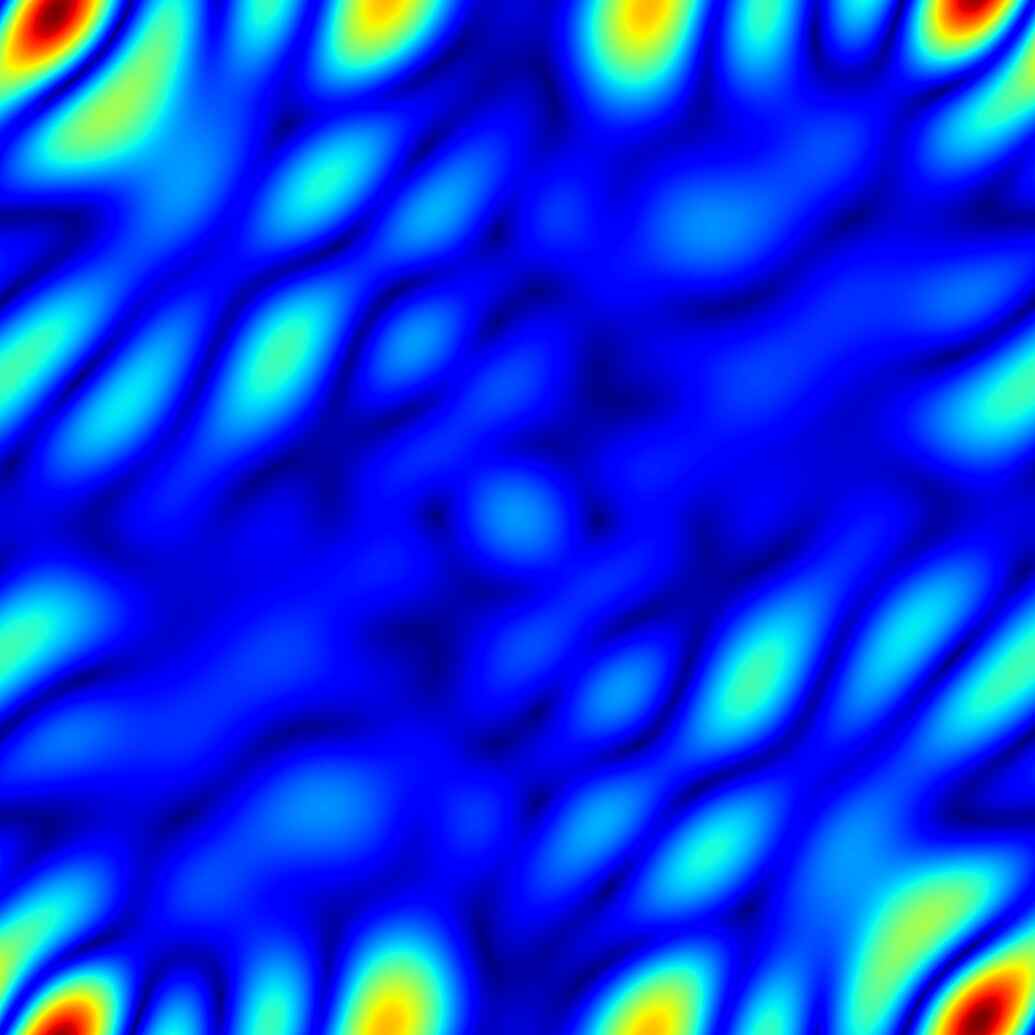}}
        & \fbox{\includegraphics[width=0.089\textwidth]{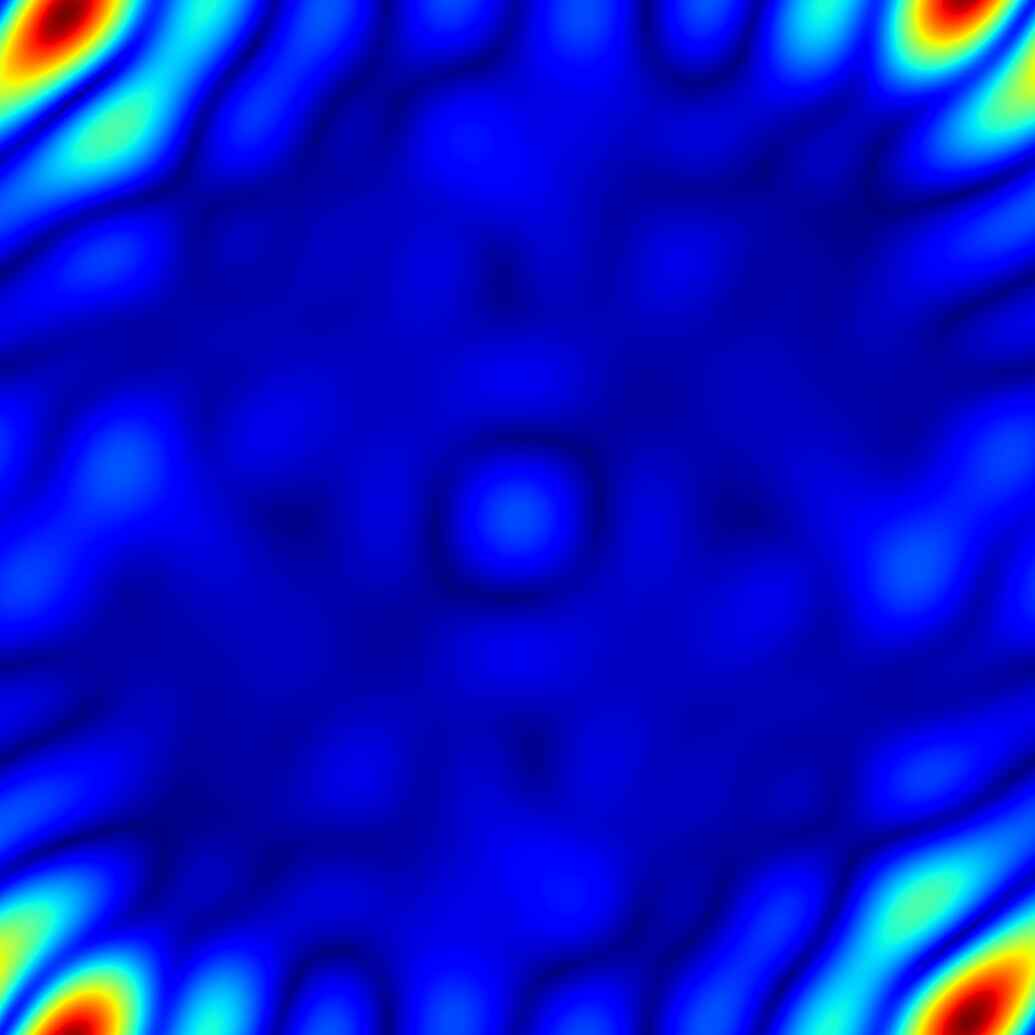}}
        & \fbox{\includegraphics[width=0.089\textwidth]{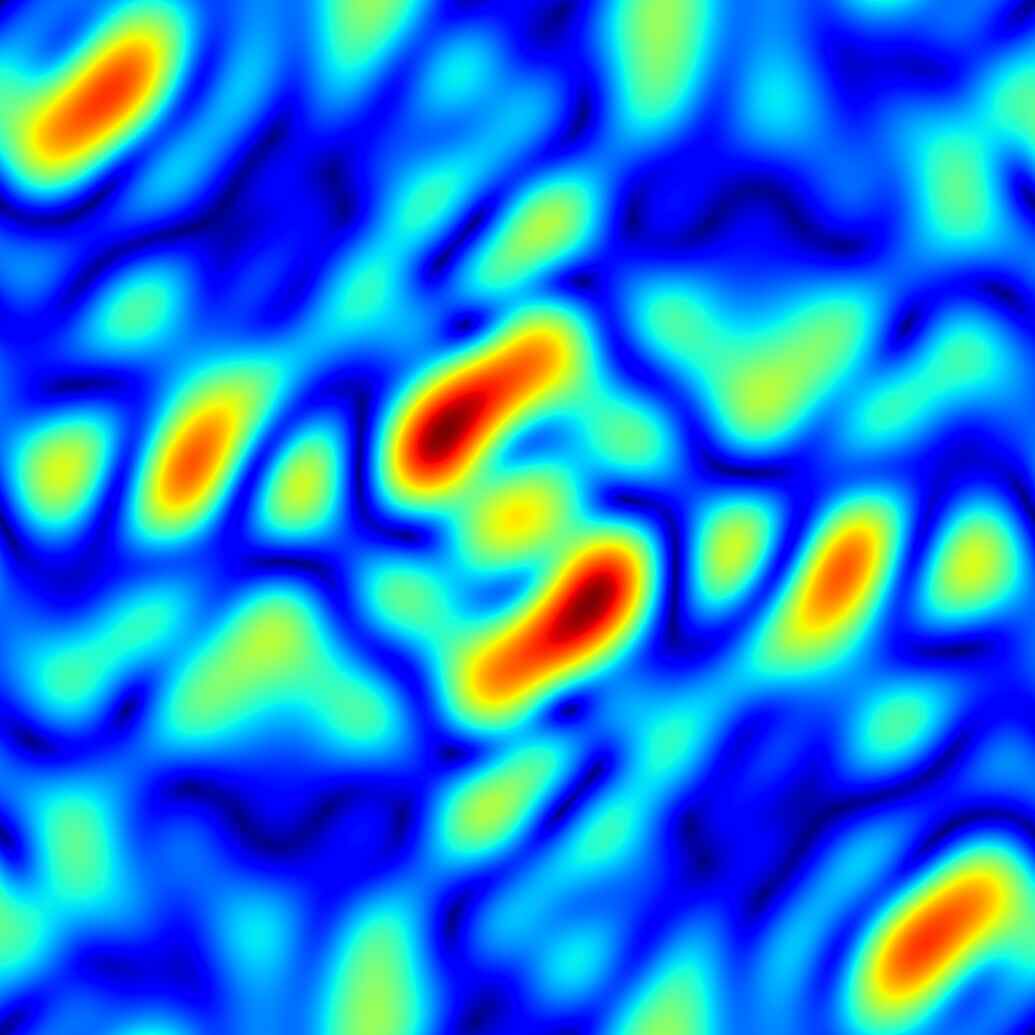}}
        & \fbox{\includegraphics[width=0.089\textwidth]{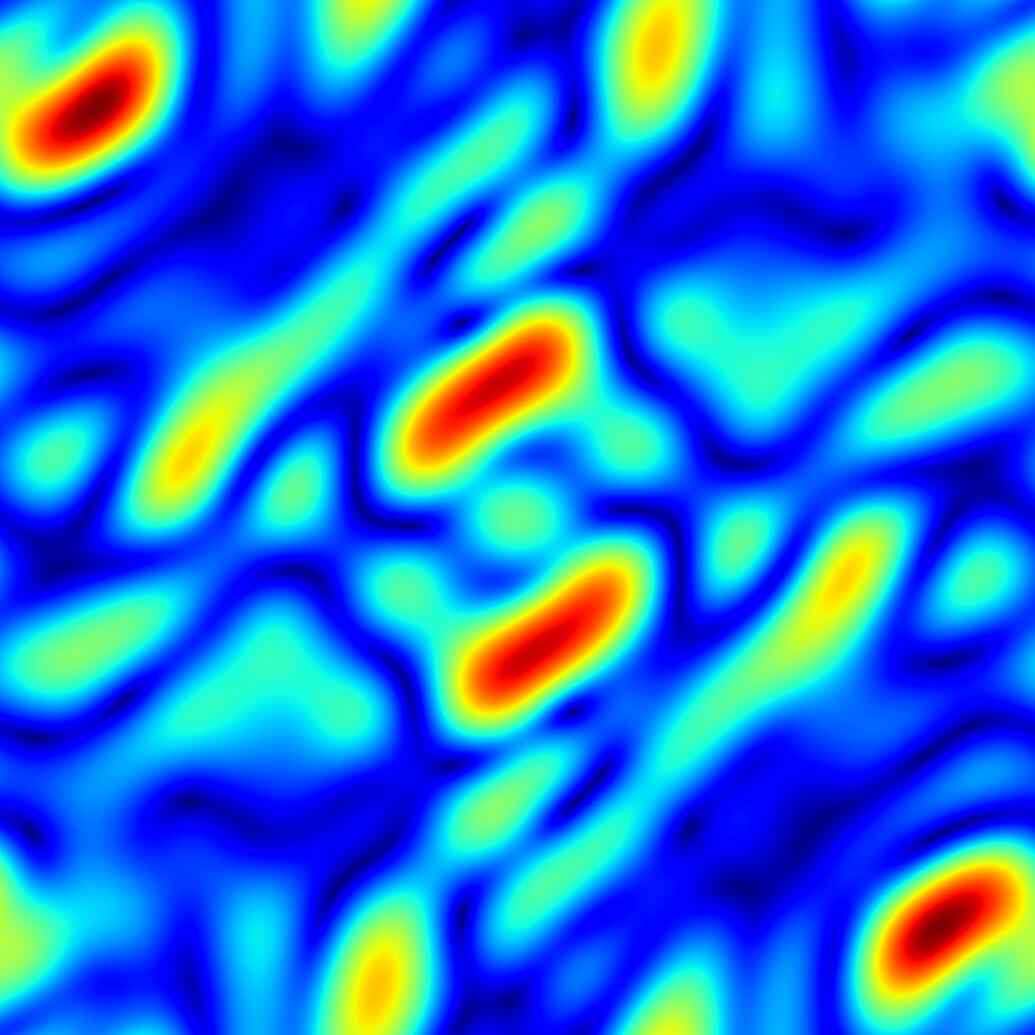}}
        & \fbox{\includegraphics[width=0.089\textwidth]{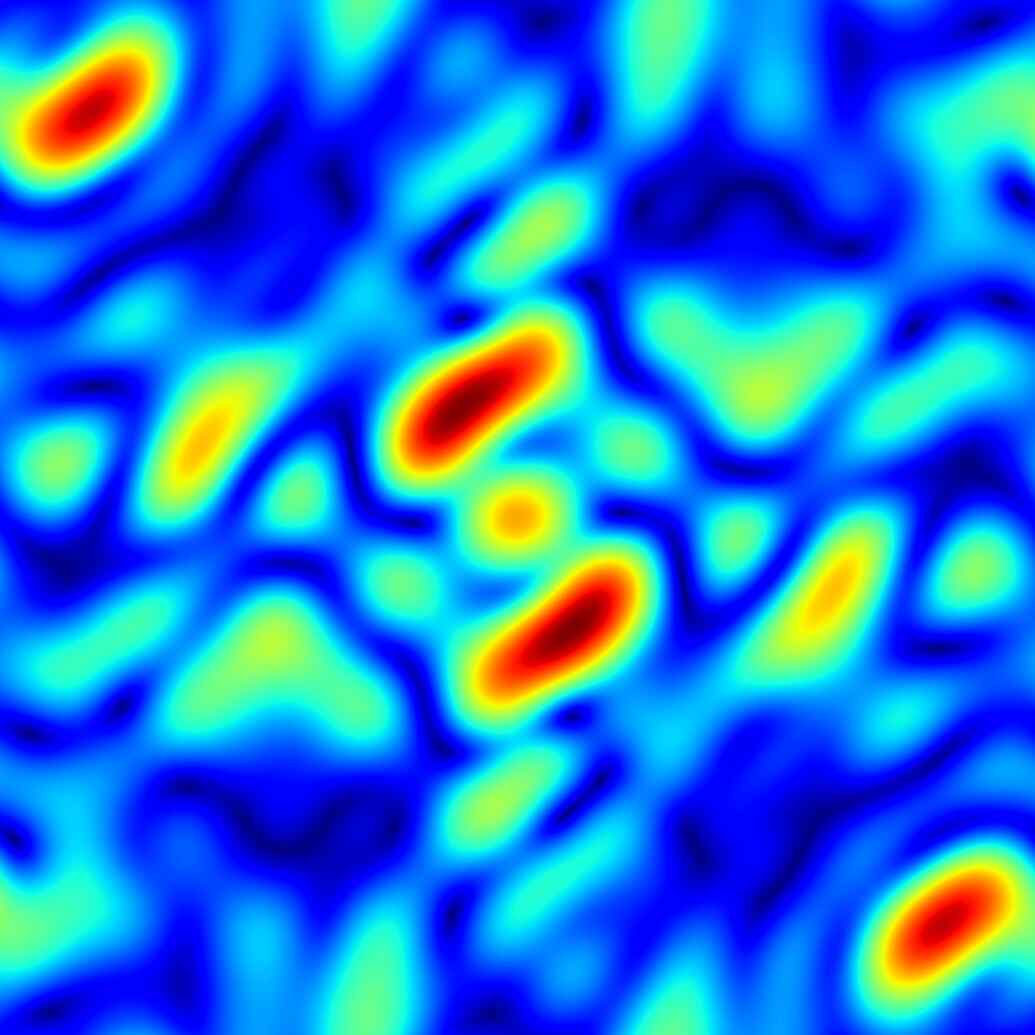}}
        & \fbox{\includegraphics[width=0.089\textwidth]{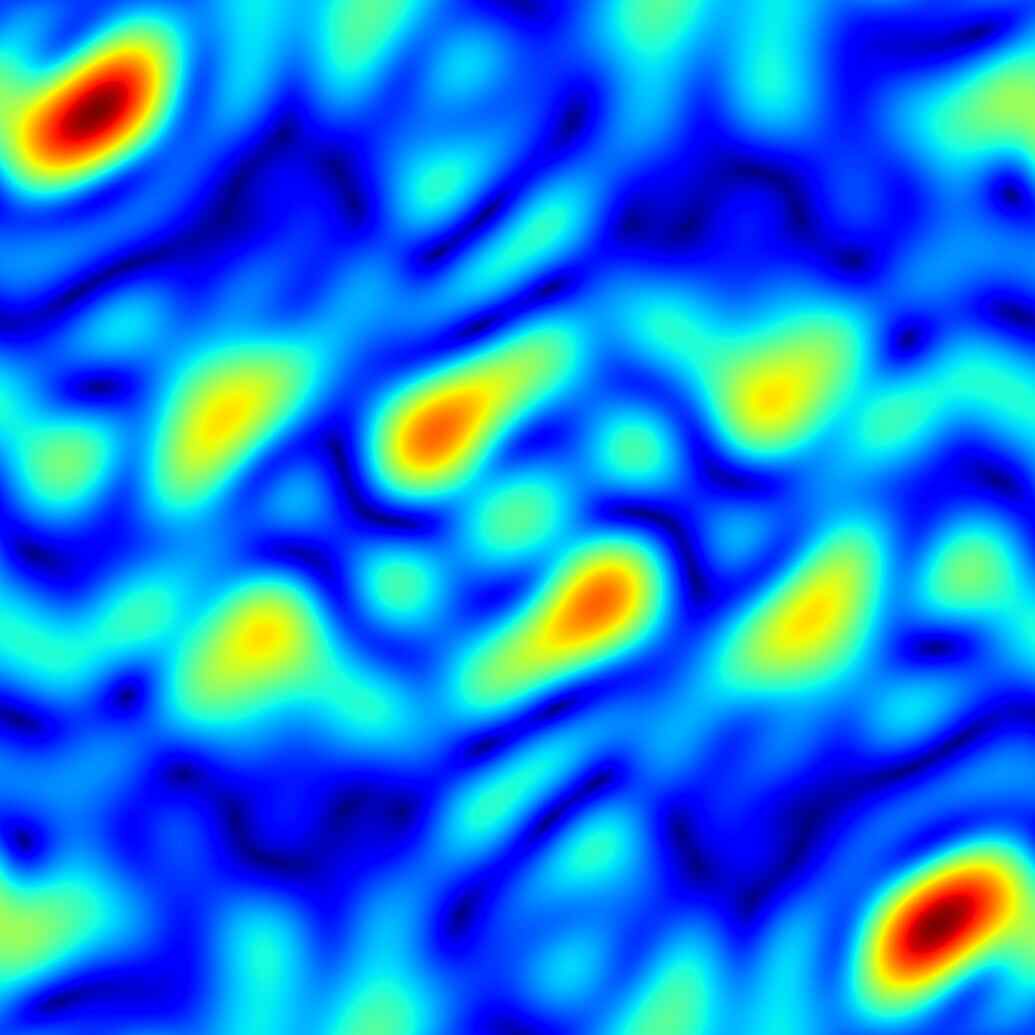}}
        & \fbox{\includegraphics[width=0.089\textwidth]{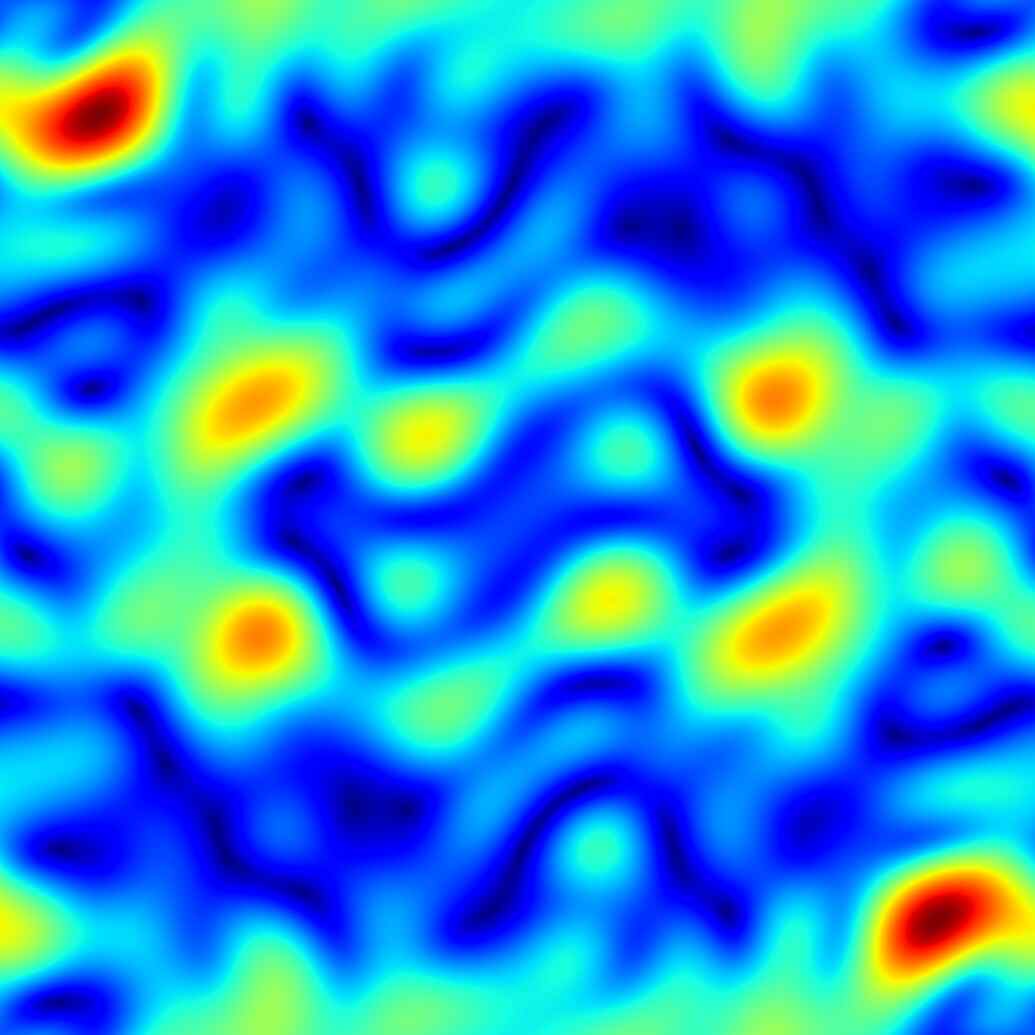}}
        & \fbox{\includegraphics[width=0.089\textwidth]{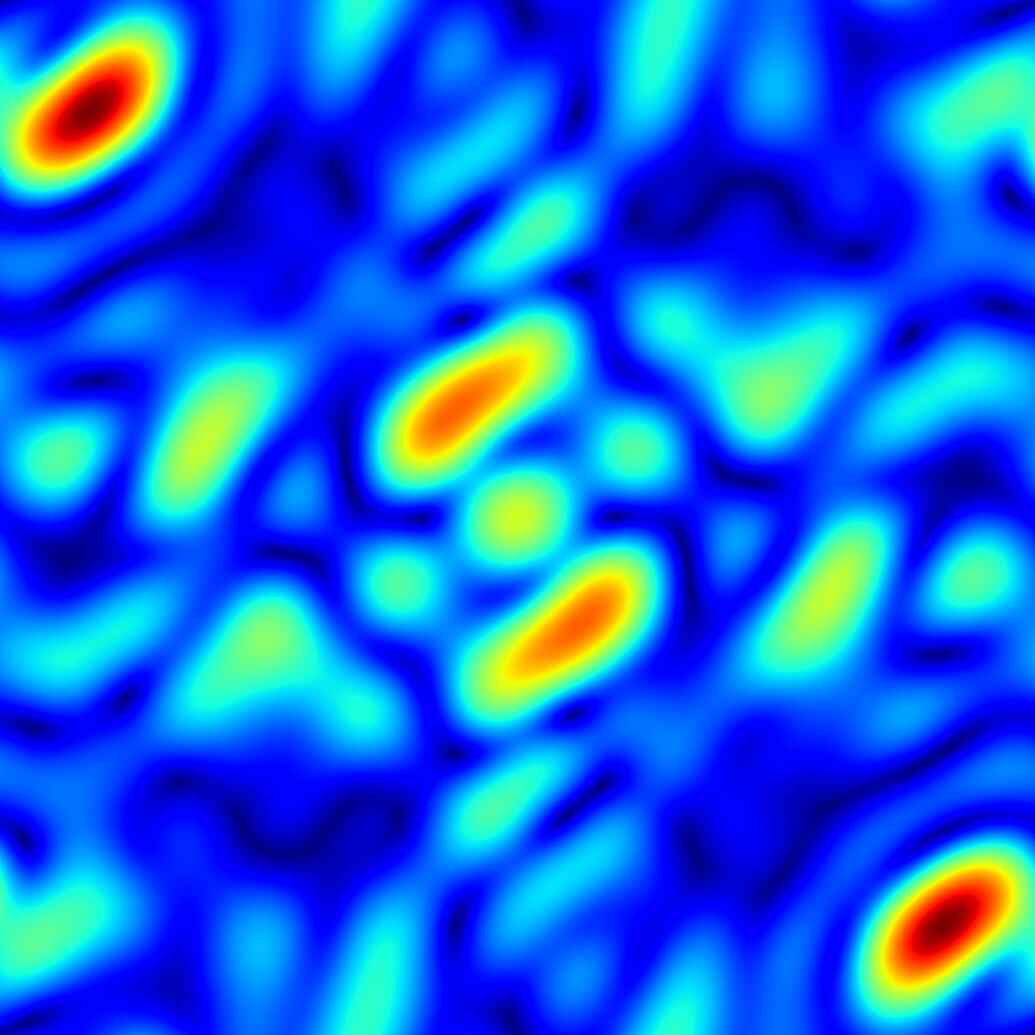}} \\

        \raisebox{19.5pt}{\fontsize{9pt}{10.5pt}\selectfont $FFT(\phi_2)$\hspace{2pt}}
        & \fbox{\includegraphics[width=0.089\textwidth]{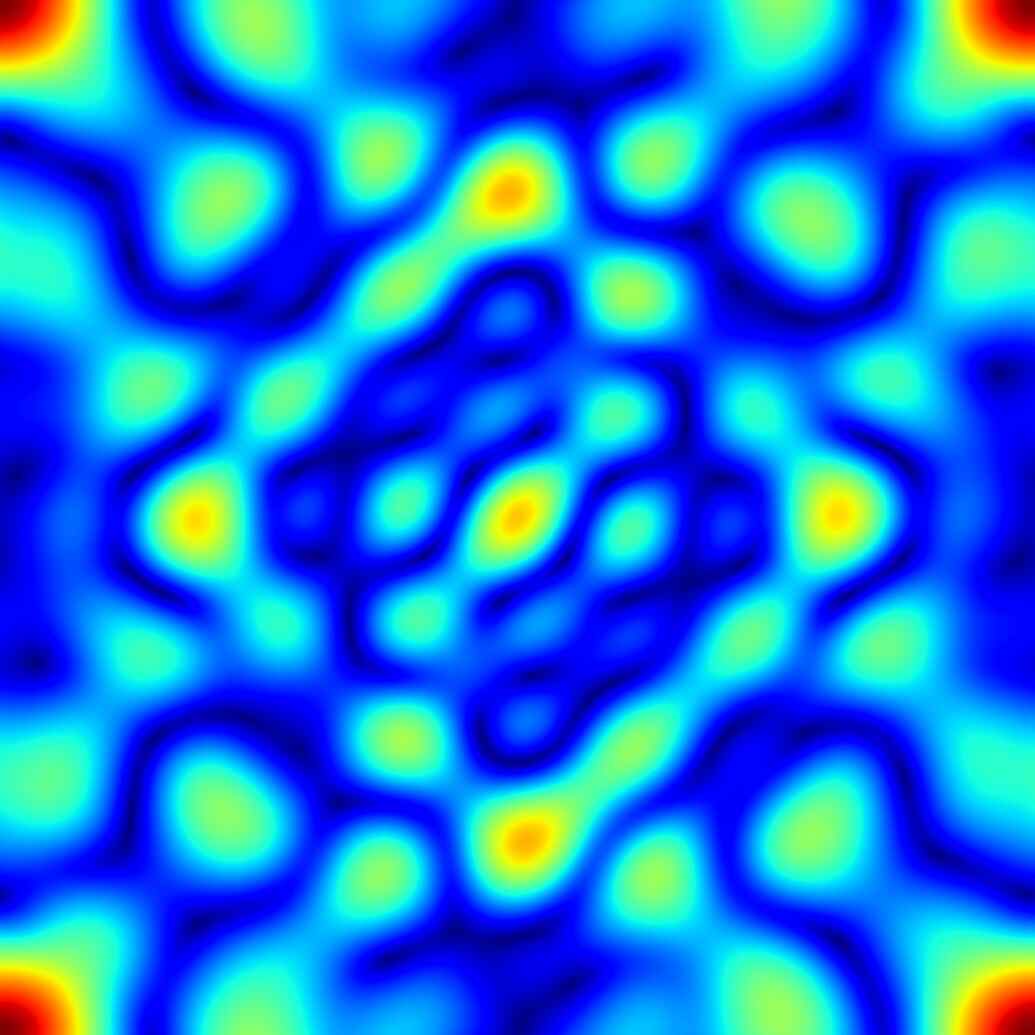}}
        & \fbox{\includegraphics[width=0.089\textwidth]{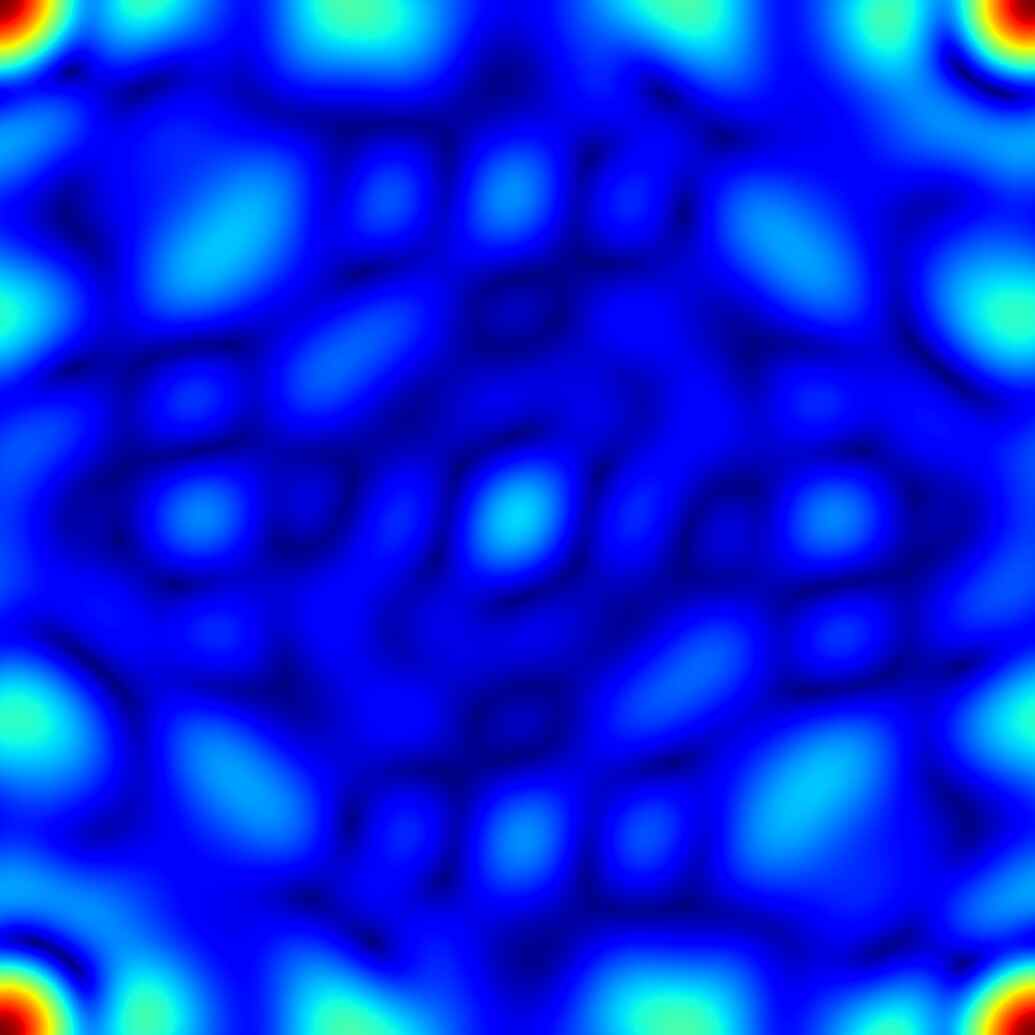}}
        & \fbox{\includegraphics[width=0.089\textwidth]{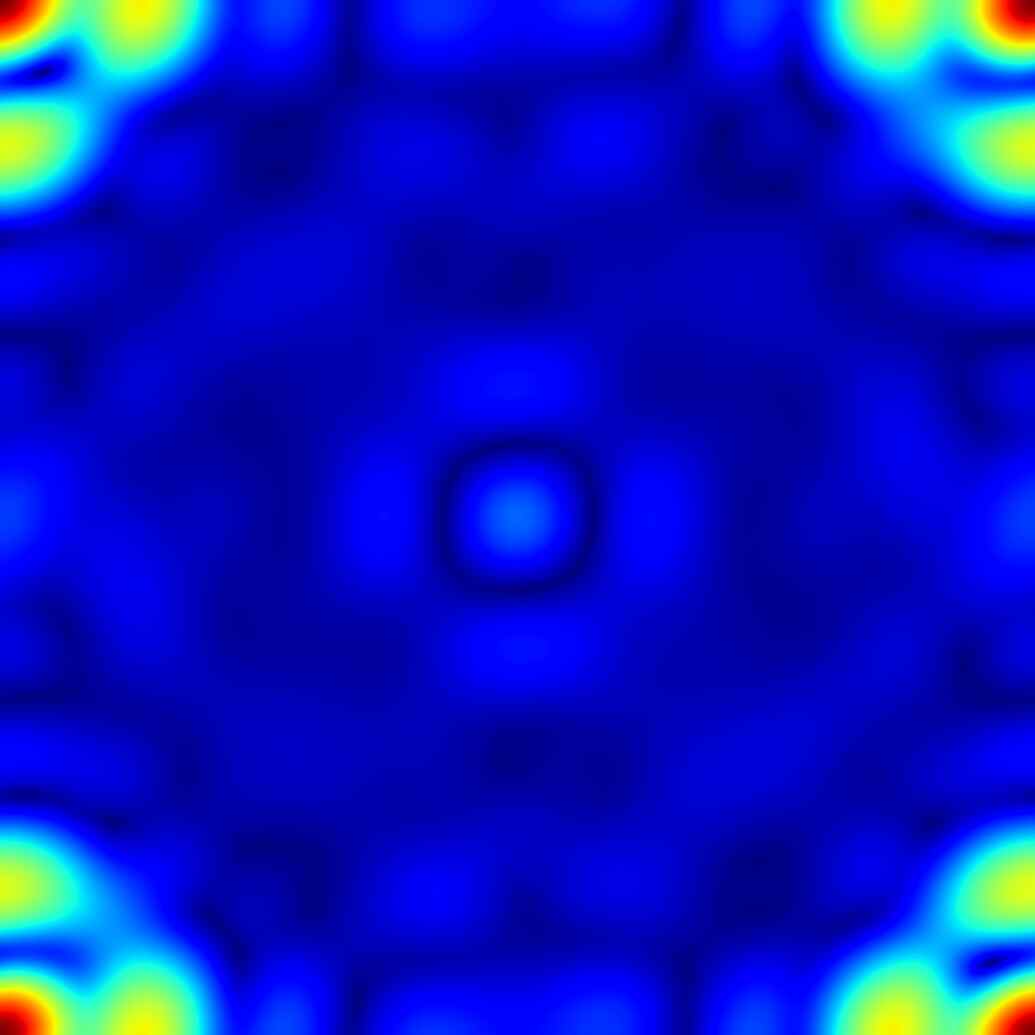}}
        & \fbox{\includegraphics[width=0.089\textwidth]{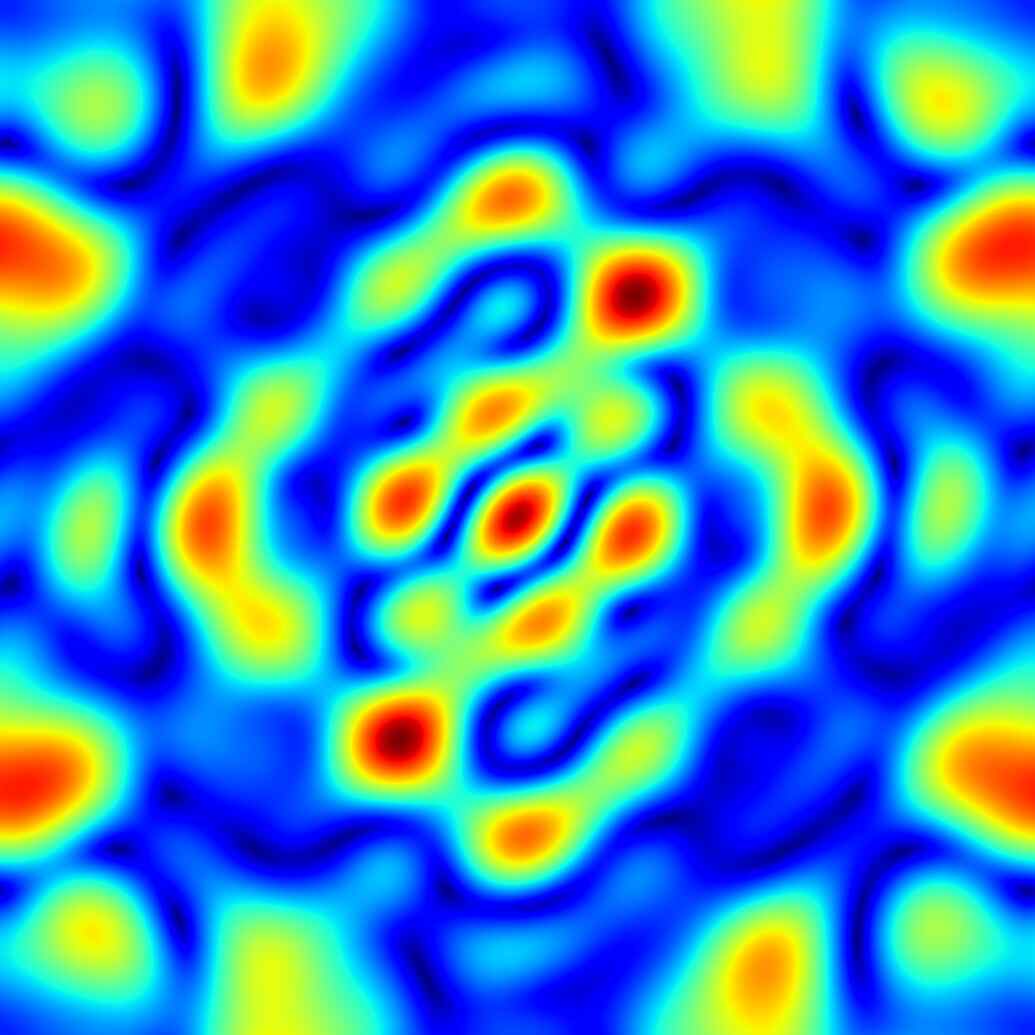}}
        & \fbox{\includegraphics[width=0.089\textwidth]{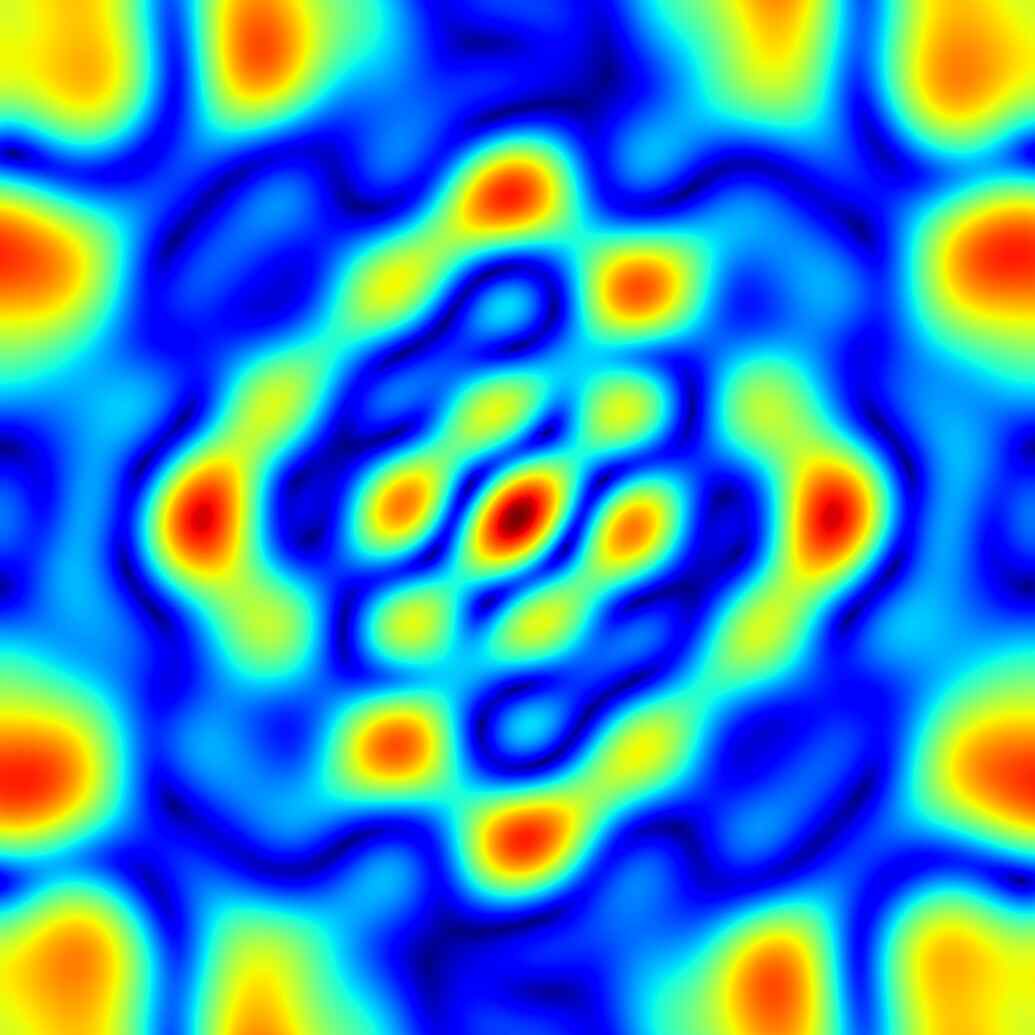}}
        & \fbox{\includegraphics[width=0.089\textwidth]{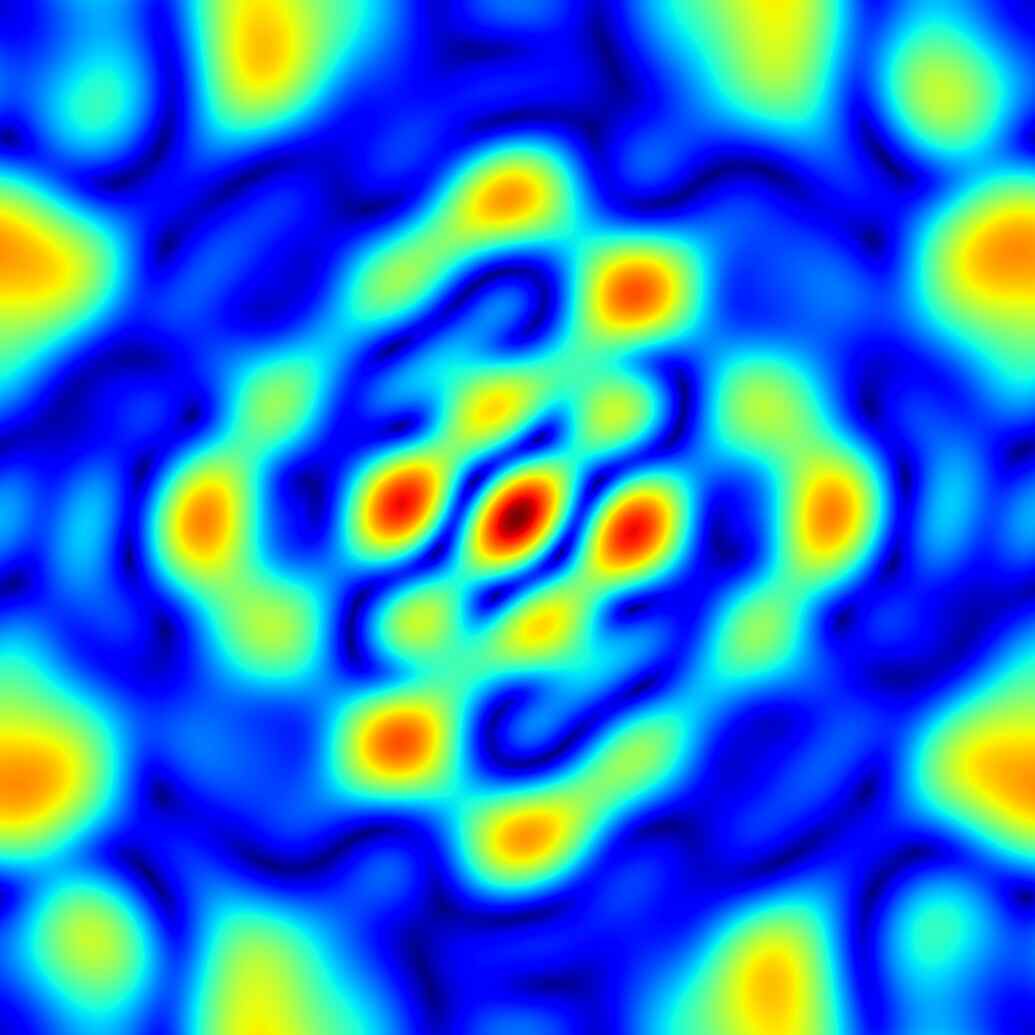}}
        & \fbox{\includegraphics[width=0.089\textwidth]{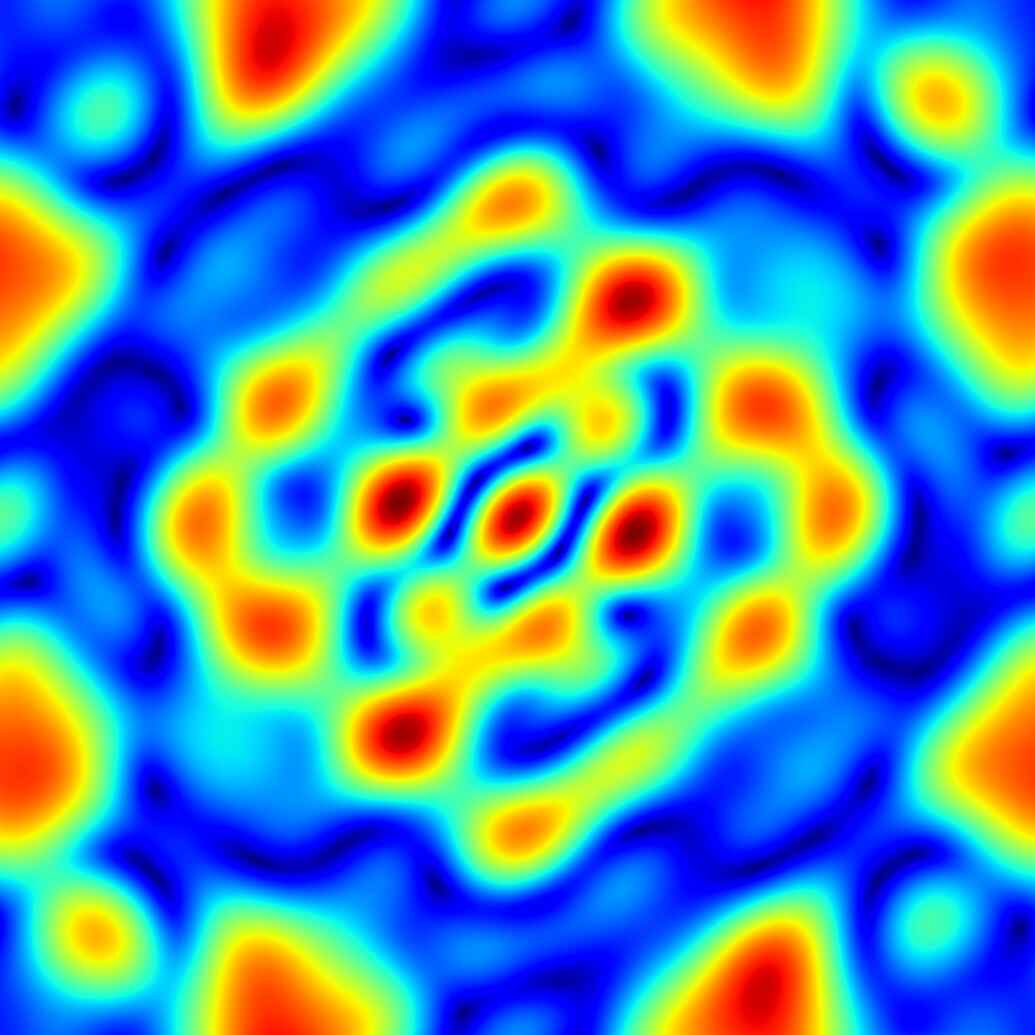}}
        & \fbox{\includegraphics[width=0.089\textwidth]{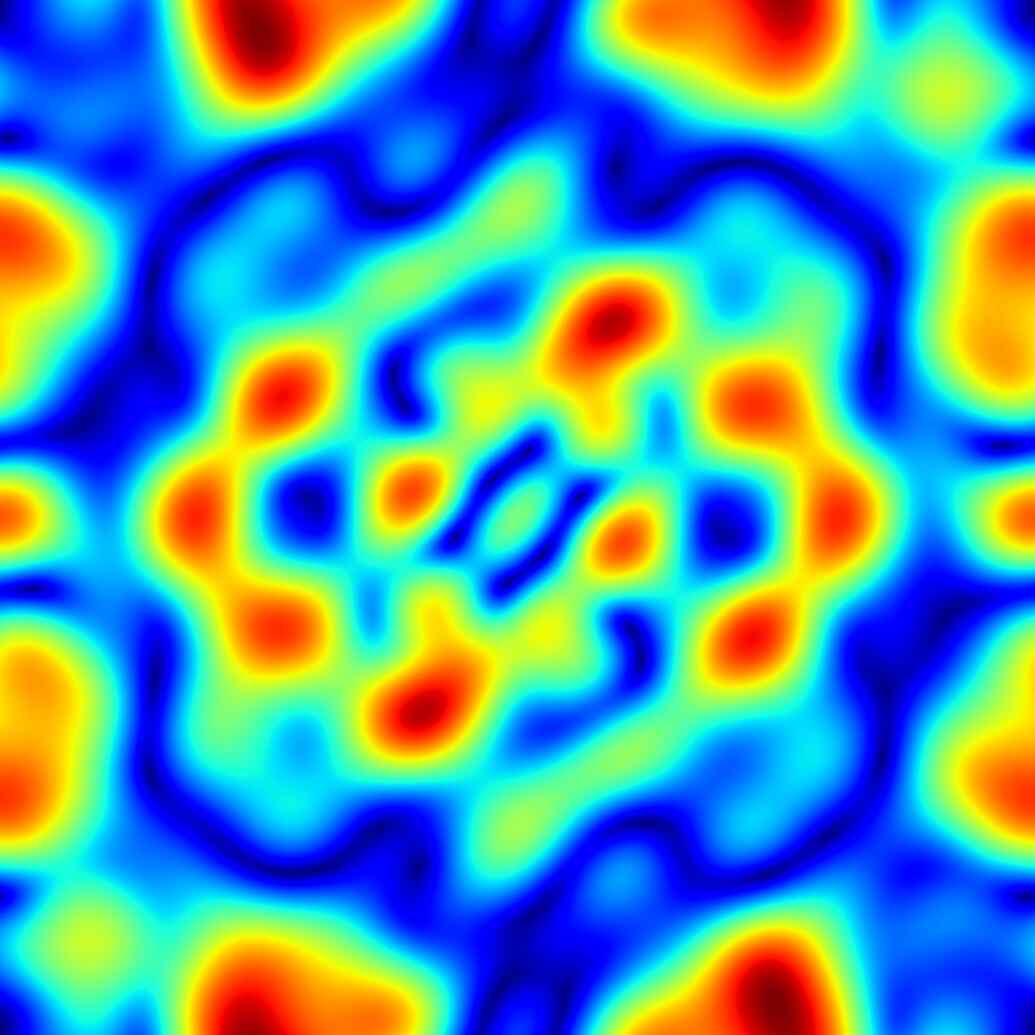}}
        & \fbox{\includegraphics[width=0.089\textwidth]{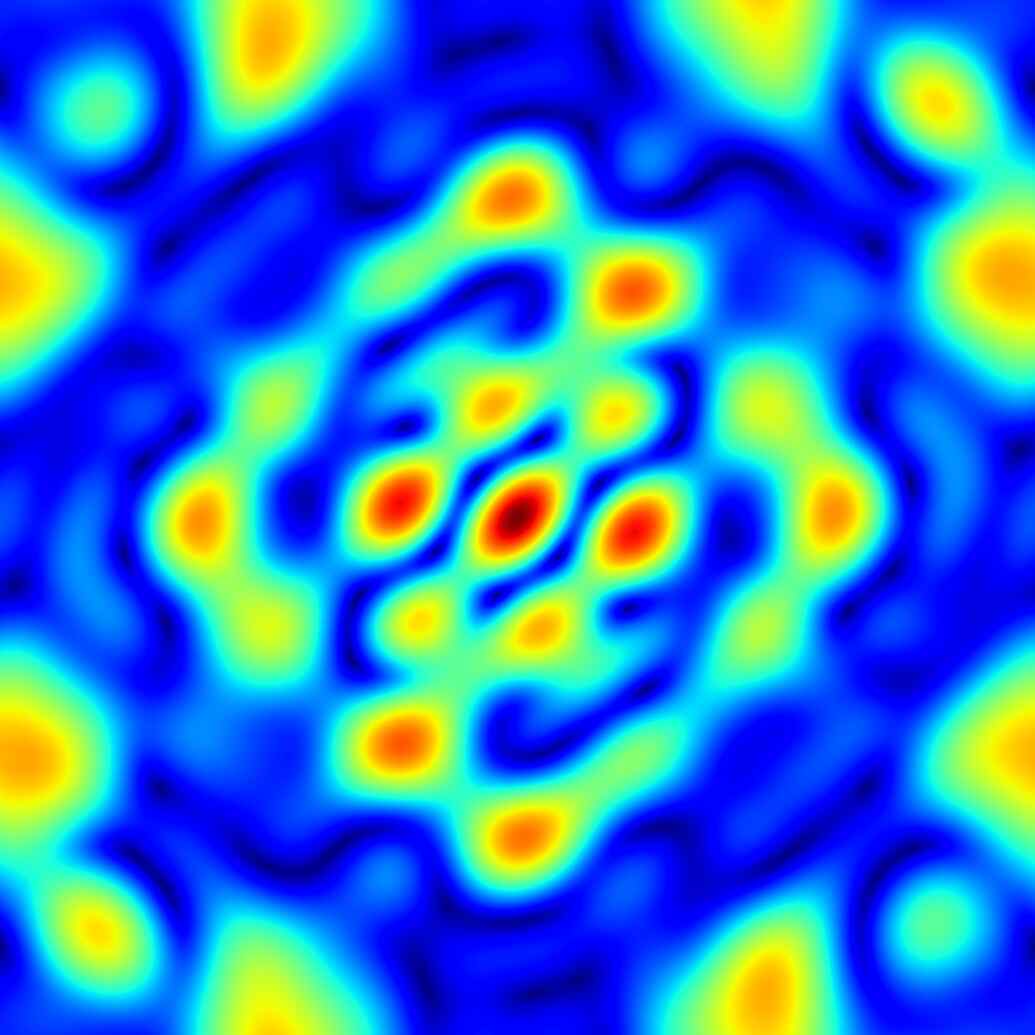}} \\

        \raisebox{19.5pt}{\fontsize{9pt}{10.5pt}\selectfont $FFT(\phi_3)$\hspace{2pt}}
        & \fbox{\includegraphics[width=0.089\textwidth]{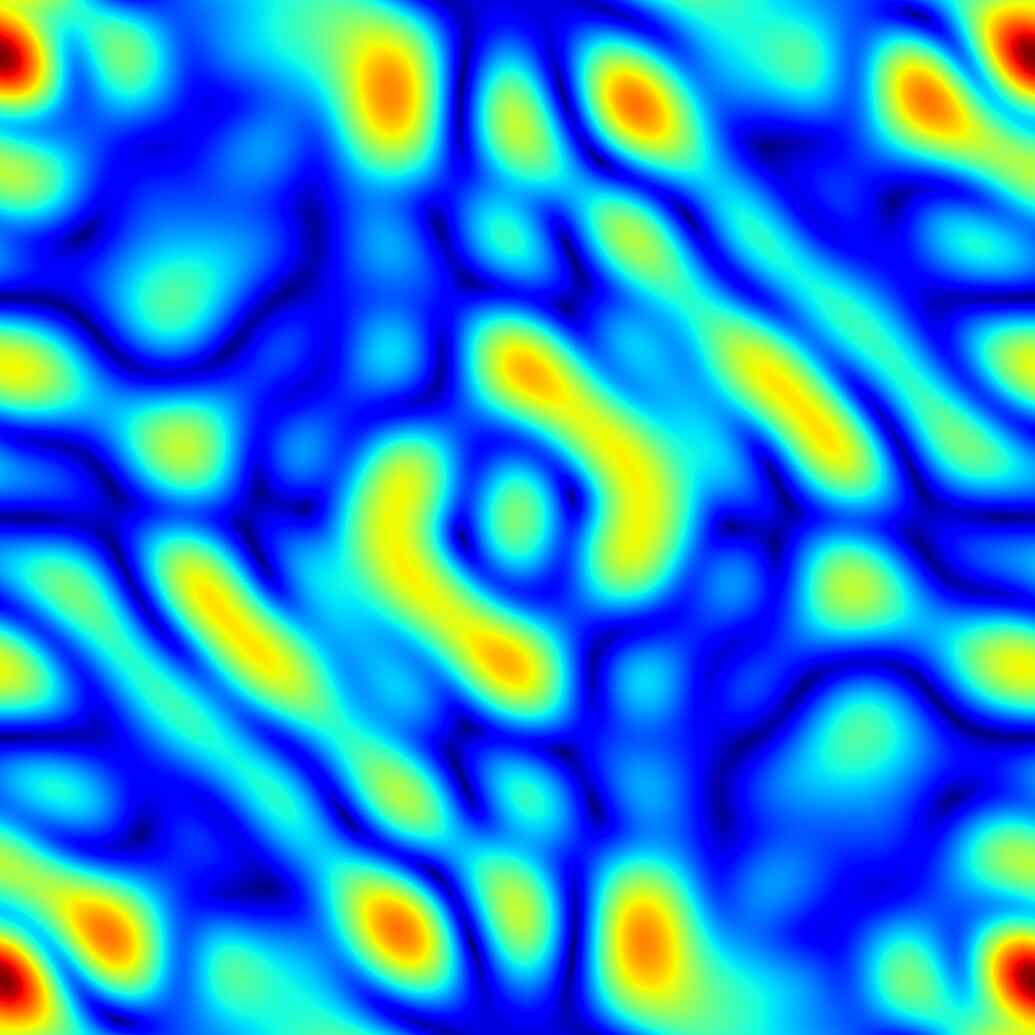}}
        & \fbox{\includegraphics[width=0.089\textwidth]{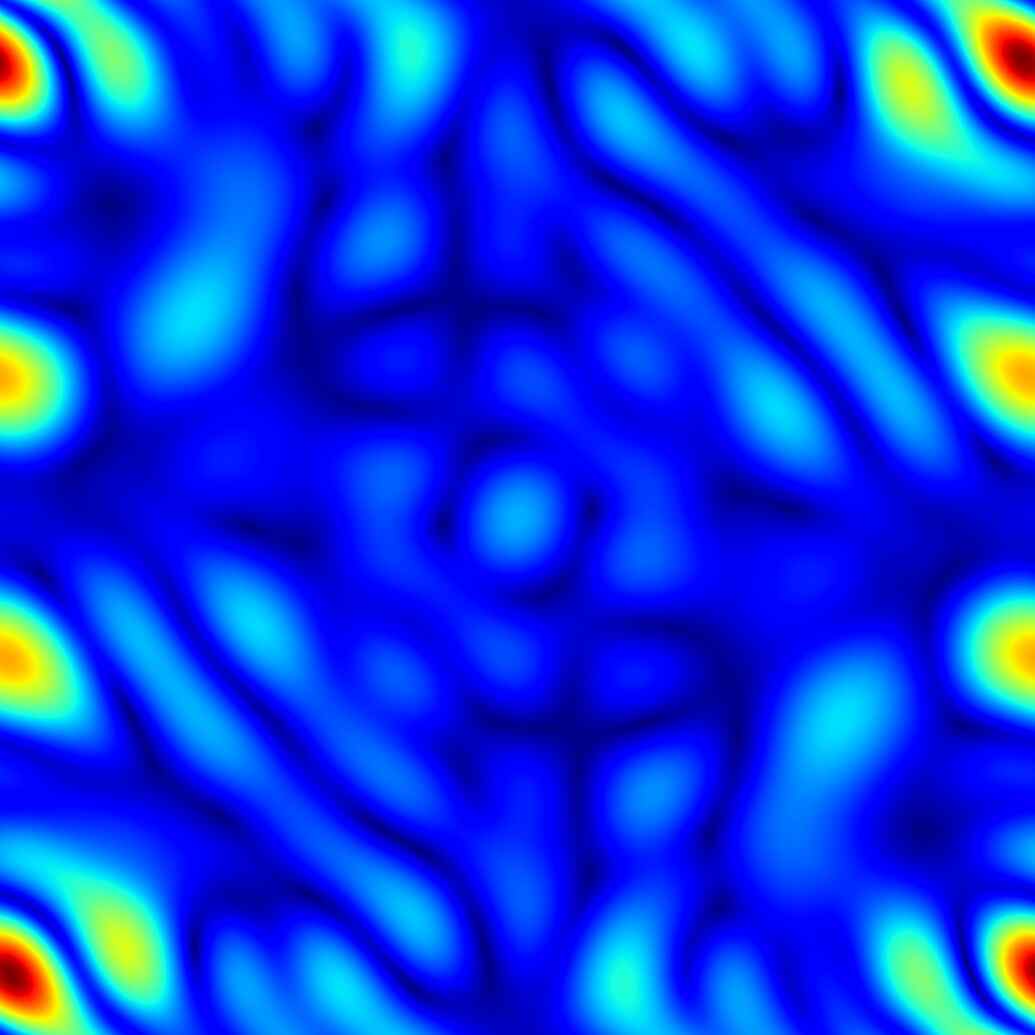}}
        & \fbox{\includegraphics[width=0.089\textwidth]{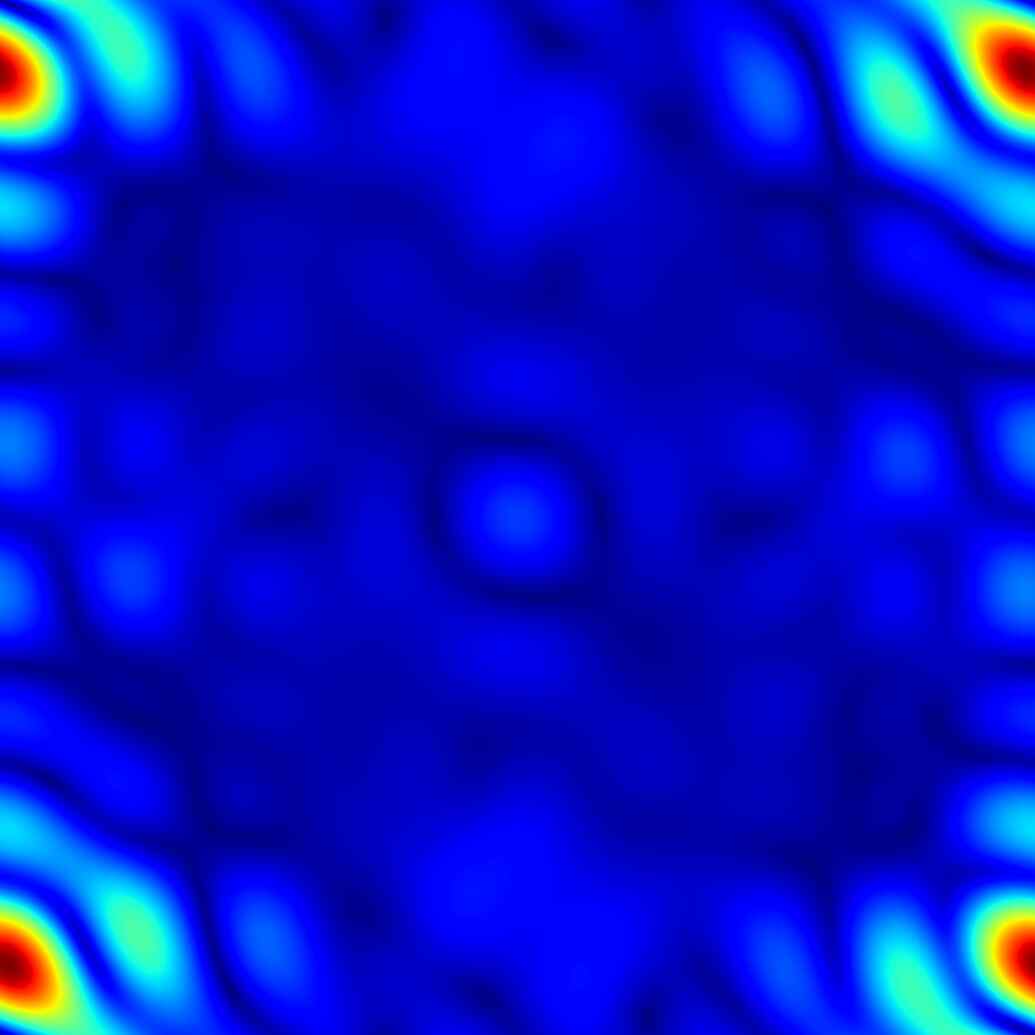}}
        & \fbox{\includegraphics[width=0.089\textwidth]{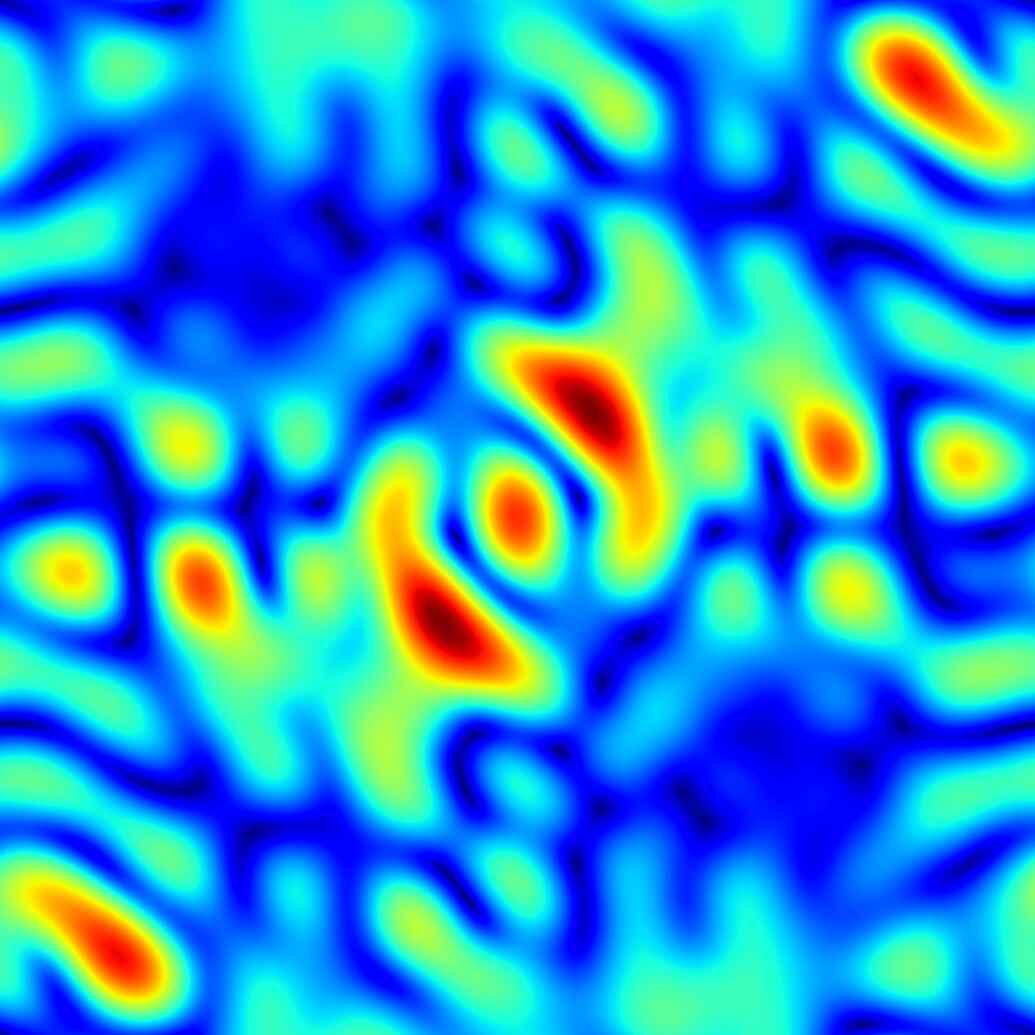}}
        & \fbox{\includegraphics[width=0.089\textwidth]{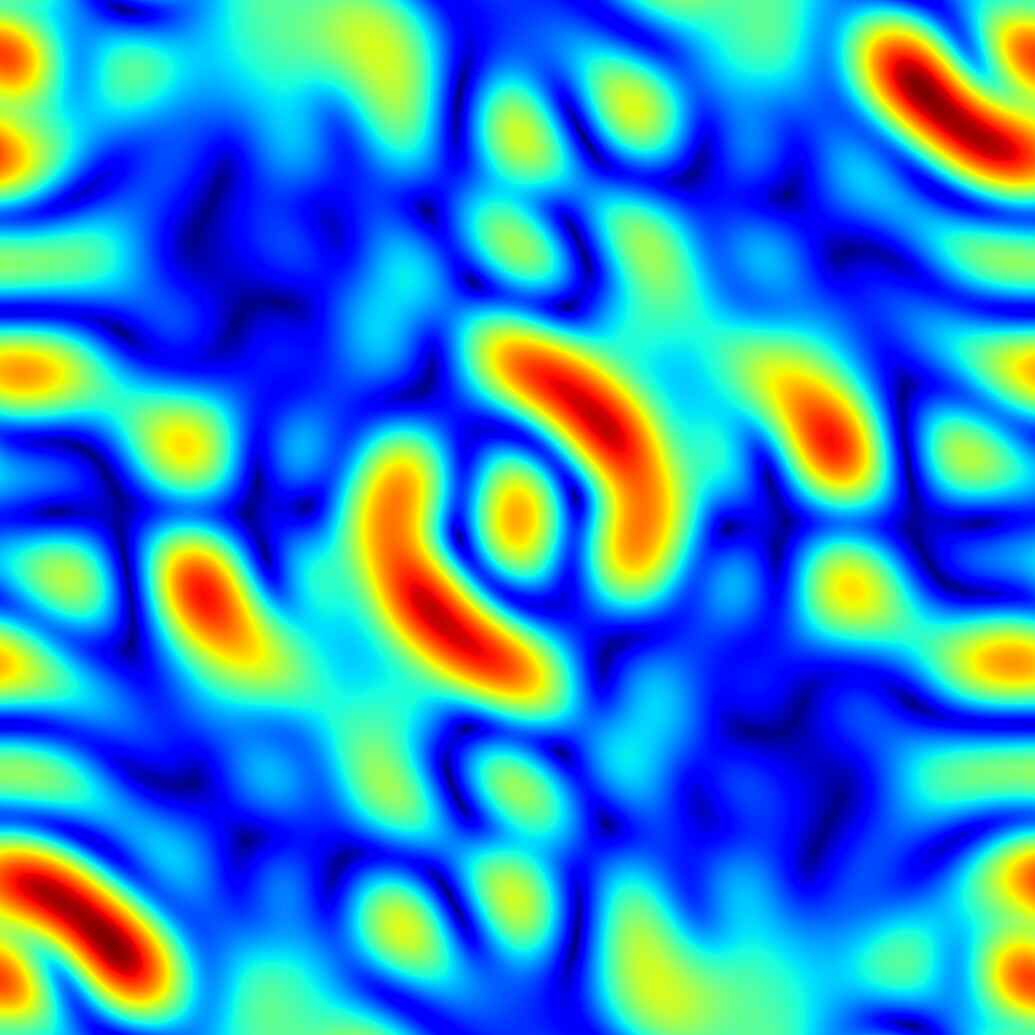}}
        & \fbox{\includegraphics[width=0.089\textwidth]{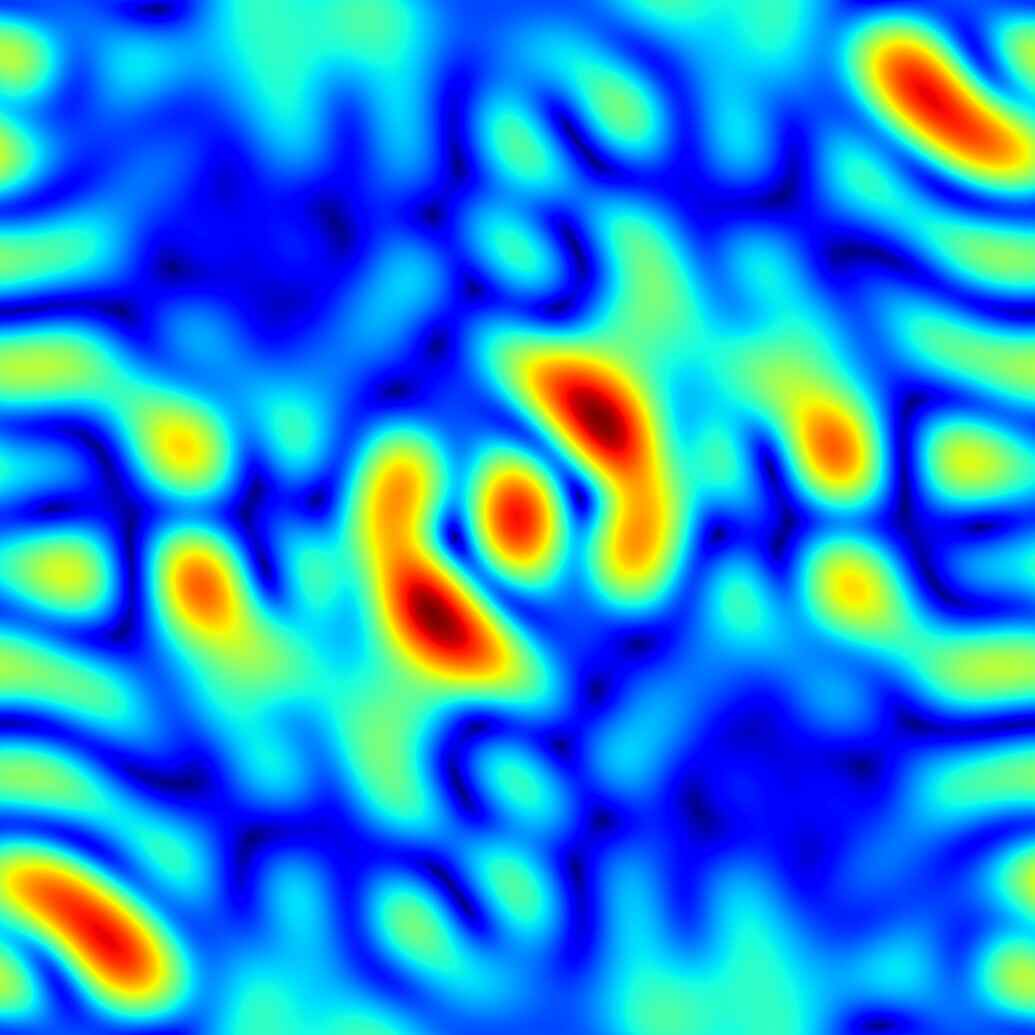}}
        & \fbox{\includegraphics[width=0.089\textwidth]{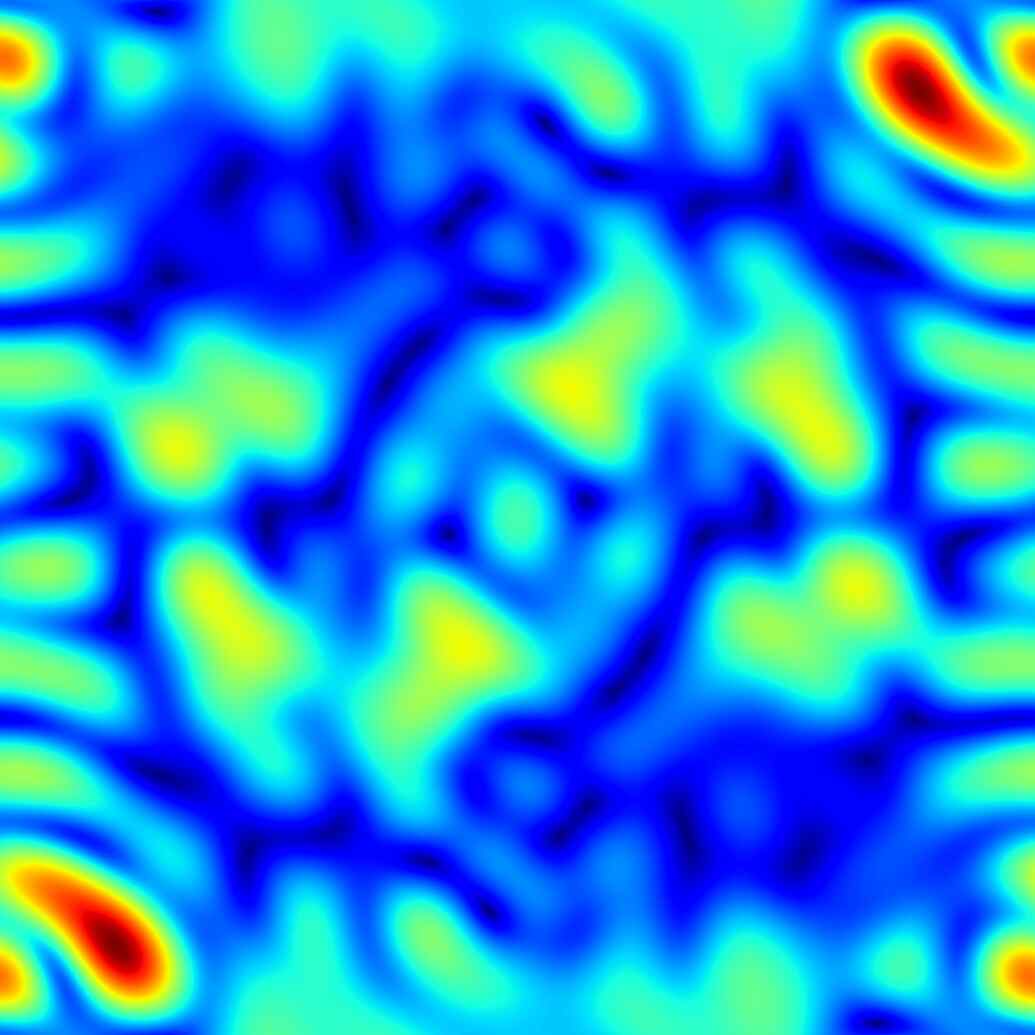}}
        & \fbox{\includegraphics[width=0.089\textwidth]{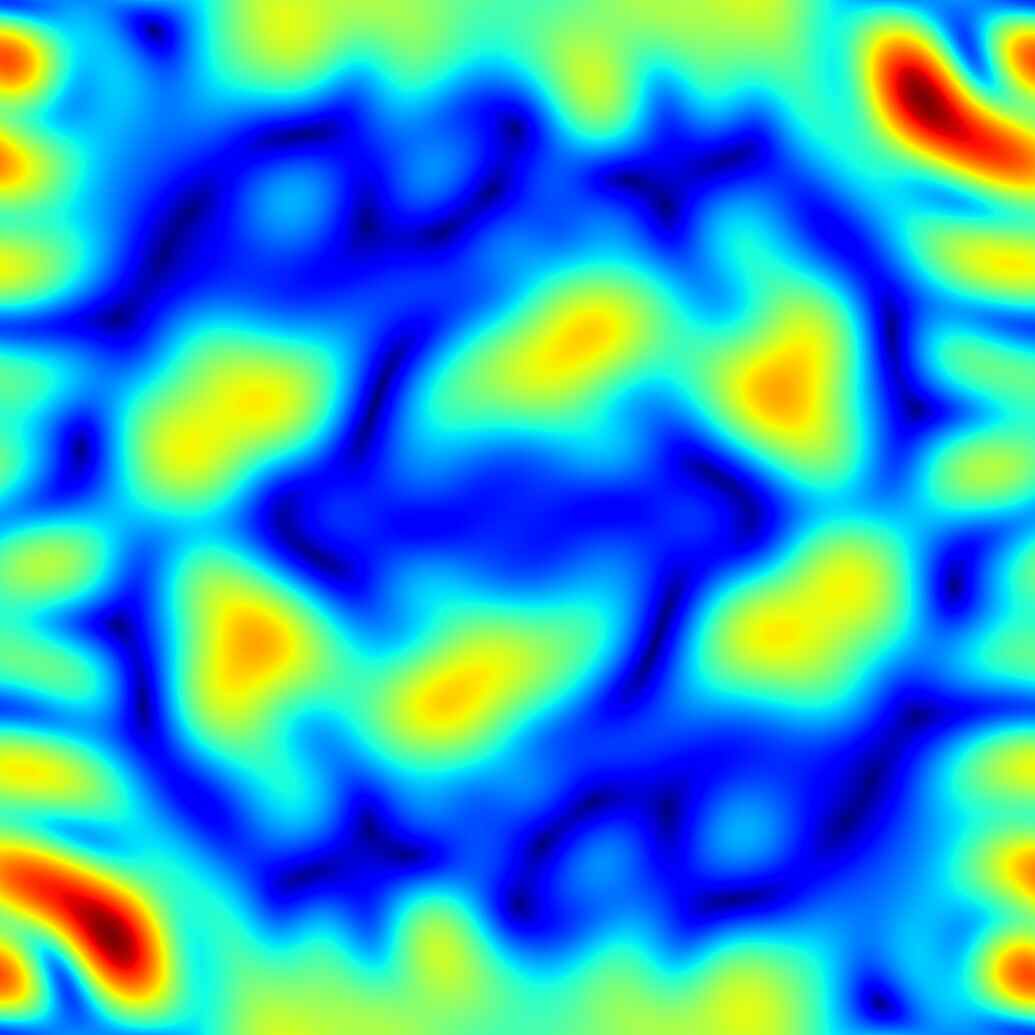}}
        & \fbox{\includegraphics[width=0.089\textwidth]{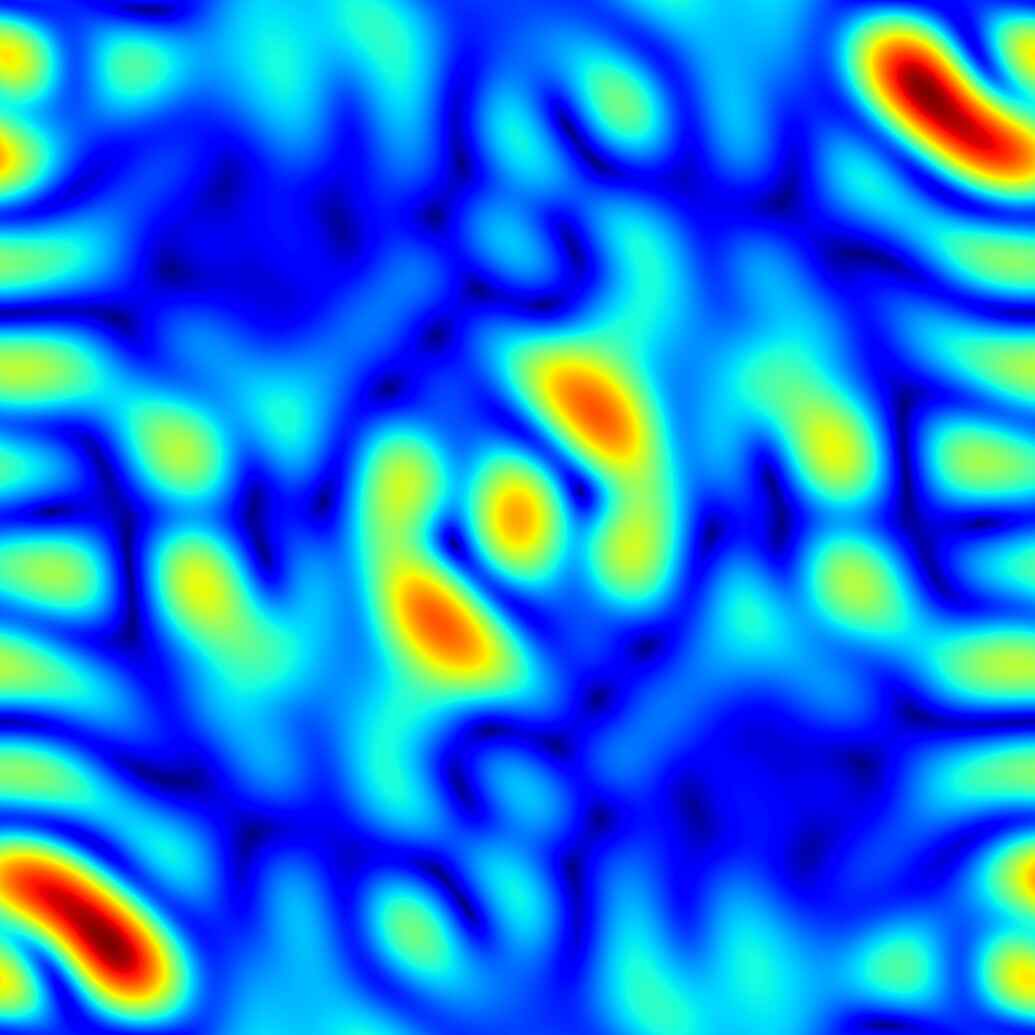}} \\

        \raisebox{19.5pt}{\fontsize{9pt}{10.5pt}\selectfont $FFT(\phi_4)$\hspace{2pt}}
        & \fbox{\includegraphics[width=0.089\textwidth]{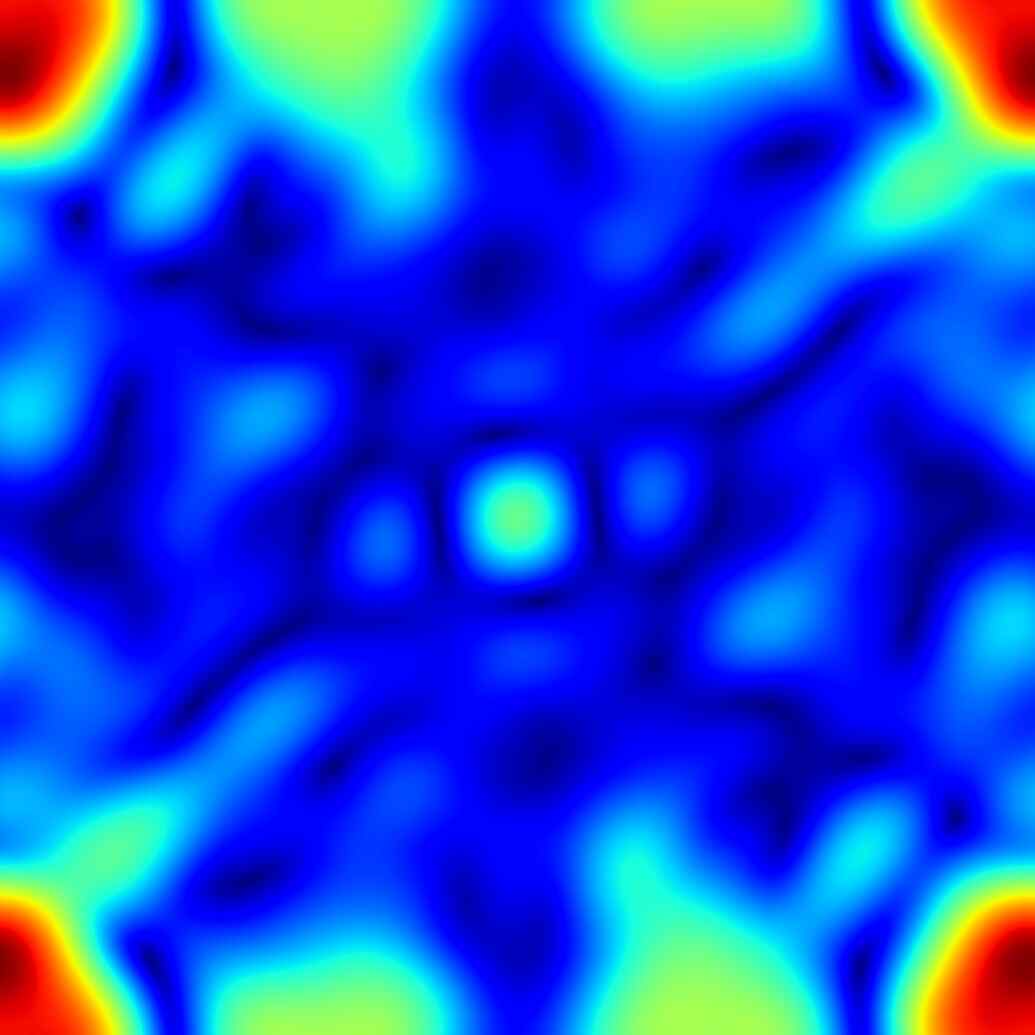}}
        & \fbox{\includegraphics[width=0.089\textwidth]{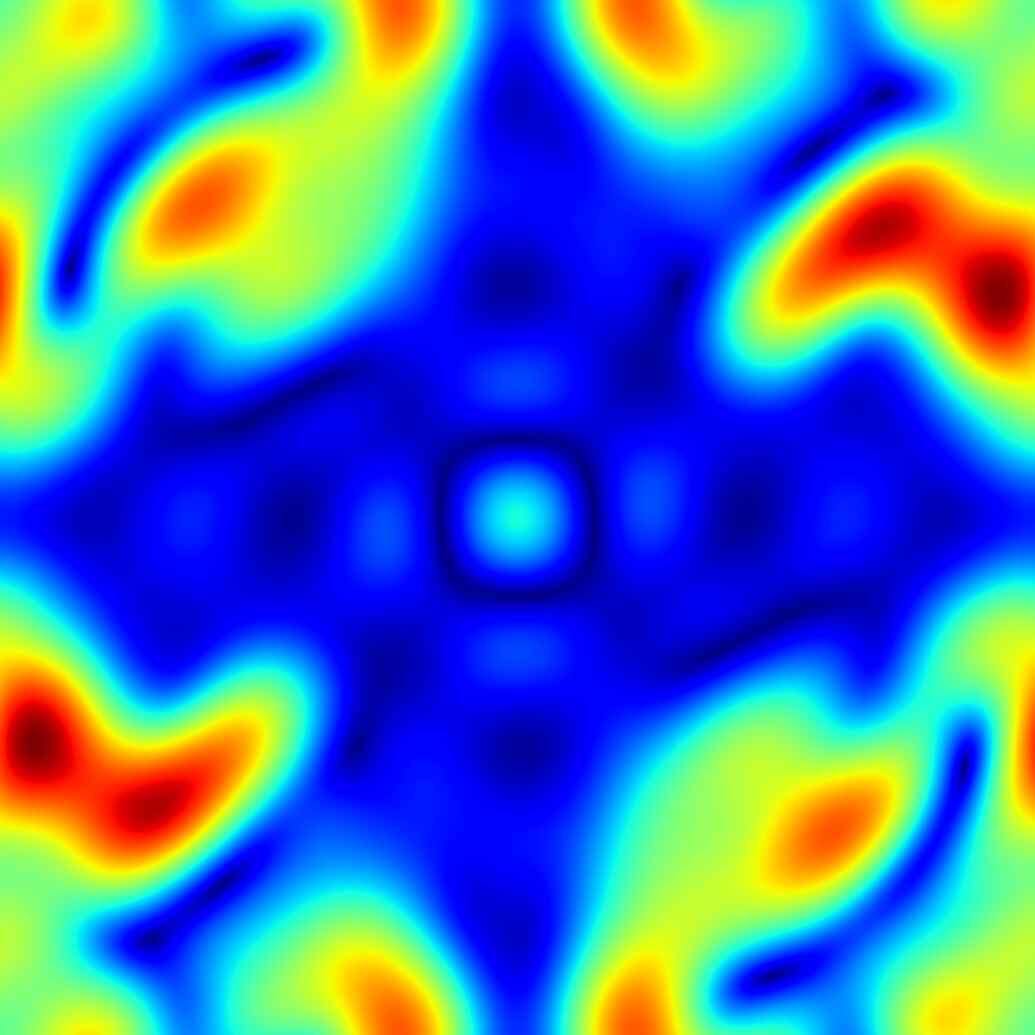}}
        & \fbox{\includegraphics[width=0.089\textwidth]{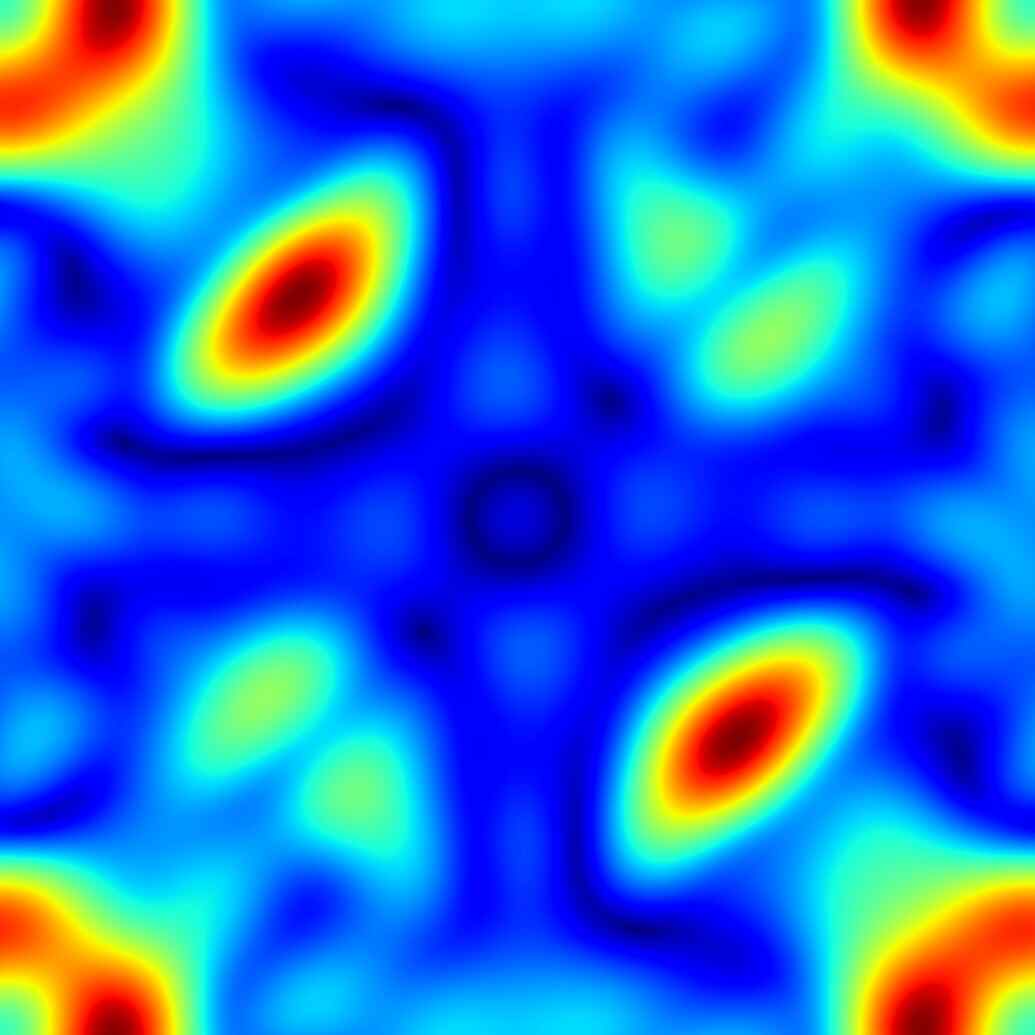}}
        & \fbox{\includegraphics[width=0.089\textwidth]{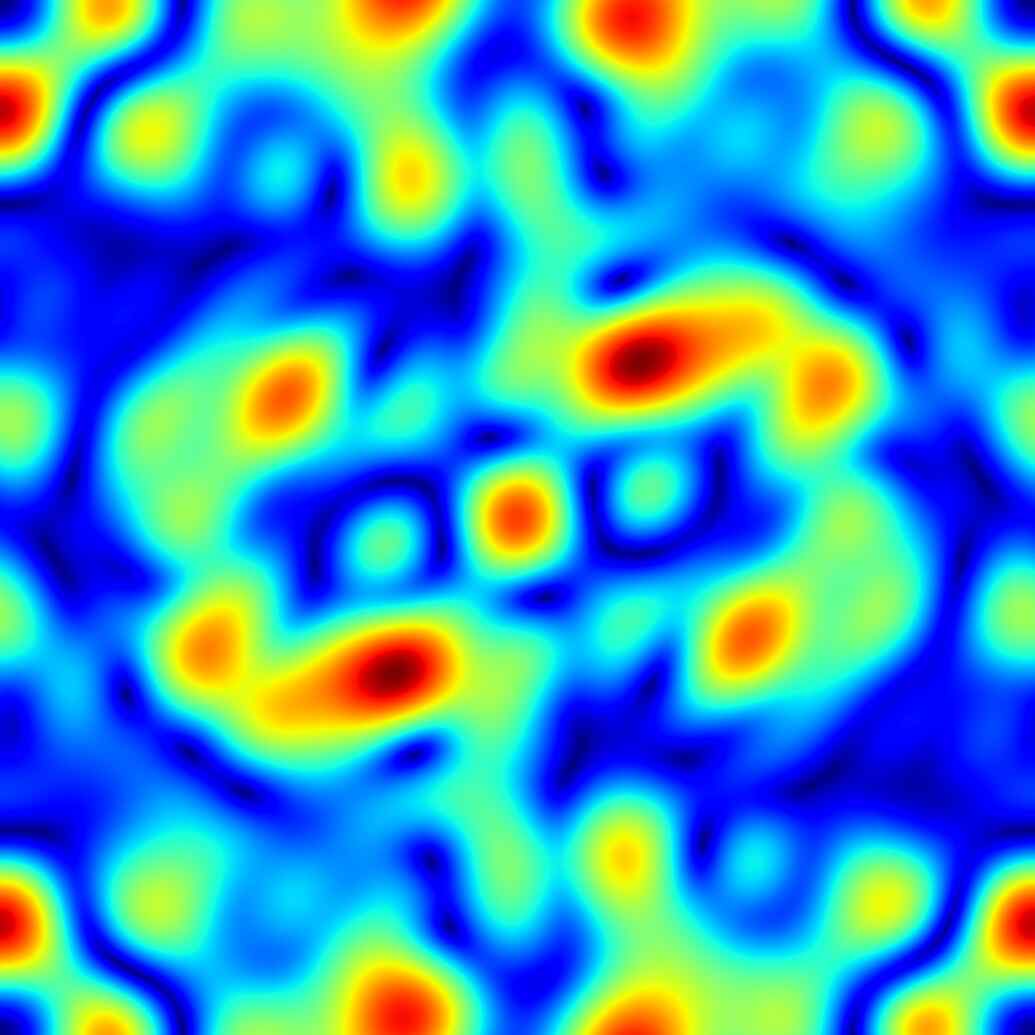}}
        & \fbox{\includegraphics[width=0.089\textwidth]{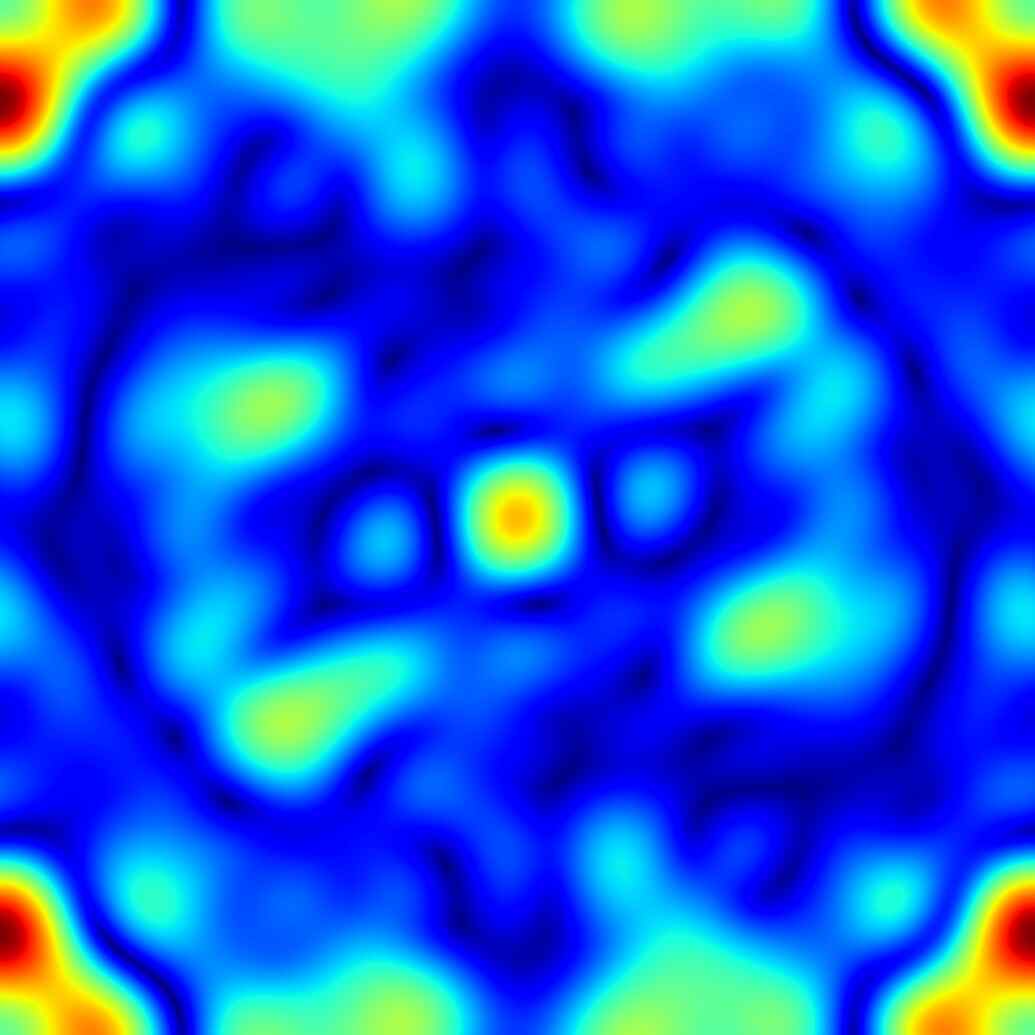}}
        & \fbox{\includegraphics[width=0.089\textwidth]{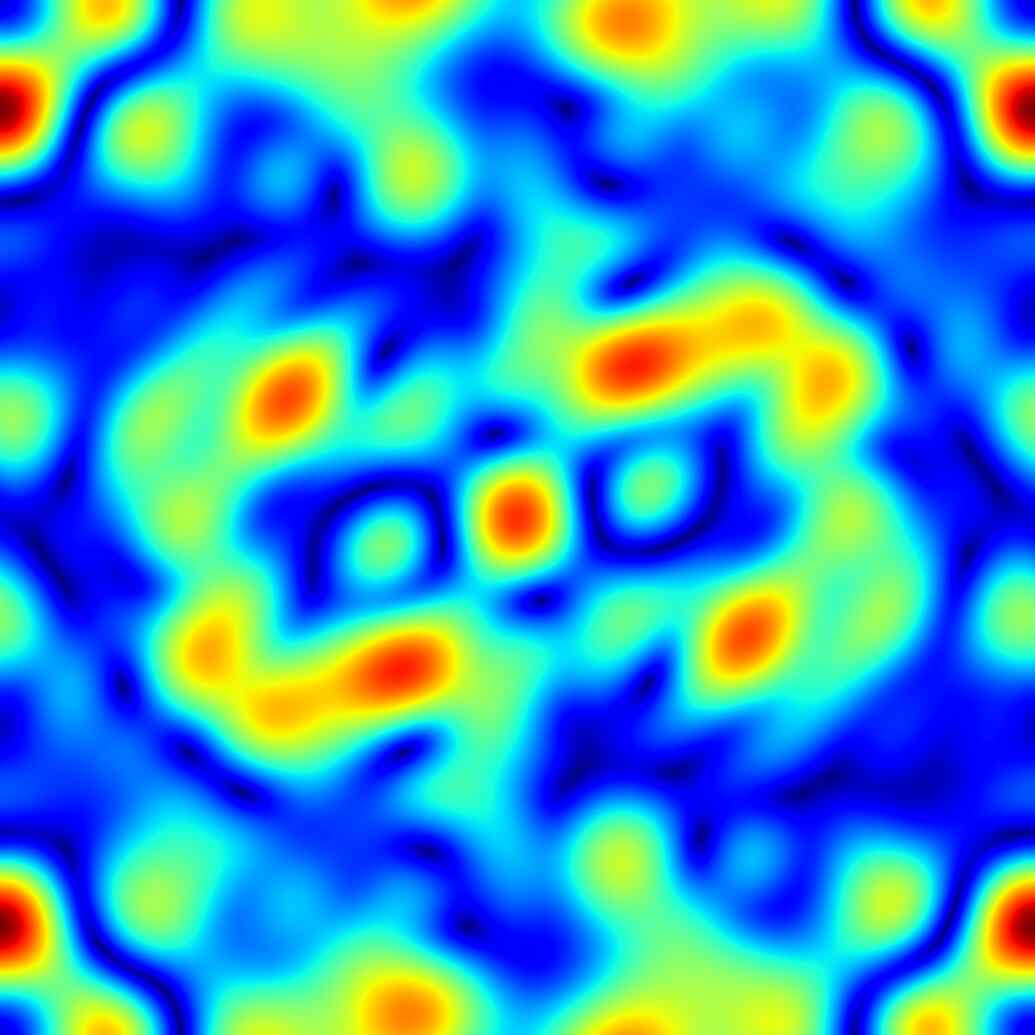}}
        & \fbox{\includegraphics[width=0.089\textwidth]{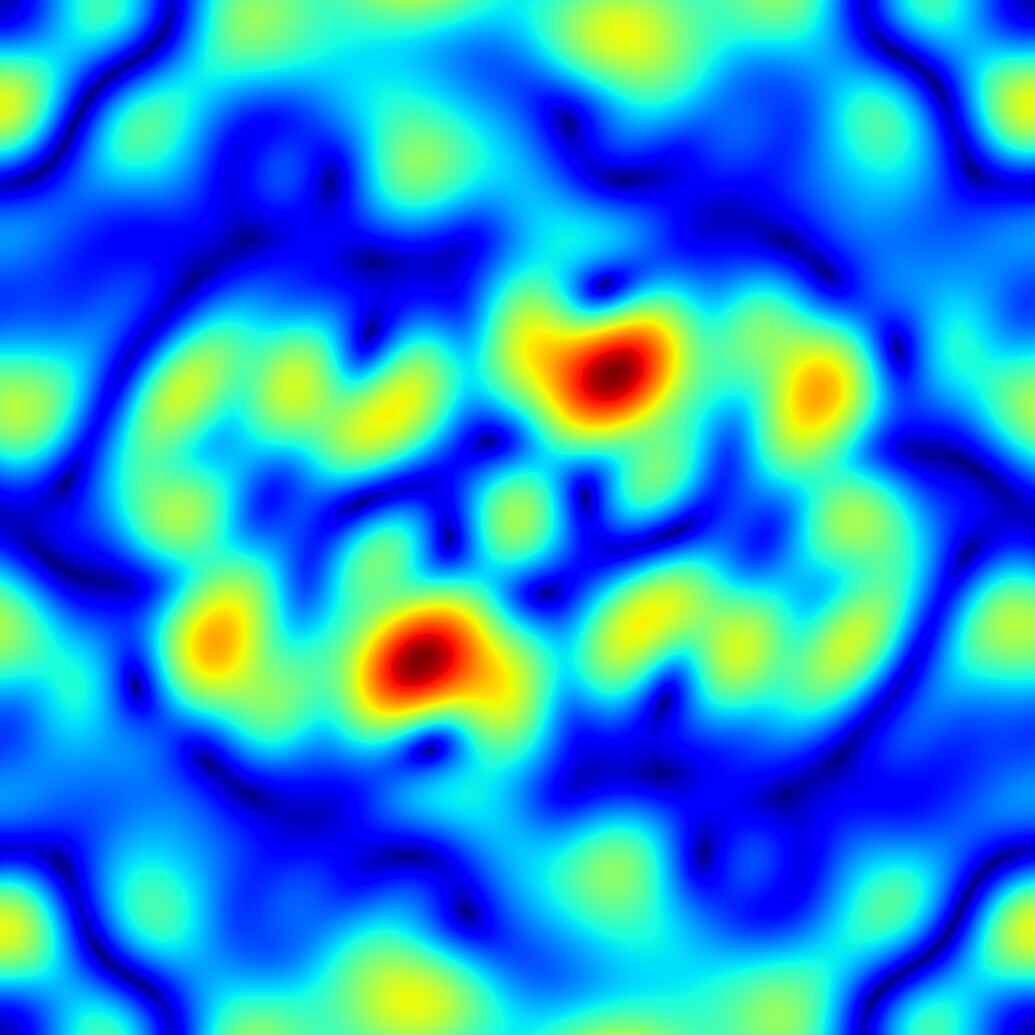}}
        & \fbox{\includegraphics[width=0.089\textwidth]{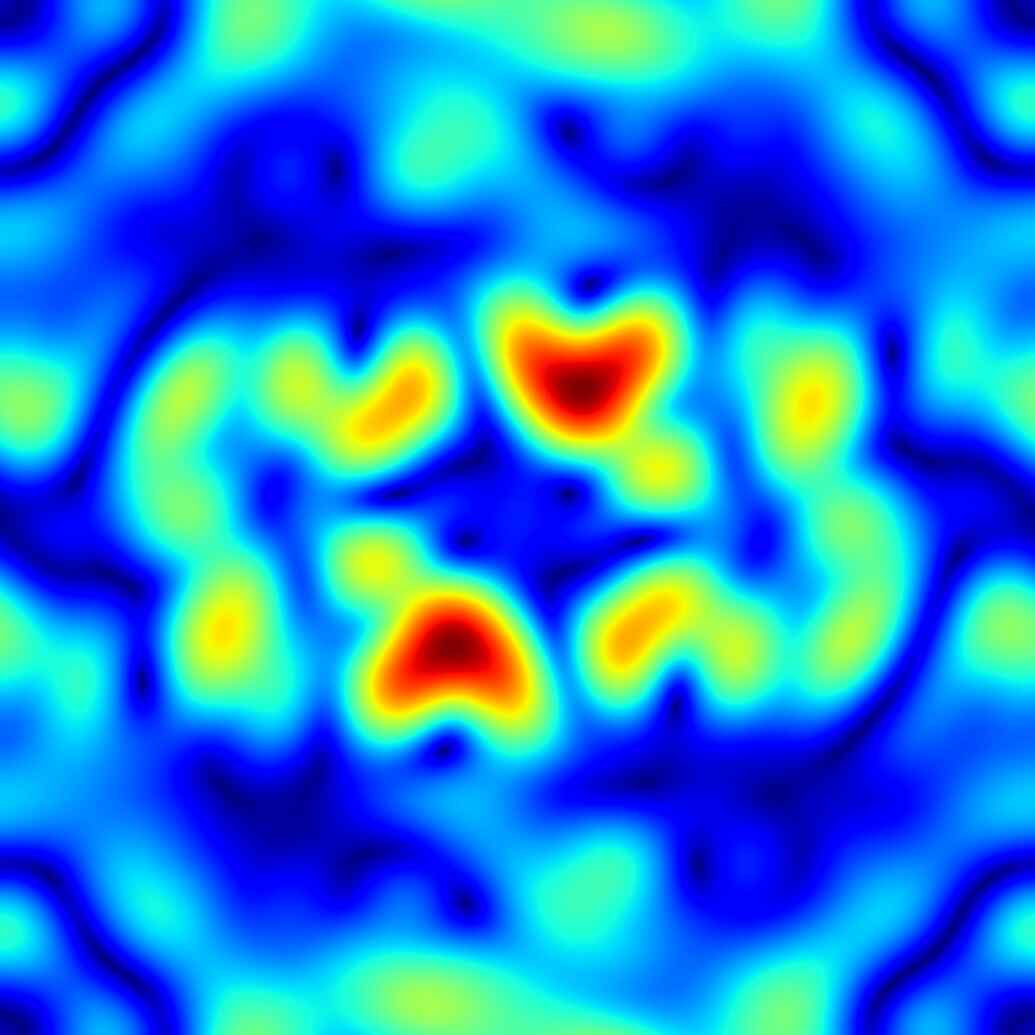}}
        & \fbox{\includegraphics[width=0.089\textwidth]{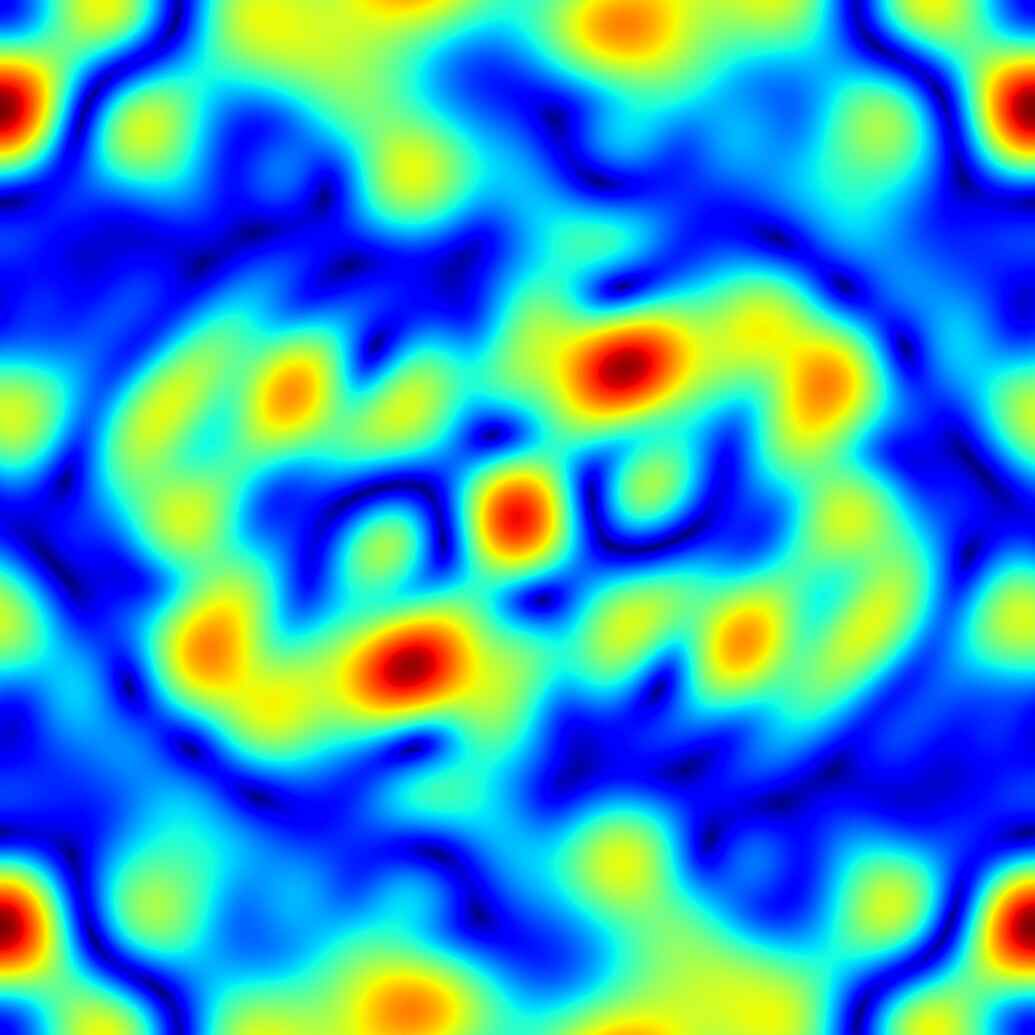}} \\

    \end{tabular}
    }
	\pullupp
    \caption{Visualization of the average power spectrum of different filters in the forensic self-descriptions obtained from various sources.}
    \label{fig:self_desc_viz}
    \pulluppp\pullupp
\end{figure*}

%% file: sections/ProposedMethod_v0.tex
\section{Proposed Method}
\label{sec:proposed}

In this paper, we propose a novel approach for detecting and attributing synthetic images without any exposure to them. As illustrated in Fig.~\ref{fig:system_diagram}, we first learn a set of diverse predictive filters using only real images to approximate scene content. We then apply these filters and extract residuals containing forensic microstructures from a single image. Finally, we jointly model these residuals across multiple scales with a parametric model to derive a unique forensic self-description for each image. This self-description captures intrinsic forensic properties, enabling precise distinction of image sources. More details are presented below.

\input{sections/ExtractForensicResiduals_v2}

\input{sections/ForensicSelfDescription_v0}

%% file: sections/ExtractForensicResiduals_v2.tex
\subsection{Forensic Microstructures Extraction}
\label{sec:method_forensic_residuals}

Prior research has shown that the process used to form an image leaves behind unique forensic microstructures~\cite{marra2019gans}. This holds true for both cameras and AI image generators~\cite{lukas2006digital, ChangWIFS2019}.
While a common strategy to identify synthetic images is to utilize the differences in these microstructures~\cite{LGrad, NPR}, they are not directly observable. However, we can estimate them using the procedure below.

We begin by modeling an image $I$ as the sum of two independent components: the scene content $S$ and the forensic microstructures $\Psi$, such that:
\skipless
\begin{equation}
	I(x,y) = S(x,y) + \Psi(x,y),
\end{equation}
where $(x, y)$ are the 2D pixel coordinates.

Using this model, we can estimate $\Psi$ by approximating $S$ and subtracting $\hat{S}$ from $I$. This subtraction results in a residual which contains forensic microstructures and estimation noise $\epsilon$. In practice, however, it is challenging to perfectly approximate the scene content, which means the estimate of the microstructures will be imperfect.

To address this problem, we use a series of $K$ distinct scene predictions to produce a set of unique residuals $\{r_k\}_{k=1}^K$, such that:
\begin{equation}
	\label{eq:residual}
	r_k(x, y) = I(x, y) - \hat{S}_k(x, y) = \Psi_k(x, y) + \epsilon_k(x, y).
\end{equation}
Since each residual captures a different aspect of the microstructures, the collection of these residuals fully describes the microstructures present.

To produce scene content estimates, we use a series of $K$ learnable linear predictive filters $\mathbf{w} = \{w_k\}_{k=1}^K$ that predict the value of each pixel based on its surrounding neighborhood, such that:
\begin{equation}
	\hat{S}_k(x,y) = \sum_{(i,j) \in \mathcal{M}} w_k(i,j) \cdot I(x + i, y + j),
\end{equation}
where $\mathcal{M}$ is the set of offsets in the $M \times M$ neighborhood around $(x,y)$ excluding $(0, 0)$. We implement these filters by constraining a convolutional layer such that the center kernel weight is always set to $0$ and the sum of all kernel weights is $1$ to preserve the energy of the output prediction.

To learn $\mathbf{w}$, we minimize the total energy across all residuals, which results in the following loss term $\mathcal{L}_E$:
\begin{equation}
	\mathcal{L}_E(\mathbf{w}) = \sum_{k=1}^K \sum_{x,y} \left( I(x,y) - \hat{S}_k(x,y) \right)^2.
\end{equation}
However, this loss term alone may produce filters that are redundant. To prevent this, we introduce a novel spectral diversity regularization term that encourages the filters to be as linearly independent as possible, maximizing the diversity of information captured.

To do this, we first construct a matrix $\mathbf{W} \in \mathbb{R}^{K \times (M^2)}$ by reorienting the weights of each filter into a vector:
\begin{equation}
	\mathbf{W} =
	\begin{bmatrix}
		\mathrm{vec}(w_1)^\top \\
		\mathrm{vec}(w_2)^\top \\
		\vdots \\
		\mathrm{vec}(w_K)^\top
	\end{bmatrix}
\end{equation}
We then perform the singular value decomposition on $\mathbf{W}$ to obtain the set of singular values $\left\{ \sigma_i \right\}$. Finally, the spectral diversity regularization term is defined as:
\begin{equation}
	\mathcal{L}_{\text{diversity}}(\mathbf{w}) = - \sum_{i=1}^{\min(K, M^2)} \log(\sigma_i + \alpha),
\end{equation}
where $\alpha$ is a small constant to prevent numerical instability. This term penalizes filter configurations where singular values are small, which would indicate greater degrees of linear dependence among filters. By minimizing $\mathcal{L}_{\text{diversity}}$, we encourage the filters to be as diverse as possible.

We combine the two terms to obtain the overall objective for learning the predictive filters:
\begin{equation}
	\mathbf{w}^* = \arg\min_{\mathbf{w}} \left[ \mathcal{L}_E(\mathbf{w}) + \lambda \mathcal{L}_{\text{diversity}}(\mathbf{w}) \right],
\end{equation}
where $\lambda$ is a hyperparameter that balances the two terms.
%We note that $\mathbf{w}$ is trained on a corpus of only real images.
We note that $\mathbf{w}$ is learned from a training set consisting of only real images.

%% file: sections/ForensicSelfDescription_v0.tex
\subsection{Forensic Self-Description}

%After $\mathbf{w}$ is pretrained on real images, we use it to extract a set of residuals $\{r_k\}_{k=1}^K$ for a single image, irrespective of whether the image is real or synthetic.
%Since these residuals contain both forensic microstructures and noise,
%we extract the structures present in these residuals by building a parametric model of them. When this model converges, then its parameters define \textbf{forensic self-description} of an image.

%After pretraining $\mathbf{w}$ on real images, we apply it to any image—real or synthetic—to extract . These residuals contain both forensic microstructures and noise. To extract the structured patterns within the residuals, we build a parametric model to represent them. When this model converges, its parameters define forensic self-description of an image.

After $\mathbf{w}$ is learned, we use it to extract a set of residuals $\{ r_k \}_{k=1}^K$ for a single image, irrespective of whether the image is real or synthetic.
To capture structures present in these residuals, we build a parametric model of these residuals and use its parameters to describe the forensic microstructures. We refer to these parameters as the \textbf{forensic self-description} of an image.

To do this, we model the $k$-th residual $r_k(x, y)$ on the basis of residual values in a $B \times B$ neighborhood around $(x, y)$, similar to an autoregressive model. Additionally, to capture structures present across different scales, we define the residual at scale $l$ as:
\begin{equation}
    r_k^{(l)} = \text{Downsample}\left( r_k, 2^{l-1} \right),
\end{equation}
where $\text{Downsample}(X, Y)$ reduces the spatial resolution of the input $X$ by a factor of $Y$.

Then, the model of the residuals at scale $l$ is defined as:
\begin{equation}
    \hat{r}_k^{(l)} = \sum_{(m, n) \in \mathcal{B}} \phi_k(m, n) \cdot r_k^{(l)}(x + m, y + n),
\end{equation}
where $\phi_k$ are the parameters of a linear convolutional filter that models $r_k$ at scale $l$, and $\mathcal{B}$ is the set of offsets in the $B \times B$ neighborhood excluding $(0, 0)$.

Although we model each residual \( r_k^{(l)} \) separately with its own filter \( \phi_k \), we optimize all filters \( \{ \phi_k \} \) jointly across all residuals and scales. This joint optimization ensures that the filters collectively capture the interdependent forensic microstructures present in the image.

Hence, the collection of all filters in the model $\Phi = \left\{ \phi_1, \phi_2, \ldots, \phi_K \right\}$ corresponds to an image's forensic self-description.

To learn $\Phi$ jointly across all scales, we first define the total model error at a location $(x, y)$ as:
\begin{equation}
    \varepsilon(x, y) = \sum_{k=1}^K \sum_{l=1}^L r_k^{(l)}(x, y) - \hat{r}_k^{(l)}(x, y).
\end{equation}
Then, we optimize the parameters $\Phi$ by minimizing the total model error power across all locations in the image:
\begin{equation}
    \Phi^* = \arg\min_{\Phi} \sum_{x}\sum_{y} \left|\varepsilon(x, y)\right|^2.
\end{equation}
The final parameter set $\Phi$ constitutes the forensic self-description of the image.

%% file: sections/Applications_v0.tex
\section{Applications of Forensic Self-Description}

Forensic self-descriptions can be used to perform a number of critical tasks related to synthetic image detection and source attribution, such as: zero-shot detection, open-set source attribution, and unsupervised clustering.

\input{sections/ApplicationZeroShot_v1}

\input{tables/ZeroShot_v1}

\input{sections/ApplicationOpenSet_v1}

\input{sections/ApplicationClustering_v1}

%% file: sections/ApplicationZeroShot_v1.tex
\subsection{Zero-Shot Synthetic Image Detection}
\label{sec:application_zeroshot}

Zero-shot detection refers to the task of determining whether an image is real or AI-generated without prior exposure to images from the generator in question. Supervised detectors struggle in this task as they typically learn representations optimized to discriminate between known sources during training.

We can perform zero-shot detection using forensic self-descriptions because they capture all aspects of the forensic microstructures in an image, not just features discriminative among known sources. By modeling the distribution of forensic self-descriptions from real images, we can flag images whose self-descriptions deviate from this distribution.
This ability is qualitatively demonstrated in Fig.~\ref{fig:self_desc_viz}, which shows the power spectra of forensic self-description filters learned from images of different sources. The figure reveals substantial differences between the self-descriptions of real images and those of AI-generated images.

%We can qualitatively see that this is possible by examining the forensic self-description visualizations in Fig.~\ref{fig:self_desc_viz}.  This figure shows the power spectra of forensic self-description filters learned from images of different sources.  From this figure, we can clearly see substantial differences between the self-description of real images and all AI-generated images.

%In this paper, we implement this approach by
%training a Gaussian Mixture Model (GMM)~\cite{GMM} on only the self-descriptions of real images.
We perform zero-shot detection by first using a Gaussian Mixture Model (GMM)~\cite{GMM} to model the distribution of the self-descriptions obtained from a set of real images.
%Then, the likelihood of any image being real is:
Detection is performed by computing the likelihood that an image is real, defined as:
$p(\Phi | \text{Real}) = \sum_{\ell}\pi_\ell \mathcal{N}(\boldsymbol{\mu}\ell, \boldsymbol{\Sigma}\ell)$, where $\Phi$ is the self-description of the image, and $\pi_\ell$, $\boldsymbol{\mu}\ell$, $\boldsymbol{\Sigma}\ell$ are the GMM's  parameters.
If $p(\Phi | \text{Real}) \geq \tau_{\text{real}}$, the image is classified as real; otherwise, it is flagged as synthetic.

%% file: tables/ZeroShot_v1.tex
\begin{table}[!t]
	\centering
	\caption{\label{tab:zero_shot} Zero-shot synthetic image detection performance, measured in average AUC over all pairs of a real dataset vs each synthetic generator source.}
	\pullup
	\resizebox{!}{1.76cm}{
		\begin{tblr}{
				width = \linewidth,
				colspec = {|m{22mm}|m{14mm}|m{13mm}|m{13mm}|m{12mm}|m{12mm}|},
				column{6} = {c},
				cell{1}{2-6} = {c},
				cell{2-10}{2-5} = {c},
				vlines,
				hline{1,2,10-11} = {-}{},
				hline{1} = {2-5}{},
			}
			\textbf{Method} & \textbf{COCO17} & \textbf{IN-1k} & \textbf{IN-22k} & \textbf{MIDB} & \textbf{Average} \\
			CNNDet~\cite{CNNDet} & 0.756 & 0.714 & 0.733 & 0.683 & 0.722\\
			PatchFor~\cite{PatchFor} & 0.833 & 0.823 & 0.845 & 0.790 & 0.823\\
			UFD~\cite{UFD} & 0.903 & 0.862 & 0.815 & 0.612 & 0.798\\
			LGrad~\cite{LGrad} & 0.819 & 0.770 & 0.866 & 0.824 & 0.820\\
			DE-FAKE~\cite{DE-Fake} & 0.765 & 0.749 & 0.617 & 0.791 & 0.731\\
			Aeroblade~\cite{Aeroblade} & 0.728 & 0.741 & 0.582 & 0.646 & 0.674\\
			ZED~\cite{Zed} & 0.751 & 0.676 & 0.716 & 0.747 & 0.723\\
			NPR~\cite{NPR} & 0.945 & 0.900 & 0.900 & 0.957 & 0.926\\
			\textbf{Ours} & \textbf{0.968} & \textbf{0.962} & \textbf{0.941} & \textbf{0.971} & \textbf{0.960}
		\end{tblr}
	}
    \pulluppp
\end{table}

%% file: sections/ApplicationOpenSet_v1.tex
\subsection{Open-Set Synthetic Image Source Attribution}
\label{sec:application_openset}

%Open-set source attribution refers to the task in which one needs to assigning an image to its source, and simultaneously identify if an image comes from an unknown source.
Open-set source attribution refers to the task of identifying the source of an image amongst a set of known source generators, or determining if the image originates from an unknown source.

We can leverage forensic self-descriptions to perform this task as images from common sources share similar forensic microstructures, while those from different sources do not~\cite{corvi2023detection}. To accomplish this, we can model the distribution of forensic self-descriptions from each source separately. Then, we can attribute an image by assigning it to the most likely source. If this likelihood is sufficiently low, we designate the source to be unknown.

We perform open-set source attribution by first collecting a set of images from known sources. Then, for images from source $S$, we model the distribution of their corresponding self-descriptions using a GMM as follows: $p(\Phi | S) = \sum_{\ell}\pi_\ell \mathcal{N}(\boldsymbol{\mu}_\ell, \boldsymbol{\Sigma}_\ell)$. This will result in one GMM for each known source. After training the GMMs, we can then use them to attribute the source of an image by computing the likelihood of its embedding under each GMM. The generator source with the highest likelihood is considered to be the candidate source of the image:
\begin{equation}
    S^* = \arg\max_{S} p(\Phi | S).
\end{equation}
If $p(\Phi | S^*) < \tau_{\text{reject}}$, the image's source is unknown, otherwise, the candidate source is accepted.

%% file: sections/ApplicationClustering_v1.tex
\subsection{Unsupervised Clustering of Image Sources}
\label{sec:application_clustering}

In many practical scenarios, we need to identify common sources in an unlabeled image dataset by applying a clustering algorithm on to the features extracted for each image. In these cases, we can also use the forensic self-descriptions of images as their features.

Particularly, in this paper, we show that we can successfully apply K-means~\cite{kmeans} to the set of forensic self-descriptions produced from individual images to group them based on their description's similarity. The number of clusters can be set based on the expected number of sources or via the elbow method~\cite{elbow_kmeans} or silhouette analysis~\cite{silhouette_kmeans}.

%% file: sections/Results_v0.tex
\pullup
\section{Experiments and Results}

% ZeroShot Table is inside Applications
\input{tables/ZeroShot_v1_min}

\subsection{Implementation Details}

\subheader{Extracting Forensic Residuals} Following Sec.~\ref{sec:method_forensic_residuals}, we trained a scene content approximator with $K=8$ learnable linear predictive filters of neighborhood size $\mathcal{M} = 11 \times 11$ on gray-scaled real images. We used the AdamW optimizer~\cite{AdamW} (learning rate $0.001$) for 10 epochs. A balance factor of $\lambda=1.0$ optimized the two loss terms.

\subheader{Extracting Forensic Self-Descriptions} For each image, we modeled the $K=8$ forensic residuals with $8$ corresponding predictive filters of neighborhood size $\mathcal{B} = 11 \times 11$, across $L=3$ scales (obtained via bilinear downsampling). The filters are optimized over multi-scale residuals using the AdamW optimizer with a learning rate of $0.1$, decaying by half on plateau, for up to 10,000 iterations.

\subsection{Datasets}
\label{sec:datasets}
To conduct our experiments, we pooled together a large composite dataset of real and synthetic images from various publicly available sources. Real images are drawn from: (1) COCO2017~\cite{coco}, (2) ImageNet-1k~\cite{imagenet}, (3), ImageNet-22k~\cite{imagenet22k}, and (4) MISL Image Database (MIDB)~\cite{mislnet, cam-openset}. Synthetic images come from: (1) OSSIA dataset~\cite{Fang_2023_BMVC}, (2) DMID dataset~\cite{corvi2023detection}, and (3) Synthbuster dataset~\cite{Synthbuster}. Overall, our set of synthetic images includes 24 generators across diverse architectures. Some notable ones are:
ProGAN~\cite{progan},
StyleGAN [1 to 3]~\cite{stylegan, stylegan2, stylegan3},
GigaGAN~\cite{Gigagan},
EG3D~\cite{eg3d},
GLIDE~\cite{glide},
Stable Diffusion (SD) [1.3 to 3.0]~\cite{stablediff, stablediffusion3, sdxl},
DALLE [M, 2, 3]~\cite{dallemini, dalle2, dalle3},
Midjourney (MJ) [5, 6]~\cite{Midjourney},
and Adobe Firefly~\cite{AdobeFirefly}.
Data composition details are available in the supplemental materials.

\input{tables/OpenSet_v0}

\input{sections/ResultsZeroShot_v4}

\input{tables/Clustering_v1}

\input{sections/ResultsOpenSet_v4}

\input{figures/TsneClusteringViz}

\input{sections/ResultsClustering_v2}

%% file: tables/ZeroShot_v1_min.tex
\begin{table}[!t]
	\centering
	\caption{\label{tab:zero_shot_min} Worst case zero-shot detection performance across all pairs of a real dataset vs each synthetic generator source. Metrics are reported in AUC.}
	\pullup
	\resizebox{!}{1.76cm}{
		\begin{tblr}{
				width = \linewidth,
				colspec = {|m{21.5mm}|m{5mm} m{14mm}|m{5mm} m{14mm}|m{5mm} m{14mm}|m{5mm} m{14mm}|},
				cell{1}{2} = {c=2}{},
				cell{1}{4} = {c=2}{},
				cell{1}{6} = {c=2}{},
				cell{1}{8} = {c=2}{},
				cell{1}{2-9} = {c},
				cell{2-10}{2-9} = {c},
				hline{1,2,10-11} = {-}{},
			}
			\textbf{Method} & \textbf{COCO17} & & \textbf{IN-1k} & & \textbf{IN-22k} && \textbf{MIDB} \\
			CNNDet~\cite{CNNDet} & 0.477 & {\smaller(DALLE 3)} & 0.424 & {\smaller(DALLE 3)} & 0.439& \smaller{(DALLE3)} & 0.373 & \smaller{(DALLE 3)}\\
			PatchFor~\cite{PatchFor} & 0.547 & \smaller{(SD 2.1)} & 0.543& \smaller{(SD 2.1)} & 0.565& \smaller{(SD2.1)} & 0.536 & \smaller{(SD 2.1)}\\
			UFD~\cite{UFD} & 0.680 & \smaller{(DALLE 3)} & 0.607 & \smaller{(DALLE 3)} & 0.527 & \smaller{(DALLE 3)} & 0.244 & \smaller{(MJ v6)} \\
			LGrad~\cite{LGrad} & 0.617 & \smaller{(SD 2.1)} & 0.625 & \smaller{(Firefly)} & \textbf{0.776} & \smaller{(Firefly)} & 0.606 & \smaller{(SD 2.1)} \\
			DE-FAKE~\cite{DE-Fake} & 0.534 & \smaller{(BigGAN)} & 0.487 & \smaller{(BigGAN)} & 0.383 & \smaller{(BigGAN)} & 0.563 & \smaller{(BigGAN)} \\
			Aeroblade~\cite{Aeroblade} & 0.425 & \smaller{(BigGAN)} & 0.458 & \smaller{(BigGAN)} & 0.336 & \smaller{(BigGAN)} & 0.360 & \smaller{(BigGAN)} \\
			ZED~\cite{Zed} & 0.462 & \smaller{(ProGAN)} & 0.402 & \smaller{(ProGAN)} & 0.375 & \smaller{(ProGAN)} & 0.331 & \smaller{(ProGAN)} \\
			NPR~\cite{NPR} & 0.396 & \smaller{(Firefly)} & 0.239 & \smaller{(Firefly)} & 0.295 & \smaller{(Firefly)} & 0.449 & \smaller{(Firefly)} \\
			\textbf{Ours} & \textbf{0.892} & \smaller{(SD 1.5)} & \textbf{0.903} & \smaller{(GigaGAN)} & 0.714 & \smaller{(GLIDE)} & \textbf{0.896} & \smaller{(MJ v6)}
		\end{tblr}
	}
    \pulluppp\pullup
\end{table}

%% file: tables/OpenSet_v0.tex
\begin{table}[!t]
    \centering
    \caption{\label{tab:open_set}Open-set source attribution performance comparisons with various techniques.}
    \pullup
    \resizebox{!}{2.10cm}{
        \begin{tblr}{
            width = \linewidth,
            colspec = {|m{19mm}|m{24mm}|m{16mm}|m{16mm}|m{17mm}|},
            column{3-5} = {c},
            cell{2}{1} = {r=2}{},
            cell{4}{1} = {r=2}{},
            cell{6}{1} = {r=2}{},
            cell{8}{1} = {r=4}{},
            vlines,
            hline{1-2,4,6,8,12} = {-}{},
            hline{11} = {2-5}{},
        }
        \textbf{Category} & \textbf{Method} & \textbf{Accuracy} & \textbf{AU-CRR} & \textbf{AU-OSCR}\\
        {Transferable\\Embeddings}
        & CLIP~\cite{CLIP} & 0.570 & 0.543 & 0.304\\
        & ResNet-50~\cite{resnet} & 0.538 & 0.605 & 0.372\\
        Supervised
        & DCTCNN~\cite{DCTCNN} & 0.855 & 0.452 & 0.406\\
        & RepMix~\cite{Repmix} & 0.982 & 0.746 & 0.741\\
        {Metric-\\learning}
        & FSM~\cite{FSM} & 0.422 & 0.565 & 0.207\\
        & EXIFNet~\cite{ExifNet} & 0.186 & 0.412 & 0.064\\
        Open-set
        & Abady et al.~\cite{Abady} & 0.828 & 0.640 & 0.555\\
        & POSE~\cite{POSE} & 0.913 & 0.629 & 0.608\\
        & Fang et al.~\cite{Fang_2023_BMVC} & \textbf{0.988} & 0.856 & 0.852\\
        & \textbf{Ours} & 0.964 & \textbf{0.933} & \textbf{0.913}
        \end{tblr}
    }
    \pulluppp
\end{table}

%% file: sections/ResultsZeroShot_v4.tex
\subsection{Zero-Shot Detection Evaluation}
\label{sec:zeroshot_eval}

\subheader{Setup}
To assess zero-shot detection performance, we divided the composite dataset, described in Sec.~\ref{sec:datasets}, into a training set of real images and a test set of both real and synthetic images. We measured performance across 96 real-synthetic dataset pairs and report the average result over all real-vs-synthetic dataset pairs per real source. A detailed breakdown of the results by generator is provided in the supplemental materials.

\subheader{Metrics}
We report the average AUC (Area Under the ROC curve) for direct comparison with prior works.

\subheader{Competing Methods}
We compared our method to 2 traditional approaches: CNNDet~\cite{CNNDet}, PatchFor~\cite{PatchFor}, and 6 state-of-the-art zero-shot methods: LGrad~\cite{LGrad}, UFD~\cite{UFD}, DE-FAKE~\cite{DE-Fake}, Aeroblade~\cite{Aeroblade}, ZED~\cite{Zed}, and NPR~\cite{NPR}.

\subheader{Results}
This experiment's results are provided in Tab.~\ref{tab:zero_shot} and~\ref{tab:zero_shot_min}.
These results show that our method
achieves the highest zero-shot detection performance, with an overall average AUC of 0.960 across all datasets. In contrast, supervised methods like CNNDet and PatchFor obtain lower performance because the features they learned during training do not transfer well to new generators.

While zero-shot methods such as ZED, DE-FAKE, and NPR show strong performance on some generators, they struggle on others.
Tab.~\ref{tab:zero_shot_min} shows the worst-case performance of each method across all real-versus-synthetic dataset pairs.
The table reveals that ZED consistently struggled with detecting ProGAN, DE-FAKE with BigGAN, and NPR with Firefly.
In contrast, by using forensic self-descriptions, we achieve consistently strong performance, with an overall worst-case AUC of 0.89 or greater, substantially higher the other methods. The only exception is IN22k, where we are slightly behind LGrad.
These results show that forensic self-descriptions offer reliable detection capability across a wide range of real and synthetic sources.

%As we can see from this table, ZED struggled with MIDB-vs-ProGAN (0.331), DE-FAKE with IN22k-vs-BigGAN (0.383), and NPR with IN22k-vs-Firefly (0.239).
%, irrespective of the generator or real source.

%% file: tables/Clustering_v1.tex
\begin{table}[!t]
    \centering
    \caption{\label{tab:clustering}Clustering performance comparisons with various techniques. Here, the ground-truth number of sources is $N = 8$.}
    \pullup
    \resizebox{1.0\linewidth}{!}{
        \begin{tblr}{
            width = \linewidth,
            colspec = {|m{21.75mm}|m{6.5mm}|m{8.3mm}|m{6.5mm}|m{6.5mm}|m{8.3mm}|m{6.5mm}|m{6.5mm}|m{8.3mm}|m{6.6mm}|},
            row{2} = {c},
            cell{1}{1} = {r=2}{},
            cell{1}{2} = {c=3}{0.30\linewidth,c},
            cell{1}{5} = {c=3}{0.30\linewidth,c},
            cell{1}{8} = {c=3}{0.30\linewidth,c},
            cell{3-13}{2-10} = {c},
            vline{1-10} = {-}{},
            vline{2,5,8} = {2}{1-13}{solid,black}, % double vertical line
            hline{1,2,3,5,9,13,14} = {-}{},
        }
        \textbf{ Method} & \textbf{\# Clusters = N} &  &  & \textbf{\# Clusters = 2N } &  &  & \textbf{\# Clusters = 4N } &  &  \\
        & {\textbf{Avg.}\\\textbf{Acc.}} & \textbf{Purity} & \textbf{NMI} & {\textbf{Avg.}\\\textbf{Acc.}} & \textbf{Purity} & \textbf{NMI} & {\textbf{Avg.}\\\textbf{Acc.}} & \textbf{Purity} & \textbf{NMI} \\
        CLIP~\cite{CLIP} & 0.68 & 0.68 & 0.60 & 0.72 & 0.72 & 0.59 & 0.73 & 0.74 & 0.52 \\
        ResNet-50~\cite{resnet} & 0.50 & 0.51 & 0.38 & 0.56 & 0.59 & 0.40 & 0.60 & 0.59 & 0.37 \\
        FSM~\cite{FSM} & 0.16 & 0.16 & 0.01 & 0.18 & 0.18 & 0.02 & 0.20 & 0.20 & 0.03 \\
        EXIFNet~\cite{ExifNet} & 0.21 & 0.22 & 0.06 & 0.24 & 0.26 & 0.08 & 0.32 & 0.28 & 0.09 \\
        Abady et al.~\cite{Abady} & 0.45 & 0.40 & 0.30 & 0.46 & 0.46 & 0.30 & 0.51 & 0.48 & 0.28 \\
        POSE~\cite{POSE} & 0.57 & 0.49 & 0.36 & 0.56 & 0.50 & 0.32 & 0.49 & 0.52 & 0.32 \\
        CNNDet~\cite{CNNDet} & 0.47 & 0.36 & 0.28 & 0.49 & 0.38 & 0.27 & 0.52 & 0.42 & 0.26 \\
        NPR~\cite{NPR} & 0.46 & 0.39 & 0.34 & 0.57 & 0.48 & 0.33 & 0.63 & 0.51 & 0.32 \\
        DE-FAKE~\cite{DE-Fake} & 0.32 & 0.25 & 0.16 & 0.24 & 0.25 & 0.14 & 0.22 & 0.25 & 0.12 \\
        UFD~\cite{UFD} & \textbf{0.78} & 0.71 & 0.68 & 0.67 & 0.69 & 0.55 & 0.71 & 0.72 & 0.50 \\
        \textbf{Ours} & \textbf{0.78} & \textbf{0.77} & \textbf{0.69} & \textbf{0.80} & \textbf{0.81} & \textbf{0.65} & \textbf{0.83} & \textbf{0.85} & \textbf{0.61}
        \end{tblr}
    }
    \pulluppp
\end{table}

%% file: sections/ResultsOpenSet_v4.tex
\subsection{Open-Set Source Attribution Evaluation}
\label{sec:openset_eval}

\subheader{Setup}
To evaluate open-set source attribution performance, we selected 9 sources (1 real and 8 synthetic) from our pooled dataset (described in Sec.~\ref{sec:datasets}), dividing them into five known (ImageNet-1k, StyleGAN, StyleGAN3, SD 1.4, ProGAN) and four unknown sources (StyleGAN2, SD 3, DALLE 3, Firefly). Supervised and open-set methods were trained on known sources and tested on both known and unknown sources.

\subheader{Metrics}
Following other open-set works~\cite{neal2018open, POSE, OSCR, Fang_2023_BMVC, advOpenset}, we show (1) the average accuracy across all known sources, and (2) the Area Under the Correct Rejection Rate curve (AU-CRR)~\cite{POSE, Fang_2023_BMVC}, and (3) the Area Under the Open Set Classification Rate curve (AU-OSCR)~\cite{POSE, OSCR}.

\subheader{Competing Methods}
We compared our method against three state-of-the-art methods designed for this task: Abady et al.~\cite{Abady}, Fang et al.~\cite{Fang_2023_BMVC}, POSE~\cite{POSE}; two supervised methods: DCTCNN~\cite{DCTCNN}, and RepMix~\cite{Repmix};
two metric-learning methods designed for image forensics: FSM~\cite{FSM}, EXIFNet~\cite{ExifNet}; and two methods which produce generic visual embeddings: CLIP~\cite{CLIP}, and a ResNet-50~\cite{resnet} trained on ImageNet1k. For methods which only produce a generic embedding, we apply the same open-set procedure proposed in Sec.~\ref{sec:application_openset} to their produced embeddings.

\subheader{Results}
Tab.~\ref{tab:open_set} shows the results of this experiment.
These results show that leveraging forensic self-descriptions leads to the highest AU-CRR (0.933) and AU-OSCR (0.913). We also obtained near-best known source accuracy (0.964), behind Fang et al.’s 0.988 and RepMix's 0.982.
These results indicate that forensic self-descriptions enable both accurate attribution of images to their sources and reliable detection of images from unknown sources.
%This is also demonstrated in Fig.~\ref{fig:self_desc_viz}, where we observe that each source has its own unique spectral properties that differs from others.
This is also qualitatively demonstrated in Fig.~\ref{fig:self_desc_viz}, where we can see that the forensic self-descriptions of each source differ from one another.
%
%Our strong results is due to the fact that forensic self-descriptions capture distinct microstructural patterns for each source, enhancing attribution even when encountering new generators.
%
%In contrast, while  perform well on known sources (0.982 accuracy), their lower AU-CRR (0.746) reflects limited transferability to unknown sources.

Both supervised methods like RepMix and
dedicated open-set methods like POSE, Abady et al., and Fang et al.
achieve moderate to strong known source accuracies but fall short in AU-CRR and AU-OSCR compared to our method. This is because they rely on embedding spaces learned from known generators to generalize to new and unknown generators, which is challenging in practice.
%Notably, while Fang et.al perform well across all metrics, our method still surpasses them on AU-CRR and AU-OSCR.
%This shows that by learning self-descriptions that capture all aspects of the forensic microstructures and not just those useful for discriminating between known sources during training, we can peform accurate open-set attribution of image sources.
In contrast, forensic self-descriptions capture all aspects of forensic microstructures, not just those useful for discriminating between known sources during training.
This enables us to perform accurate open-set attribution of image sources.

%forensic self-descriptions which capture all aspects of forensic microstructures, rather than only those useful for differentiating known sources, is crucial
%for accurate open-set attribution of image sources.

%% file: figures/TsneClusteringViz.tex
\begin{figure}[!t]
    \centering
    \includegraphics[width=0.72\linewidth]{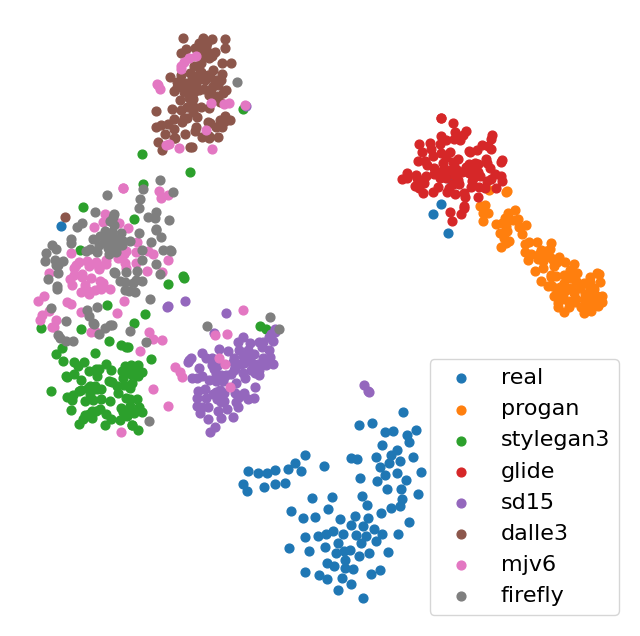}
    \pullupp
    \caption{2D t-SNE plot showing the distribution of the self-descriptions among real and synthetic sources.}
    \label{fig:tsne_clustering}
    \pulluppp\pullupp
\end{figure}

%% file: sections/ResultsClustering_v2.tex
\subsection{Unsupervised Clustering Evaluation}
\label{sec:clustering_eval}

\subheader{Setup}
To evaluate clustering, we used 8 sources representing distinct generation techniques from our composite dataset described in Sec.~\ref{sec:datasets} ~(Real: ImageNet-1k; Synthetic: ProGAN, StyleGAN3, GLIDE, SD 1.5, DALLE 3, MJ v6, Firefly). Our method, applied in an unsupervised manner, does not have training data. For other methods that require synthetic images in their training data, we retrained them on other sources not seen during testing.

\subheader{Metrics}
We present clustering accuracy, purity, and Normalized Mutual Information (NMI), measured across integer multiples of the true number of sources (N, 2N, and 4N) to benchmark performance under different scenarios.

\subheader{Competing Methods}
We evaluated our method against four methods in the zero-shot experiment: NPR~\cite{NPR}, UFD~\cite{UFD}, DE-FAKE~\cite{DE-Fake} \& CNNDet~\cite{CNNDet}, four metric-learning-based methods: FSM~\cite{FSM}, EXIFNet~\cite{ExifNet}, Abady et al.~\cite{Abady} \& POSE~\cite{POSE}, as well as general vision embeddings: CLIP~\cite{CLIP}, and ResNet-50~\cite{resnet} trained on ImageNet1k. For each method, we extracted embeddings from either the specified embedder network or the penultimate layer and applied K-means clustering using Euclidean distance or the method’s provided distance metric.

\subheader{Results}
We present the results in Tab.~\ref{tab:clustering}, which show that clustering based on forensic self-descriptions achieves the highest performance across all metrics and cluster sizes. This is because these descriptions effectively capture forensic microstructures, causing images from the same source to cluster naturally. This behavior is further illustrated in Fig.~\ref{fig:tsne_clustering}, where the t-SNE plot~\cite{tsne} reveals clear separation between real and synthetic images, with each synthetic generator forming a tight, well-defined cluster.

Notably, when the number of clusters equals the number of sources, UFD performs competitively and CLIP shows moderate clustering ability. This is not surprising, as UFD was designed for enhanced source-separability and CLIP was demonstrated in recent works to have promising detection capabilities~\cite{UFD, Cozzolino_2024_CVPR, AmorosoACM}.

In more realistic scenarios where the number of sources is unknown, clustering is often performed with an overestimated number of clusters followed by merging. Under these conditions, our method continues to improve with larger cluster counts, whereas others show modest gains (Abady et al., CLIP) or performance declines (UFD, POSE). This trend highlights the suitability of forensic self-descriptions for accurate, unsupervised source clustering.

%% file: sections/Ablation_v0.tex
\section{Ablation Study}
\label{sec:ablation}

We conducted an ablation study to understand the impact of different design choices on the performance of forensic self-descriptions. To do this, we measured the performance of the zero-shot detection task in terms of average AUC over a subset of real-vs-synthetic dataset pairs (ImageNet-1k versus ProGAN, SDXL, DALLE 3, MJ v6, and Firefly).
We also calculated the relative error reduction (RER) in detection AUC of our method compared to alternative design choices.
%, defined as:
%$\text{RER} = (\text{AUC}_{R} - \text{AUC}_{N})/(1 - \text{AUC}_{N})$,
%where $\text{AUC}_R$ is our method average AUC, and $\text{AUC}_{N}$ is the alternative version's average AUC.
%with $\text{AUC}_R$​ representing the average AUC of our method and $\text{AUC}_N$​ representing the average AUC of the alternative version.
%with $\text{AUC}_R$ representing the average AUC of our method and $\text{AUC}_{N}$ representing the average AUC of the alternative version.
The results are provided in Tab.~\ref{tab:ablation}.

\input{tables/Ablation_v0}

\subheader{Residual Extraction Method}
We examined the detection performance impact of various design choices in the forensic residual extraction process.
Results in Tab.~\ref{tab:ablation} show that our method of learning a set of diverse linear predictive filters from a corpus of real images is essential for optimal performance. Nonetheless, we observe that even with a simple high-pass filter to extract residuals, our forensic self-descriptions still achieve strong performance.

\subheader{Obtaining Self-Descriptions}
We explored different design choices and their impact on obtaining forensic self-descriptions.
Tab.~\ref{tab:ablation}'s results show that using multiple filters to capture underlying structures in the forensic residuals is essential for optimal performance. Additionally, we observe that the self-description extracted from multi-scaled residuals yielded significant performance gains.
Overall, these findings highlight that the combination of multi-scale modeling, an adequate number of learnable filters, and an appropriate neighborhood size is vital for obtaining effective forensic self-descriptions.

\subheader{Utilizing Self-Descriptions}
We analyzed several out-of-distribution detection methods using forensic self-descriptions. This is important because different approaches offer unique trade-offs between space-time complexity, practicality, and performance.
The results in Tab.~\ref{tab:ablation} show that forensic self-descriptions are versatile and can also be used with a One-Class SVM or an Isolation Forest with minimal performance loss.

%% file: tables/Ablation_v0.tex
\begin{table}
    \centering
    \caption{\label{tab:ablation} Zero-shot detection performance of our proposed forensic self-description and its alternative design choices.}
    \pullupp
    \resizebox{0.80\linewidth}{!}{
        \begin{tblr}{
            width = \linewidth,
            colspec = {|m{25mm}|m{42mm}|m{8mm}|m{9mm}|},
            column{3-4} = {c},
            cell{1}{1} = {r=2}{},
            cell{3}{1} = {r=4}{},
            cell{7}{1} = {r=5}{},
            cell{12}{1} = {r=2}{},
            vlines,
            hline{1,3,7,12,14} = {-}{},
            hline{2} = {2-4}{},
        }
        \textbf{Component} & \textbf{Method} & \textbf{AUC} & \textbf{RER\%} \\
        {} & \textbf{Proposed} & \textbf{0.986} & -- \\

        {Residual\\Extraction}
        & 5$\times$5 high-pass filter~\cite{FridrichRichModels, kodovsky2011dangers} & 0.913  & 83.38  \\
        & 3$\times$3 high-pass filter~\cite{FridrichRichModels, kodovsky2011dangers} & 0.955  & 67.70	  \\
        & Neighbor Pixel Relations~\cite{NPR} & 0.952 & 70.22\ \\
        & No spectral diversity & 0.969  & 53.34\ \\

        {Obtaining\\Self-Descriptions}
        & No multi-scale & 0.956  & 67.51\ \\
        & 1 learnable filter & 0.951  & 70.47  \\
        & 4 learnable filters & 0.931  & 79.08  \\
        & 7$\times$7 neighborhood & 0.961 & 63.28  \\
        & 5$\times$5 neighborhood & 0.897 & 85.98  \\

        {Utilizing\\Self-Descriptions}
        & One-Class SVM~\cite{ocsvm} & 0.968  & 55.00 \\
        & Isolation Forest~\cite{liu2008isolation} & 0.968 & 55.00 \\
        \end{tblr}
    }
    \pulluppp\pullup
\end{table}

%% file: sections/Discussion_v0.tex
\section{Discussion}

\subheader{Qualitative Analysis}
To qualitatively analyze the characteristics of the microstructures captured by forensic self-descriptions, we visualize the average power spectrum of each filter, computed from 100 images across various sources. The resulting power spectra are presented in  Fig.~\ref{fig:self_desc_viz}

As shown in Fig.~\ref{fig:self_desc_viz}, the power spectra of all filters in the self-descriptions of real images are significantly distinct from those of synthetic images. Among synthetic sources, each generator exhibits at least one unique spectral characteristic that differ from others. For instance, StyleGAN3 and SD 1.5 have similar spectral responses in filter 1-3 but differ in filter 4.
This property of the forensic self-descriptions is confirmed by our experimental results above and further illustrated in the t-SNE plot in Fig.~\ref{fig:tsne_clustering}. In this plot, we observe the same property: real images cluster distinctly apart from synthetic images, with each synthetic source forming tight, easily distinguishable clusters.

\subheader{JPEG Robustness}
To assess the robustness of forensic self-descriptions to compression at various JPEG quality factors, we evaluated our method's zero-shot detection performance by measuring the average AUC across quality factors ranging from 50 to 100. This was done on the same subset of real-vs-synthetic dataset pairs used in Sec.~\ref{sec:ablation}.

As shown in Tab.~\ref{tab:jpeg_robust}, our method consistently achieves high AUC scores across all JPEG quality factors with an overall average AUC of 0.972. Even at a low quality factor of 60, our method maintains an AUC of 0.972, showing minimal degradation in detection performance. These results show that forensic microstructures of real and synthetic images still remain distinct and detectable even after compression. This demonstrates that forensic self-descriptions are highly robust and suitable for practical use.

\input{tables/JpegRobustness}

\subheader{Limitations and Future Work}
One possible limitation of forensic self-descriptions is their reliance on accurate and diverse forensic residuals, which in turn depend on training the scene content predictors with a high-quality, diverse set of real images. Future work could explore adaptive filter learning to accommodate new data distributions or develop domain-specific filters for targeted forensic tasks. Extending the approach to handle more complex scenarios, such as post-processed or social media–shared images, could further improve its robustness in real-world settings.

%% file: tables/JpegRobustness.tex
\begin{table}[!t]
	\centering
	\caption{\label{tab:jpeg_robust} Average Zero-Shot AUC of our method over different JPEG quality factors.}
	\pullupp
	\resizebox{1.0\linewidth}{!}{
	\begin{tblr}{
			width = \linewidth,
			colspec = {|m{21mm}|m{8mm}|m{7mm}|m{7mm}|m{7mm}|m{7mm}|m{7mm}|m{7mm}|m{9mm}|},
			column{even} = {c},
			column{3} = {c},
			column{5} = {c},
			column{7} = {c},
			column{9} = {c},
			hlines,
			vlines,
		}
		\textbf{JPEG Quality}  & \textbf{None} & \textbf{100}   & \textbf{90}    & \textbf{80}    & \textbf{70}    & \textbf{60}    & \textbf{50}    & \textbf{Avg.} \\
		Our method & 0.986   & 0.968 & 0.963 & 0.960 & 0.979 & 0.972 & 0.979 & \textbf{0.972}
	\end{tblr}
	}
    \pulluppp
\end{table}

%% file: sections/Conclusion_v0.tex
\section{Conclusion}

We introduced forensic self-descriptions as a robust approach for zero-shot detection, open-set attribution, and unsupervised clustering of synthetic images. By using a self-supervised process to extract residuals containing forensic microstructures, our approach constructs a compact, representative model, that accurately distinguishes real from synthetic images, identifies unknown sources, and clusters images by origin without any supervision. Experimental results confirm forensic self-descriptions resilience to compression artifacts and adaptability across diverse generative models, establishing them as a powerful tool for combating the proliferation of AI-generated fake images.

%Experimental results confirm their resilience to compression artifacts and adaptability across diverse generative models, making them a powerful tool for combating the rise of fake AI-generated images.

%% file: sections/X_suppl.tex
\clearpage
\setcounter{page}{1}
\appendix

\def\maketitlesupplementary
{
	\newpage
	\twocolumn[
	\centering
	\Large
	\textbf{\thetitle}\\
	\vspace{0.5em}
	Supplementary Material \\
	\vspace{0.5em}
	\resizebox*{1.0\linewidth}{!}{
		\begin{tblr}{
				colspec = {m{10mm}m{20mm}m{200mm}},
				rowsep = 2pt,    % Slight row separation
				colsep = 10pt,    % Space between columns
				width = \linewidth, % Table width
			}
			\textbf{Page} & \textbf{Appendix} & \textbf{Title} \\
			\pageref{supp:data} & A & Data Composition \\
			\pageref{supp:cat_and_tax} & B & Method Categories and Taxonomy \\
			\pageref{supp:full_zeroshot} & C & Full Zero-Shot Results \\
			\pageref{supp:zeroshot_vs_thresholds} & D & Zero-Shot Performance vs. Thresholds \\
			\pageref{supp:effects_of_diff_reals} & E & Impact of Choice of Real Training Datasets \\
			\pageref{supp:spec_of_diff_reals} & F & Qualitative Study of Forensic Self-Descriptions of Different Real Datasets \\
			\pageref{supp:space_time_analysis} & I & Space-Time Complexity Analysis \\
		\end{tblr}
	}
	\vspace{1.0em}
	] %< twocolumn
}

\maketitlesupplementary

\input{tables/CatAndTax}

\input{sections/suppl/Data_v0}

\input{figures/CrossValDiffReals}

\input{sections/suppl/CategoryTaxonomy_v0}

\input{sections/suppl/FullZeroShotResults_v0}

\input{figures/SpecOfReals}

\input{figures/ZeroShotAccVsThresholds}

\input{sections/suppl/ZeroShotPerfVsThresholds_v0}

\input{sections/suppl/EffectsOfRealDatasets_v0}

\input{sections/suppl/SpectralOfDiffReals_v0}

\input{sections/suppl/SpaceTimeAnalysis_v0}

%% file: tables/CatAndTax.tex
\newcommand{\newcheckmark}{\textrm{\faCheck}}
\newcommand{\newcrossmark}{\textrm{\faTimes}}
\newcommand{\newhalfcirc}{\textrm{\faAdjust}}

\definecolor{Ebb}{rgb}{0.909,0.89,0.89}

\begin{table*}
	\centering
	\caption{\label{tab:cat_and_tax} Categorization of different capabilities, training data requirement, training paradigm, and high-level idea/approach of competing methods and ours. \newcrossmark ~means no ability or achieving poor performance, \newhalfcirc ~means having moderate ability or performance, and \newcheckmark ~means having good to strong ability or performance.}
	\resizebox{\linewidth}{!}{%
		\begin{tblr}{
			width = \linewidth,
			colspec = {m{23mm}m{17mm}m{17mm}m{17mm}m{28mm}m{24mm}m{77mm}},
			row{2} = {c},
			row{4,6,8,10,12,14,16,18,20} = {Ebb},
			column{5} = {c},
			column{6} = {c},
			cell{1}{1} = {r=2}{},
			cell{1}{2} = {c=3}{0.235\linewidth,c},
			cell{1}{5} = {r=2}{},
			cell{1}{6} = {r=2}{},
			cell{1}{7} = {r=2}{c},
			cell{3-20}{2,3,4} = {c},
			vlines,
			hline{1,3,5,11,13,16,18,20-21} = {-}{},
			hline{2} = {2-4}{},
		}
			\textbf{Method}                   & \textbf{Capabilities} &                   &                     & \textbf{Training Data Requirement} & \textbf{Training Paradigm} & \textbf{Idea / Approach}                                                                                                                                                                                                                     \\
			                                  & \textbf{Zero-Shot}    & \textbf{Open-Set} & \textbf{Clustering} &                                    &                            &                                                                                                                                                                                                                                              \\
			CnnDet~\cite{CNNDet}              & \newhalfcirc          & \newcrossmark     & \newcrossmark       & Real + Synthetic                   & Supervised                 & Standard Classifier trained on 1 GAN can generalize to some other GANs                                                                                                                                                                       \\
			PatchFor~\cite{PatchFor}          & \newcheckmark         & \newcrossmark     & \newcrossmark       & Real + Synthetic                   & Supervised                 & Ensemble of Patch-based classifiers trained on low-level artifacts                                                                                                                                                                           \\
			LGrad~\cite{LGrad}                & \newcheckmark         & \newcrossmark     & \newcrossmark       & Real + Synthetic                   & Supervised                 & Classifier trained on 2D gradients of a common CNN as forensic features                                                                                                                                                                                             \\
			UFD~\cite{UFD}                    & \newcheckmark         & \newcrossmark     & \newhalfcirc        & Real + Synthetic                   & Supervised                 & {Classifier trained based on CLIP's embedding distances to real and fake reference embeddings}                                                                                                                                               \\
			DE-FAKE~\cite{DE-Fake}            & \newcheckmark         & \newcrossmark     & \newcrossmark       & Real + Synthetic                   & Supervised                 & Classifier trained based on CLIP's and BLIP's text and visual embeddings                                                                                                                                                                     \\
			Aeroblade~\cite{Aeroblade}        & \newcheckmark         & \newcrossmark     & \newcrossmark       & No Data Required                   & Training-Free              & {The reconstruction errors using pretrained Diffusion models of synthetic images are lower than that of real images}                                                                                                                         \\
			ZED~\cite{Zed}                    & \newcheckmark         & \newcrossmark     & \newcrossmark       & Real                               & Self-Supervised            & {The coding costs using a lossless neural compressor (trained on real images) of real images are lower than that of synthetic images}                                                                                                        \\
			NPR~\cite{NPR}                    & \newcheckmark         & \newcrossmark     & \newcrossmark       & Real + Synthetic                   & Supervised                 & {Classifier trained on neighboring pixel relationships, which is extracted by subtracting the image by its down-up-sampled version}                                                                                                          \\
			DCTCNN~\cite{DCTCNN}              & \newcrossmark         & \newcrossmark     & \newcrossmark       & Real + Synthetic                   & Supervised                 & Classifier trained on DCT of real and synthetic images                                                                                                                                                                                       \\
			RepMix~\cite{Repmix}              & \newcrossmark         & \newcheckmark     & \newcrossmark       & Real + Synthetic                   & Supervised                 & Classifier trained with representational mixing                                                                                                                                                                                              \\
			POSE~\cite{POSE}                  & \newcrossmark         & \newcheckmark     & \newcheckmark       & Real + Synthetic                   & Open-Set                   & {Progressively enlarge the embedding space of classes using learned augmentations}                                                                                                                                                           \\
			Fang et al.~\cite{Fang_2023_BMVC} & \newcrossmark         & \newcheckmark     & \newcheckmark       & Real + Synthetic                   & Open-Set                   & Learned transferable embeddings using ProxyNCA applied on a CNN                                                                                                                                                                              \\
			Abady et al.~\cite{Abady}         & \newcrossmark         & \newcheckmark     & \newcheckmark       & Real + Synthetic                   & Open-Set                   & {Learned embedding space of classes using siamese network with learned distance metric}                                                                                                                                                      \\
			FSM~\cite{FSM}                    & \newcrossmark         & \newhalfcirc      & \newcrossmark       & Real                               & Supervised                 & {Learned embedding space of different camera models using siamese network with learned distance metric}                                                                                                                                      \\
			ExifNet~\cite{ExifNet}            & \newcrossmark         & \newhalfcirc      & \newcrossmark       & Real                               & Supervised                 & {Learned embedding space of images' Exif data using siamese network with learned distance metric}                                                                                                                                            \\
			CLIP~\cite{CLIP}                  & \newcrossmark         & \newcheckmark     & \newhalfcirc        & Real                               & Self-Supervised            & Learned transferable visual embeddings grounded by text captions                                                                                                                                                                             \\
			ResNet-50~\cite{imagenet}         & \newcrossmark         & \newcheckmark     & \newhalfcirc        & Real                               & Supervised                 & {Learned transferable visual embeddings by training on large corpus of real images with many classes}                                                                                                                                        \\
			\textbf{Ours}                     & \newcheckmark         & \newcheckmark     & \newcheckmark       & Real                               & Self-Supervised            & {The self-descriptions of the forensic microstructures in real images are naturally different than those of synthetic images. Self-descriptions of images created by different generators are also distinct, attributable and cluster-able.}
		\end{tblr}
	}
\end{table*}

%% file: sections/suppl/Data_v0.tex
\section{Data Composition}
\label{supp:data}

\input{tables/RealDatasetsComp}

\input{tables/SynthDatasetsComp}

In this section, we discuss the composition of the datasets used in our paper.

Tab.~\ref{tab:real_data_comp} summarizes the real image datasets used in our experiments, highlighting their diverse range of resolutions and topics. The datasets include COCO2017~\cite{coco}, IN-1k~\cite{imagenet}, IN-22k~\cite{imagenet22k}, and MIDB~\cite{mislnet, cam-openset}, covering resolutions from as low as \(32 \times 25\) to as high as \(5248 \times 6016\). This diversity ensures that our method is trained and evaluated on real images that represent a broad variety of scenes, resolutions, and domains, minimizing potential biases and enhancing its generalizability. Notably, our method is trained exclusively on the training samples of real images and does not see the synthetic images during training, supporting its zero-shot detection capability.

Tab.~\ref{tab:synth_data_comp} provides an overview of the synthetic image datasets used in our study, which are drawn from OSSIA~\cite{Fang_2023_BMVC}, DMID~\cite{corvi2023detection}, SB~\cite{Synthbuster}, and our own generations. These datasets include synthetic images generated by a wide range of models, such as BigGAN, DALLE variants, StyleGAN, and Stable Diffusion versions, covering diverse resolutions from \(256 \times 256\) to \(1792 \times 1792\). Notably, DMID and SB datasets are primarily evaluation-only, with no training samples, except for Latent Diffusion and Guided Diffusion from DMID. This comprehensive collection ensures robust evaluation across diverse generative models, demonstrating the adaptability and generalization of our method to various synthetic sources.

%% file: tables/RealDatasetsComp.tex
\begin{table}
	\centering
	\caption{\label{tab:real_data_comp} Composition of datasets of real images used in this paper.  We note that our method only sees the training samples of real images during training.}
	\resizebox*{1.0\linewidth}{!}{
		\begin{tblr}{
			width = \linewidth,
			colspec = {m{23mm}m{32mm}m{22mm}m{20mm}},
			row{1} = {c},
			cell{1}{1} = {c=4}{0.938\linewidth},
			cell{2-6}{3-4} = {r},
			hlines,
			vlines,
			hline{1-3,7} = {-}{black},
				}
			\textbf{Real Images Datasets}     &                      &                        &                       \\
			\textbf{Source}                   & \textbf{Image Sizes} & \textbf{Train Samples} & \textbf{Test Samples} \\
			COCO2017~\cite{coco}              & 51-640 x 59-640      & 100000                 & 1000                  \\
			IN-1k~\cite{imagenet}             & 32-5980 x 25-4768    & 100000                 & 1000                  \\
			IN-22k~\cite{imagenet22k}         & 56-1857 x 56-2091    & 100000                 & 1000                  \\
			MIDB~\cite{mislnet, cam-openset}  & 480-5248 x 640-6016  & 22329                  & 1000
		\end{tblr}
	}
\end{table}

%% file: tables/SynthDatasetsComp.tex
\begin{table}
	\centering
	\caption{\label{tab:synth_data_comp} Composition of datasets of synthetic images used in this paper. These datasets are pooled together from OSSIA~\cite{Fang_2023_BMVC}, DMID~\cite{corvi2023detection}, SB~\cite{Synthbuster}, and our own generations. We note that in the zero-shot experiment, our method \underline{does not} see any synthetic images during training.}
	\resizebox{1.0\linewidth}{!}{%
		\begin{tblr}{
			width = \linewidth,
			colspec = {m{20mm}m{16mm}m{35mm}m{22mm}m{20mm}},
			row{1} = {c},
			cell{1}{1} = {c=5}{0.934\linewidth},
			cell{2-27}{4-5} = {r},
			cell{27}{1} = {c=3}{0.512\linewidth},
			vlines,
			hline{1-3,27-28} = {-}{},
				}
			\textbf{Synthetic Image Datasets } &                  &                       &                        &                       \\
			\textbf{Generator}                 & \textbf{Sources} & \textbf{Image Sizes}  & \textbf{Train Samples} & \textbf{Test Samples} \\
			BigGAN                             & DMID             & 256-512 x 256-512     & 0                      & 1000                  \\
			DALLE 2                            & DMID, SB         & 1024-1024 x 1024-1024 & 0                      & 2000                  \\
			DALLE 3                            & Ours, SB         & 1024-1792 x 1024-1792 & 4000                   & 2000                  \\
			DALLE M                            & DMID             & 256-256 x 256-256     & 0                      & 1000                  \\
			EG3D                               & DMID             & 512-512 x 512-512     & 0                      & 1000                  \\
			FireFly                            & SB               & 1536-2304 x 1792-2688 & 0                      & 1000                  \\
			GigaGAN                            & DMID             & 256-1024 x 256-1024   & 0                      & 1000                  \\
			GLIDE                              & DMID, SB         & 256-256 x 256-256     & 0                      & 2000                  \\
			Guided Dif                         & DMID             & 256-256 x 256-256     & 1000                   & 1000                  \\
			Latent Dif                         & DMID             & 256-256 x 256-256     & 2000                   & 1000                  \\
			MJ v5                              & SB               & 896-1360 x 896-1360   & 0                      & 1000                  \\
			MJ v6                              & Ours             & 768-1344 x 896-1536   & 25000                  & 1000                  \\
			ProGAN                             & OSSIA            & 256-256 x 256-256     & 25000                  & 1000                  \\
			Proj.GAN                           & OSSIA            & 256-256 x 256-256     & 25000                  & 1000                  \\
			SD1.3                              & SB               & 512-512 x 512-512     & 0                      & 1000                  \\
			SD1.4                              & OSSIA, SB        & 512-512 x 512-512     & 25000                  & 2000                  \\
			SD1.5                              & Ours             & 768-768 x 768-768     & 10000                  & 1000                  \\
			SD2.1                              & SB               & 576-1408 x 704-1728   & 0                      & 1000                  \\
			SD3.0                              & Ours             & 1024-1024 x 1024-1024 & 10000                  & 1000                  \\
			SDXL                               & Ours, SB         & 576-1408 x 704-1728   & 25000                  & 2000                  \\
			StyleGAN                           & OSSIA            & 256-1024 x 256-1024   & 25000                  & 1000                  \\
			StyleGAN2                          & OSSIA            & 512-1024 x 512-1024   & 25000                  & 1000                  \\
			StyleGAN3                          & OSSIA            & 256-1024 x 256-1024   & 25000                  & 1000                  \\
			Tam.Xformer                        & OSSIA            & 256-256 x 256-256     & 25000                  & 1000                  \\
			\textbf{Total }                    &                  &                       & \textbf{252000}        & \textbf{29000}
		\end{tblr}
	}
\end{table}

%% file: figures/CrossValDiffReals.tex
\begin{figure*}[!ht]
	\centering
	\includegraphics[width=0.7\linewidth]{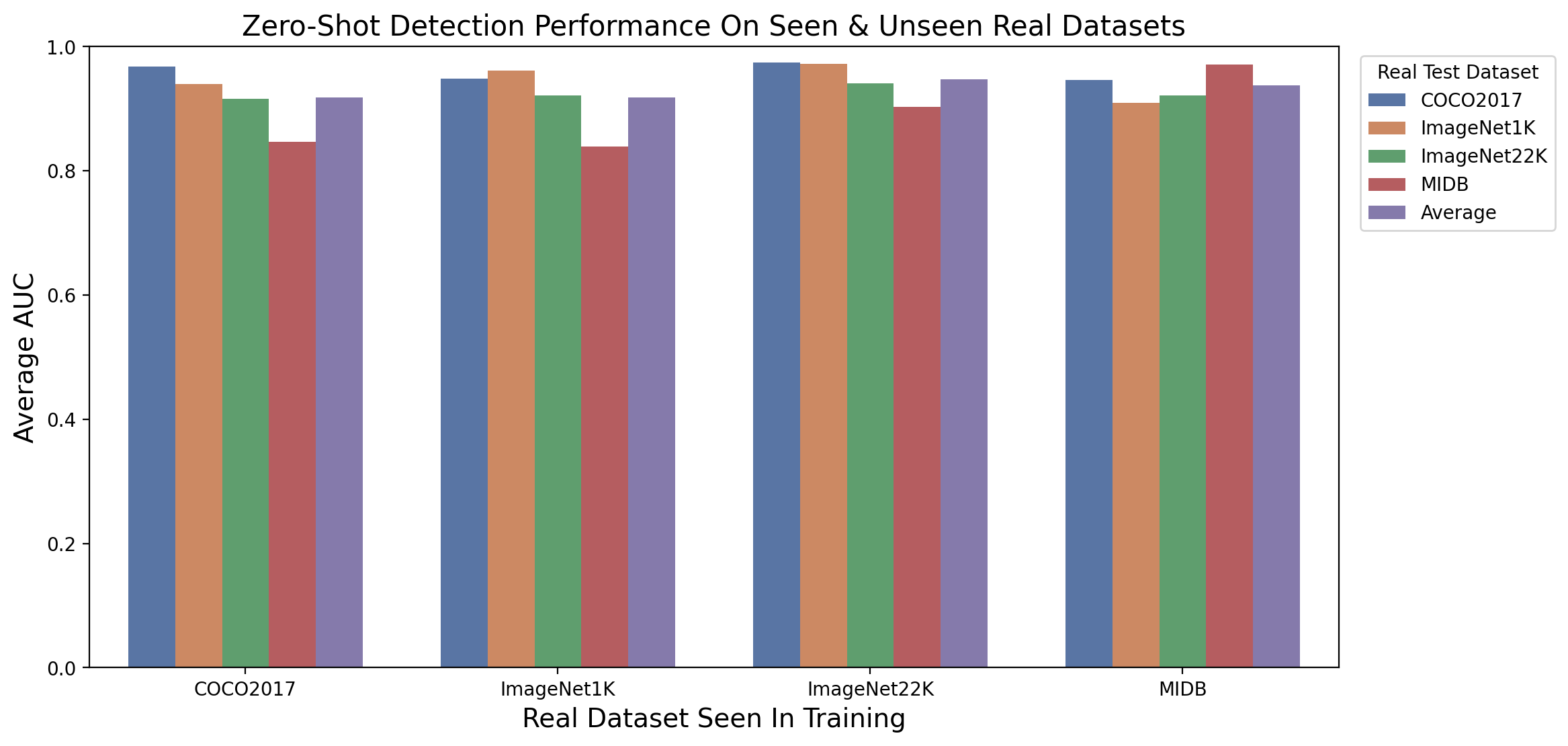}
	\pulluppp
	\caption{Zero-shot detection performance of our method evaluated on real datasets that are not seen during training. Performance on seen dataset is also provided for comparison.}
	\label{fig:cross_val_diff_reals}
	\pulluppp\pullup
\end{figure*}

%% file: sections/suppl/CategoryTaxonomy_v0.tex
\section{Competing Methods Categories and Taxonomy}
\label{supp:cat_and_tax}

Tab.~\ref{tab:cat_and_tax} presents a comprehensive comparison of various methods for synthetic image detection and source attribution, categorizing them based on their capabilities, training data requirements, training paradigms, and underlying approaches. The capabilities considered are zero-shot detection, open-set recognition, and clustering—key features that determine a method's ability to generalize to unseen data and accurately attribute sources.

Most existing methods rely on supervised learning paradigms and require both real and synthetic images for training. For instance, CnnDet~\cite{CNNDet} and PatchFor~\cite{PatchFor} train classifiers on known synthetic sources, focusing on low-level artifacts or standard classification techniques. While these methods can sometimes generalize to similar generative models, they lack zero-shot capabilities and struggle with open-set scenarios where new types of synthetic images emerge. They also do not support clustering, limiting their utility in organizing images based on source similarities.

Some methods, like LGrad~\cite{LGrad}, UFD~\cite{UFD}, DE-FAKE~\cite{DE-Fake}, Aeroblade~\cite{Aeroblade}, ZED~\cite{Zed}, and NPR~\cite{NPR}, offer zero-shot detection capabilities. LGrad trains classifiers on gradients of a common CNN, while DE-FAKE and UFD leverage embeddings from models like CLIP and BLIP. Aeroblade is unique in being training-free, using reconstruction errors from pretrained diffusion models. ZED employs a self-supervised approach, using a lossless neural compressor trained on real images. However, despite their zero-shot capabilities, these methods generally do not support open-set recognition or clustering. They are limited to distinguishing real from synthetic images and often cannot attribute images to specific unknown sources or organize them based on source characteristics.

Open-set recognition and clustering are addressed by methods like RepMix~\cite{Repmix}, POSE~\cite{POSE}, Fang et al.~\cite{Fang_2023_BMVC}, and Abady et al.~\cite{Abady}. These methods utilize supervised or open-set training paradigms and require both real and synthetic images for training. RepMix introduces representational mixing to handle unseen classes, while POSE progressively enlarges the embedding space using learned augmentations. Fang et al. and Abady et al. focus on learning transferable embeddings through techniques like ProxyNCA and siamese networks with learned distance metrics. Although these methods can perform open-set recognition and clustering, they lack zero-shot detection capabilities, meaning they require prior exposure to synthetic sources to function effectively.

Our proposed method distinguishes itself by offering all three capabilities: zero-shot detection, open-set source attribution, and clustering, while requiring only real images for training. By modeling forensic microstructures through diverse predictive filters, we extract residuals that encapsulate intrinsic forensic properties unique to the image creation process. These residuals are used to compute forensic self-descriptions, which naturally differ between real and synthetic images and across different generators. This enables robust zero-shot detection by modeling real-image self-description distributions and detecting deviations. Additionally, the distinctiveness of self-descriptions supports open-set attribution and clustering, providing a generalizable and efficient solution without relying on synthetic training data.

%% file: sections/suppl/FullZeroShotResults_v0.tex
\section{Full Zero-Shot Results}
\label{supp:full_zeroshot}

\input{tables/FullZeroShotCOCO2017}

\input{tables/FullZeroShotIN1k}

\input{tables/FullZeroShotIN22k}

\input{tables/FullZeroShotMIDB}

In this section, we present zero-shot performances between all real-vs-synthetic dataset pairs. These results are shown in Tab.~\ref{tab:full_zs_coco2017},~\ref{tab:full_zs_in1k},~\ref{tab:full_zs_in22k}, and~\ref{tab:full_zs_midb}.

These results, in conjunction with those presented in Tab.~\ref{tab:zero_shot} and~\ref{tab:zero_shot_min} of the main paper, highlight the exceptional generalizability and consistency of our method across a wide range of real sources and synthetic generators. While some other methods achieve high overall average AUC scores, their performance often drops significantly in worst-case scenarios. For instance, NPR demonstrates a strong overvall average AUC of 0.926 but fails on the Firefly generator, with worst-case AUCs as low as 0.239 on the IN-1k dataset. In contrast, our method not only achieves the highest overall average AUC of 0.960 but also maintains consistently high worst-case AUCs, with a minimum of 0.714 on IN-22k, even for challenging generators like GLIDE. This stability reflects our method's ability to generalize effectively to unseen generators.

Compared to other methods that also rely solely on real images for training, such as ZED, our approach demonstrates significant advantages. ZED achieves an average AUC of 0.723 but struggles with specific generators like ProGAN, with worst-case AUCs around 0.375. By leveraging forensic self-descriptions, our method captures intrinsic forensic properties that remain robust across diverse generators, avoiding the pitfalls of methods that depend on synthetic training data or fail to generalize to new generators.

Additionally, our method shows exceptional adaptability in handling challenging cases that cause other methods to fail, such as BigGAN and Firefly. The ability to achieve strong performance even in worst-case scenarios underscores the effectiveness of our forensic self-description approach. This resilience, combined with the exclusive use of real images during training, positions our method as a reliable and generalizable solution for zero-shot detection of synthetic images.

%% file: tables/FullZeroShotCOCO2017.tex
\begin{table*}[!t]
	\centering
	\caption{\label{tab:full_zs_coco2017} Zero-shot detection performance, measured in AUC, between each synthetic generator and COCO2017.}
	\resizebox{\linewidth}{!}{%
		\begin{tblr}{
			width = \linewidth,
			colspec = {m{15mm} *{25}{m{8.5mm}}},
			column{even} = {c},
			column{odd} = {c},
			column{1} = {l},
			vline{1-3,27} = {-}{},
			hline{1-2,10-11} = {-}{},
				}
			\textbf{Method} & \textbf{Avg.}  & \textbf{ProG}  & \textbf{Prj.G} & \textbf{SG}    & \textbf{SG2}   & \textbf{SG3}   & \textbf{BigG}  & \textbf{GigaG} & \textbf{Eg3d}  & \textbf{Tm.Xf} & \textbf{Glide} & \textbf{G.Dif.} & \textbf{L.Dif.} & \textbf{SD1.3} & \textbf{SD1.4} & \textbf{SD1.5} & \textbf{SD2.1} & \textbf{SDXL}  & \textbf{SD3.0} & \textbf{DLEM}  & \textbf{DLE2}  & \textbf{DLE3}  & \textbf{MJv5}  & \textbf{MJv6}  & \textbf{Firefly} \\
			CnnDet          & 0.756          & 0.999          & 0.803          & 0.994          & 0.944          & 0.940          & 0.923          & 0.726          & 0.939          & 0.654          & 0.733          & 0.775           & 0.752           & 0.702          & 0.685          & 0.521          & 0.683          & 0.725          & 0.702          & 0.657          & 0.804          & 0.477          & 0.598          & 0.570          & 0.834            \\
			PatchFor        & 0.833          & 0.806          & 0.953          & \textbf{0.995} & 0.845          & 0.772          & 0.939          & 0.831          & 0.890          & 0.918          & 0.850          & 0.819           & 0.952           & 0.917          & 0.896          & 0.885          & 0.547          & 0.887          & 0.751          & 0.943          & 0.884          & 0.564          & 0.687          & 0.846          & 0.620            \\
			LGrad           & 0.819          & 0.954          & 0.800          & 0.972          & 0.896          & 0.890          & 0.862          & 0.837          & 0.913          & 0.729          & 0.819          & 0.773           & 0.871           & 0.818          & 0.818          & 0.827          & 0.617          & 0.808          & 0.859          & 0.778          & 0.851          & 0.734          & 0.795          & 0.774          & 0.657            \\
			UFD             & 0.903          & \textbf{1.000} & 0.976          & \textbf{0.995} & 0.896          & \textbf{0.990} & \textbf{0.997} & 0.964          & 0.988          & 0.976          & 0.872          & 0.894           & 0.916           & 0.934          & 0.928          & 0.740          & \textbf{0.946} & 0.813          & 0.732          & 0.976          & 0.980          & 0.680          & 0.780          & 0.709          & \textbf{0.992}   \\
			DE-FAKE         & 0.765          & 0.728          & 0.799          & 0.727          & 0.894          & 0.590          & 0.534          & 0.646          & 0.601          & 0.839          & 0.905          & 0.723           & 0.812           & 0.795          & 0.839          & 0.850          & 0.694          & 0.791          & 0.943          & 0.795          & 0.560          & 0.922          & 0.775          & 0.900          & 0.694            \\
			Aeroblade       & 0.728          & 0.520          & 0.718          & 0.891          & 0.472          & 0.664          & 0.425          & 0.537          & 0.714          & 0.566          & 0.883          & 0.720           & 0.719           & 0.811          & 0.872          & \textbf{0.982} & 0.828          & 0.792          & 0.741          & 0.730          & 0.596          & 0.745          & 0.900          & 0.938          & 0.706            \\
			ZED             & 0.751          & 0.462          & 0.667          & 0.880          & 0.811          & 0.840          & 0.713          & 0.727          & 0.824          & 0.766          & 0.663          & 0.682           & 0.729           & 0.812          & 0.814          & 0.777          & 0.702          & 0.798          & 0.813          & 0.830          & 0.847          & 0.715          & 0.803          & 0.801          & 0.563            \\
			NPR             & 0.945          & 0.993          & \textbf{0.988} & 0.994          & \textbf{0.992} & 0.986          & 0.981          & 0.959          & \textbf{0.993} & \textbf{0.992} & 0.984          & 0.916           & \textbf{0.992}  & \textbf{0.986} & \textbf{0.985} & 0.971          & 0.921          & \textbf{0.975} & 0.982          & 0.970          & 0.985          & 0.844          & 0.935          & 0.969          & 0.396            \\
			\textbf{Ours}   & \textbf{0.968} & 0.989          & 0.979          & 0.905          & 0.942          & 0.973          & 0.990          & \textbf{0.987} & 0.955          & 0.991          & \textbf{0.992} & \textbf{0.991}  & 0.989           & 0.951          & 0.944          & 0.892          & 0.926          & 0.971          & \textbf{0.994} & \textbf{0.987} & \textbf{0.993} & \textbf{0.963} & \textbf{0.977} & \textbf{0.976} & 0.987
		\end{tblr}
	}
\end{table*}

%% file: tables/FullZeroShotIN1k.tex
\begin{table*}[!t]
	\centering
	\caption{\label{tab:full_zs_in1k} Zero-shot detection performance, measured in AUC, between each synthetic generator and ImageNet-1K.}
	\resizebox{\linewidth}{!}{%
		\begin{tblr}{
			width = \linewidth,
			colspec = {m{15mm} *{25}{m{8.5mm}}},
			column{even} = {c},
			column{odd} = {c},
			column{1} = {l},
			vline{1-3,27} = {-}{},
			hline{1-2,10-11} = {-}{},
				}
			\textbf{Method} & \textbf{Avg.}  & \textbf{ProG}  & \textbf{Prj.G} & \textbf{SG}    & \textbf{SG2}   & \textbf{SG3}   & \textbf{BigG}  & \textbf{GigaG} & \textbf{Eg3d}  & \textbf{Tm.Xf} & \textbf{Glide} & \textbf{G.Dif.} & \textbf{L.Dif.} & \textbf{SD1.3} & \textbf{SD1.4} & \textbf{SD1.5} & \textbf{SD2.1} & \textbf{SDXL}  & \textbf{SD3.0} & \textbf{DLEM}  & \textbf{DLE2}  & \textbf{DLE3}  & \textbf{MJv5}  & \textbf{MJv6}  & \textbf{Firefly} \\
			CnnDet          & 0.714          & 0.999          & 0.751          & \textbf{0.995} & 0.946          & 0.926          & 0.903          & 0.673          & 0.922          & 0.599          & 0.678          & 0.729           & 0.702           & 0.644          & 0.626          & 0.458          & 0.627          & 0.675          & 0.646          & 0.600          & 0.760          & 0.424          & 0.539          & 0.510          & 0.792            \\
			PatchFor        & 0.823          & 0.799          & 0.948          & 0.994          & 0.841          & 0.763          & 0.934          & 0.821          & 0.876          & 0.907          & 0.829          & 0.804           & 0.942           & 0.905          & 0.882          & 0.871          & 0.543          & 0.874          & 0.739          & 0.933          & 0.868          & 0.564          & 0.679          & 0.834          & 0.613            \\
			LGrad           & 0.770          & 0.914          & 0.738          & 0.938          & 0.891          & 0.820          & 0.774          & 0.782          & 0.812          & 0.676          & 0.787          & 0.728           & 0.809           & 0.720          & 0.731          & 0.839          & 0.658          & 0.777          & 0.769          & 0.731          & 0.803          & 0.696          & 0.716          & 0.742          & 0.625            \\
			UFD             & 0.862          & \textbf{1.000} & 0.952          & 0.985          & 0.850          & 0.978          & \textbf{0.993} & \textbf{0.939} & 0.971          & 0.953          & 0.811          & 0.804           & 0.874           & 0.895          & 0.884          & 0.661          & 0.913          & 0.751          & 0.643          & \textbf{0.956} & 0.960          & 0.607          & 0.705          & 0.623          & 0.982            \\
			DE-FAKE         & 0.749          & 0.641          & 0.725          & 0.768          & 0.872          & 0.627          & 0.487          & 0.554          & 0.581          & 0.778          & 0.814          & 0.644           & 0.738           & 0.823          & 0.841          & 0.880          & 0.710          & 0.834          & 0.911          & 0.735          & 0.635          & 0.894          & 0.810          & 0.889          & 0.785            \\
			Aeroblade       & 0.741          & 0.554          & 0.734          & 0.884          & 0.508          & 0.690          & 0.458          & 0.566          & 0.733          & 0.598          & 0.883          & 0.735           & 0.732           & 0.814          & 0.869          & 0.973          & 0.828          & 0.802          & 0.753          & 0.744          & 0.618          & 0.759          & 0.896          & 0.931          & 0.721            \\
			ZED             & 0.676          & 0.402          & 0.562          & 0.790          & 0.741          & 0.750          & 0.632          & 0.646          & 0.743          & 0.692          & 0.594          & 0.618           & 0.672           & 0.740          & 0.733          & 0.690          & 0.623          & 0.732          & 0.756          & 0.752          & 0.783          & 0.651          & 0.719          & 0.734          & 0.473            \\
			NPR             & 0.900          & 0.979          & \textbf{0.969} & 0.983          & 0.978          & 0.964          & 0.943          & 0.902          & \textbf{0.980} & \textbf{0.975} & \textbf{0.954} & 0.882           & \textbf{0.974}  & \textbf{0.960} & 0.964          & 0.917          & 0.816          & 0.938          & 0.948          & 0.908          & 0.956          & 0.713          & 0.847          & 0.918          & 0.239            \\
			\textbf{Ours}   & \textbf{0.962} & 0.955          & 0.930          & 0.984          & \textbf{0.995} & \textbf{0.999} & 0.912          & 0.903          & 0.975          & 0.927          & 0.949          & \textbf{0.922}  & 0.925           & 0.923          & \textbf{0.979} & \textbf{0.977} & \textbf{0.978} & \textbf{0.993} & \textbf{0.978} & 0.944          & \textbf{0.976} & \textbf{1.000} & \textbf{0.985} & \textbf{0.986} & \textbf{0.994}
		\end{tblr}
	}
\end{table*}

%% file: tables/FullZeroShotIN22k.tex
\begin{table*}[!t]
	\centering
	\caption{\label{tab:full_zs_in22k} Zero-shot detection performance, measured in AUC, between each synthetic generator and ImageNet-22k.}
	\resizebox{\linewidth}{!}{%
		\begin{tblr}{
			width = \linewidth,
			colspec = {m{15mm} *{25}{m{8.5mm}}},
			column{even} = {c},
			column{odd} = {c},
			column{1} = {l},
			vline{1-3,27} = {-}{},
			hline{1-2,10-11} = {-}{},
				}
			\textbf{Method} & \textbf{Avg.}  & \textbf{ProG}  & \textbf{Prj.G} & \textbf{SG}    & \textbf{SG2}   & \textbf{SG3}   & \textbf{BigG}  & \textbf{GigaG} & \textbf{Eg3d}  & \textbf{Tm.Xf} & \textbf{Glide} & \textbf{G.Dif.} & \textbf{L.Dif.} & \textbf{SD1.3} & \textbf{SD1.4} & \textbf{SD1.5} & \textbf{SD2.1} & \textbf{SDXL}  & \textbf{SD3.0} & \textbf{DLEM}  & \textbf{DLE2}  & \textbf{DLE3}  & \textbf{MJv5}  & \textbf{MJv6}  & \textbf{Firefly} \\
			CnnDet          & 0.733          & \textbf{0.999} & 0.779          & 0.997          & 0.956          & 0.940          & 0.918          & 0.694          & 0.936          & 0.622          & 0.704          & 0.751           & 0.727           & 0.670          & 0.650          & 0.474          & 0.651          & 0.697          & 0.668          & 0.622          & 0.783          & 0.439          & 0.560          & 0.530          & 0.817            \\
			PatchFor        & 0.845          & 0.821          & \textbf{0.958} & \textbf{0.998} & 0.852          & 0.789          & 0.945          & 0.844          & 0.897          & 0.925          & 0.859          & 0.832           & 0.957           & 0.925          & 0.904          & 0.894          & 0.565          & 0.895          & 0.769          & \textbf{0.949} & 0.892          & 0.594          & 0.709          & 0.856          & 0.643            \\
			LGrad           & 0.866          & 0.951          & 0.850          & 0.965          & 0.936          & 0.897          & 0.871          & 0.876          & 0.895          & 0.812          & 0.859          & 0.836           & 0.893           & 0.840          & 0.845          & 0.910          & 0.798          & 0.873          & 0.867          & 0.844          & 0.886          & 0.816          & 0.836          & 0.849          & 0.776            \\
			UFD             & 0.815          & \textbf{0.999} & 0.921          & 0.972          & 0.772          & 0.959          & \textbf{0.988} & 0.904          & 0.949          & 0.919          & 0.732          & 0.771           & 0.807           & 0.845          & 0.838          & 0.568          & 0.875          & 0.676          & 0.553          & 0.931          & 0.933          & 0.527          & 0.614          & 0.534          & 0.970            \\
			DE-FAKE         & 0.617          & 0.584          & 0.648          & 0.558          & 0.753          & 0.424          & 0.383          & 0.492          & 0.431          & 0.706          & 0.782          & 0.580           & 0.672           & 0.643          & 0.699          & 0.706          & 0.533          & 0.642          & 0.825          & 0.644          & 0.396          & 0.795          & 0.618          & 0.769          & 0.527            \\
			Aeroblade       & 0.582          & 0.405          & 0.544          & 0.713          & 0.378          & 0.499          & 0.336          & 0.420          & 0.527          & 0.437          & 0.752          & 0.584           & 0.583           & 0.617          & 0.696          & 0.862          & 0.637          & 0.637          & 0.605          & 0.579          & 0.468          & 0.588          & 0.742          & 0.792          & 0.565            \\
			ZED             & 0.716          & 0.375          & 0.603          & 0.830          & 0.771          & 0.789          & 0.789          & 0.689          & 0.775          & 0.738          & 0.643          & 0.665           & 0.729           & 0.765          & 0.766          & 0.725          & 0.668          & 0.782          & 0.791          & 0.791          & 0.809          & 0.686          & 0.757          & 0.752          & 0.507            \\
			NPR             & 0.900          & 0.966          & \textbf{0.958} & 0.969          & 0.966          & 0.953          & 0.936          & 0.903          & 0.967          & \textbf{0.962} & \textbf{0.947} & \textbf{0.891}  & \textbf{0.962}  & 0.949          & 0.948          & 0.915          & 0.844          & 0.968          & 0.940          & 0.908          & 0.929          & 0.750          & 0.867          & 0.917          & 0.295            \\
			\textbf{Ours}   & \textbf{0.941} & 0.930          & 0.895          & 0.933          & \textbf{0.975} & \textbf{0.991} & 0.912          & \textbf{0.917} & \textbf{0.970} & 0.917          & 0.714          & 0.852           & 0.893           & \textbf{0.971} & \textbf{0.969} & \textbf{0.977} & \textbf{0.966} & \textbf{0.988} & \textbf{0.983} & 0.913          & \textbf{0.976} & \textbf{0.971} & \textbf{0.982} & \textbf{0.989} & \textbf{0.992}
		\end{tblr}
	}
\end{table*}

%% file: tables/FullZeroShotMIDB.tex
\begin{table*}[!t]
	\centering
	\caption{\label{tab:full_zs_midb} Zero-shot detection performance, measured in AUC, between each synthetic generator and MISL Image Database (MIDB).}
	\resizebox{\linewidth}{!}{%
		\begin{tblr}{
			width = \linewidth,
			colspec = {m{15mm} *{25}{m{8.5mm}}},
			column{even} = {c},
			column{odd} = {c},
			column{1} = {l},
			vline{1-3,27} = {-}{},
			hline{1-2,10-11} = {-}{},
				}
			\textbf{Method} & \textbf{Avg.}  & \textbf{ProG}  & \textbf{Prj.G} & \textbf{SG}    & \textbf{SG2}   & \textbf{SG3}   & \textbf{BigG}  & \textbf{GigaG} & \textbf{Eg3d}  & \textbf{Tm.Xf} & \textbf{Glide} & \textbf{G.Dif.} & \textbf{L.Dif.} & \textbf{SD1.3} & \textbf{SD1.4} & \textbf{SD1.5} & \textbf{SD2.1} & \textbf{SDXL}  & \textbf{SD3.0} & \textbf{DLEM}  & \textbf{DLE2}  & \textbf{DLE3}  & \textbf{MJv5}  & \textbf{MJv6}  & \textbf{Firefly} \\
			CnnDet          & 0.683          & \textbf{1.000} & 0.720          & 0.999          & 0.950          & 0.932          & 0.900          & 0.635          & 0.927          & 0.551          & 0.637          & 0.696           & 0.664           & 0.597          & 0.581          & 0.407          & 0.581          & 0.638          & 0.604          & 0.555          & 0.734          & 0.373          & 0.487          & 0.457          & 0.769            \\
			PatchFor        & 0.790          & 0.777          & 0.919          & 0.970          & 0.819          & 0.741          & 0.897          & 0.786          & 0.836          & 0.855          & 0.779          & 0.765           & 0.892           & 0.856          & 0.832          & 0.820          & 0.536          & 0.832          & 0.713          & 0.886          & 0.818          & 0.573          & 0.665          & 0.790          & 0.610            \\
			UFD             & 0.612          & 0.994          & 0.745          & 0.856          & 0.504          & 0.831          & 0.947          & 0.727          & 0.776          & 0.723          & 0.425          & 0.495           & 0.547           & 0.621          & 0.608          & 0.272          & 0.690          & 0.415          & 0.255          & 0.786          & 0.776          & 0.270          & 0.312          & 0.244          & 0.883            \\
			LGrad           & 0.824          & 0.959          & 0.808          & 0.978          & 0.900          & 0.900          & 0.872          & 0.844          & 0.923          & 0.730          & 0.815          & 0.771           & 0.881           & 0.828          & 0.826          & 0.839          & 0.606          & 0.815          & 0.864          & 0.780          & 0.859          & 0.732          & 0.802          & 0.777          & 0.655            \\
			DE-FAKE         & 0.791          & 0.753          & 0.825          & 0.759          & 0.915          & 0.624          & 0.563          & 0.675          & 0.636          & 0.862          & 0.924          & 0.748           & 0.836           & 0.823          & 0.863          & 0.875          & 0.725          & 0.818          & 0.960          & 0.822          & 0.594          & \textbf{0.941} & 0.804          & 0.921          & 0.728            \\
			Aeroblade       & 0.646          & 0.440          & 0.606          & 0.813          & 0.406          & 0.547          & 0.360          & 0.457          & 0.578          & 0.477          & 0.826          & 0.645           & 0.645           & 0.695          & 0.783          & 0.954          & 0.719          & 0.708          & 0.669          & 0.647          & 0.517          & 0.657          & 0.831          & 0.885          & 0.627            \\
			ZED             & 0.747          & 0.331          & 0.599          & 0.872          & 0.801          & 0.835          & 0.729          & 0.744          & 0.898          & 0.763          & 0.699          & 0.745           & 0.760           & 0.836          & 0.803          & 0.774          & 0.647          & 0.800          & 0.812          & 0.855          & 0.891          & 0.713          & 0.730          & 0.775          & 0.513            \\
			NPR             & 0.957          & 0.994          & 0.990          & 0.995          & \textbf{0.994} & 0.991          & 0.985          & 0.966          & 0.994          & 0.994          & 0.987          & 0.963           & 0.993           & \textbf{0.990} & \textbf{0.986} & \textbf{0.980} & \textbf{0.947} & \textbf{0.990} & \textbf{0.987} & 0.977          & 0.988          & 0.876          & 0.955          & \textbf{0.989} & 0.449            \\
			\textbf{Ours}   & \textbf{0.971} & \textbf{1.000} & \textbf{1.000} & \textbf{1.000} & 0.989          & \textbf{0.998} & \textbf{0.993} & \textbf{0.995} & \textbf{1.000} & \textbf{0.998} & \textbf{1.000} & \textbf{0.993}  & \textbf{0.996}  & 0.959          & 0.941          & 0.952          & 0.903          & 0.962          & 0.956          & \textbf{0.995} & \textbf{0.993} & 0.931          & \textbf{0.965} & 0.896          & \textbf{0.896}
		\end{tblr}
	}
\end{table*}

%% file: figures/SpecOfReals.tex
\begin{figure*}[!t]
	\pullupp
	\centering
	\setlength{\fboxsep}{0pt}
	\setlength{\fboxrule}{0pt}
	\resizebox{0.80\linewidth}{!}{
		\begin{tabular}{
				*{9}{@{\hskip 2pt} p{0.090\textwidth} @{\hskip 2pt}}
			}
			\makebox[0.089\textwidth]
			& \makebox[0.089\textwidth]{\raisebox{1pt}{\fontsize{9pt}{10.5pt}\selectfont $FFT(\phi_1)$}}
			& \makebox[0.089\textwidth]{\raisebox{1pt}{\fontsize{9pt}{10.5pt}\selectfont $FFT(\phi_2)$}}
			& \makebox[0.089\textwidth]{\raisebox{1pt}{\fontsize{9pt}{10.5pt}\selectfont $FFT(\phi_3)$}}
			& \makebox[0.089\textwidth]{\raisebox{1pt}{\fontsize{9pt}{10.5pt}\selectfont $FFT(\phi_4)$}}
			& \makebox[0.089\textwidth]{\raisebox{1pt}{\fontsize{9pt}{10.5pt}\selectfont $FFT(\phi_5)$}}
			& \makebox[0.089\textwidth]{\raisebox{1pt}{\fontsize{9pt}{10.5pt}\selectfont $FFT(\phi_6)$}}
			& \makebox[0.089\textwidth]{\raisebox{1pt}{\fontsize{9pt}{10.5pt}\selectfont $FFT(\phi_7)$}}
			& \makebox[0.089\textwidth]{\raisebox{1pt}{\fontsize{9pt}{10.5pt}\selectfont $FFT(\phi_8)$}} \\

			\raisebox{19.5pt}{\fontsize{9pt}{10.5pt}\selectfont COCO2017\hspace{2pt}}
			& \fbox{\includegraphics[width=0.089\textwidth]{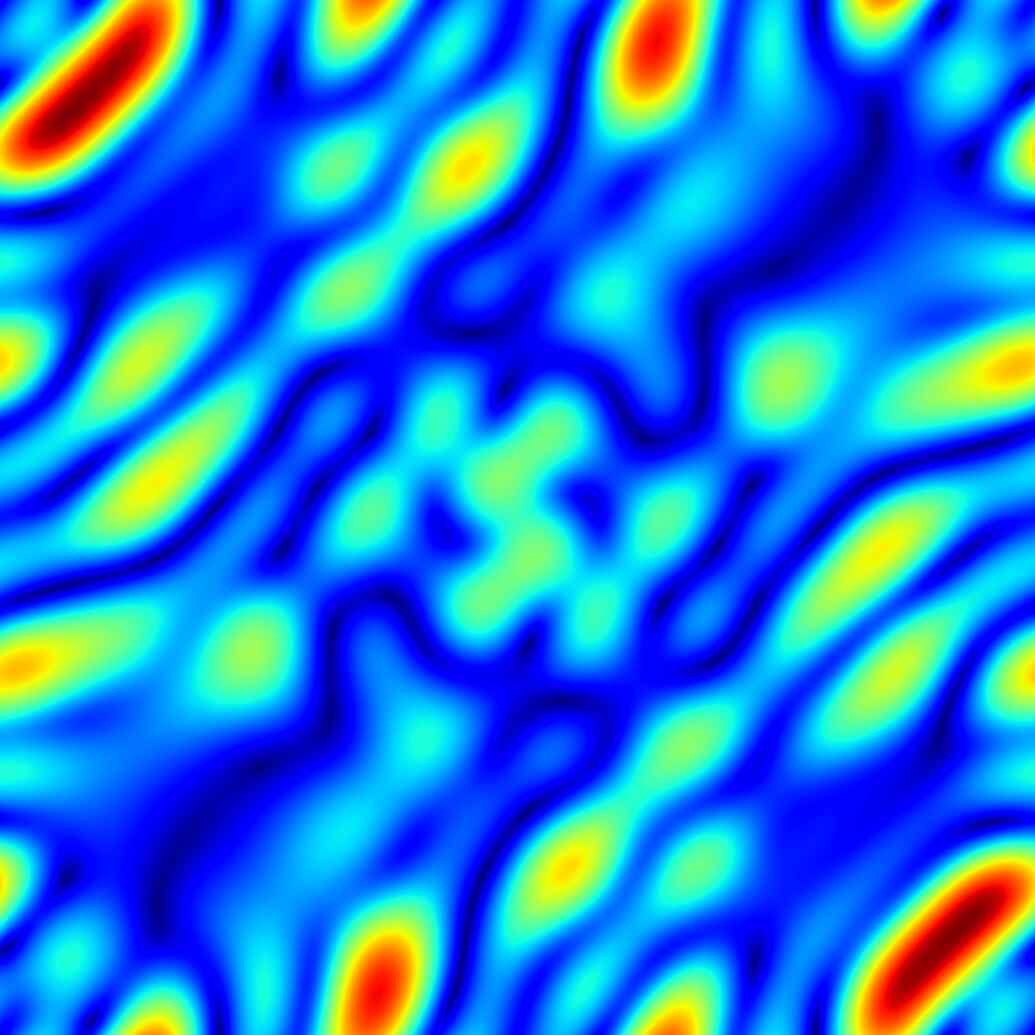}}
			& \fbox{\includegraphics[width=0.089\textwidth]{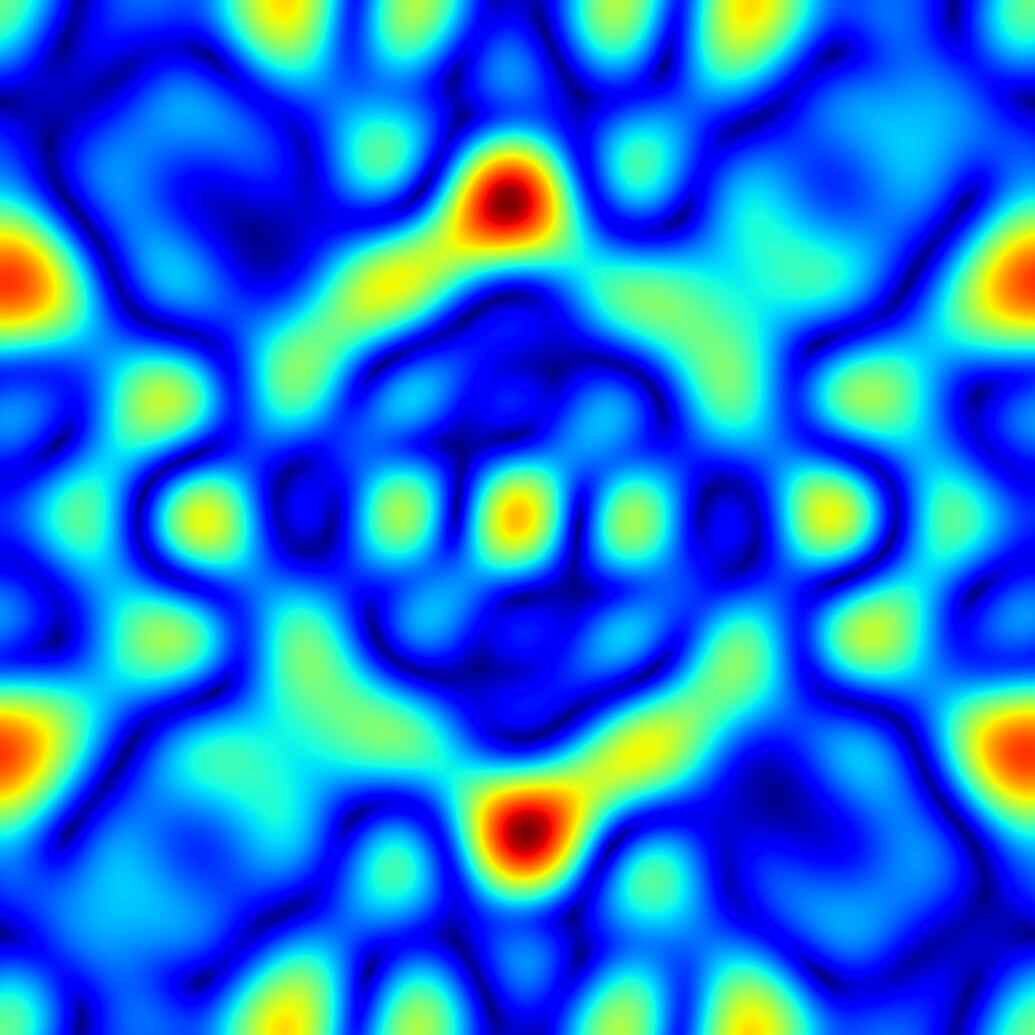}}
			& \fbox{\includegraphics[width=0.089\textwidth]{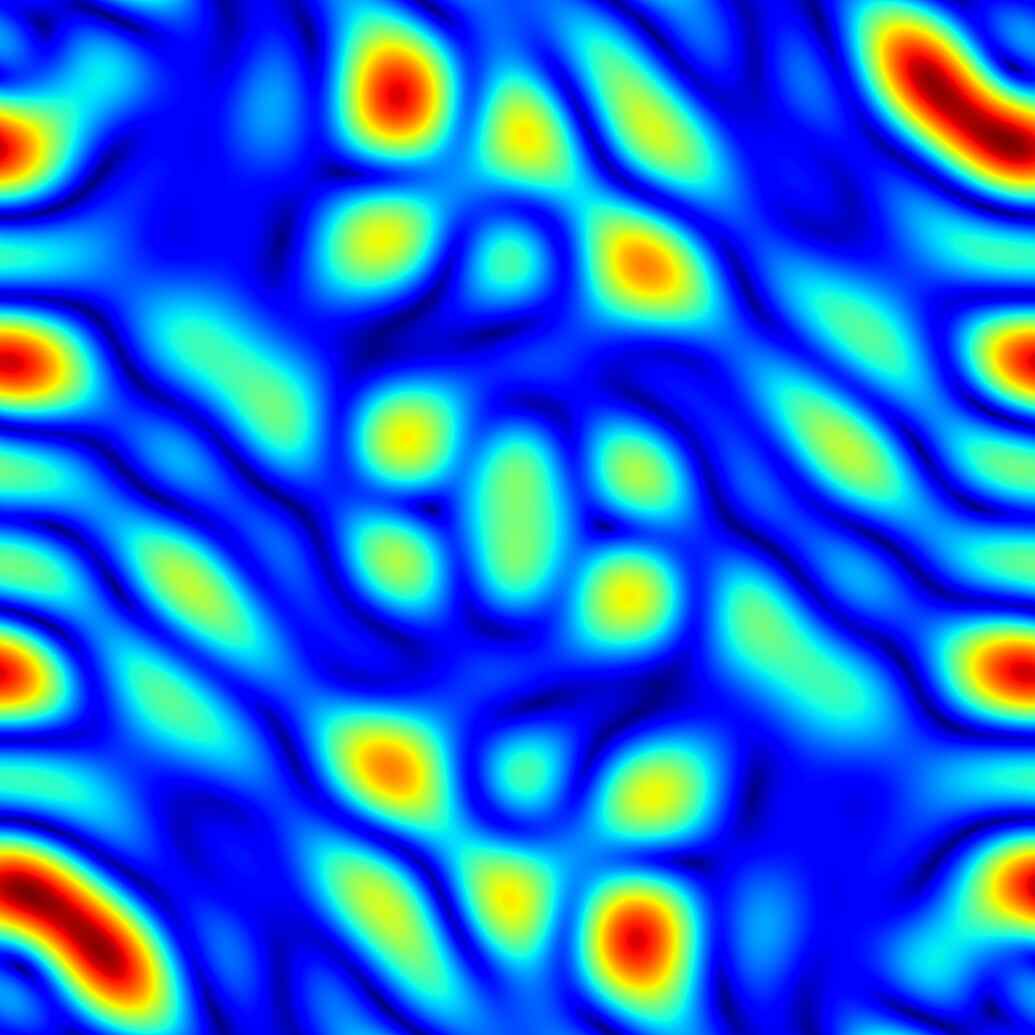}}
			& \fbox{\includegraphics[width=0.089\textwidth]{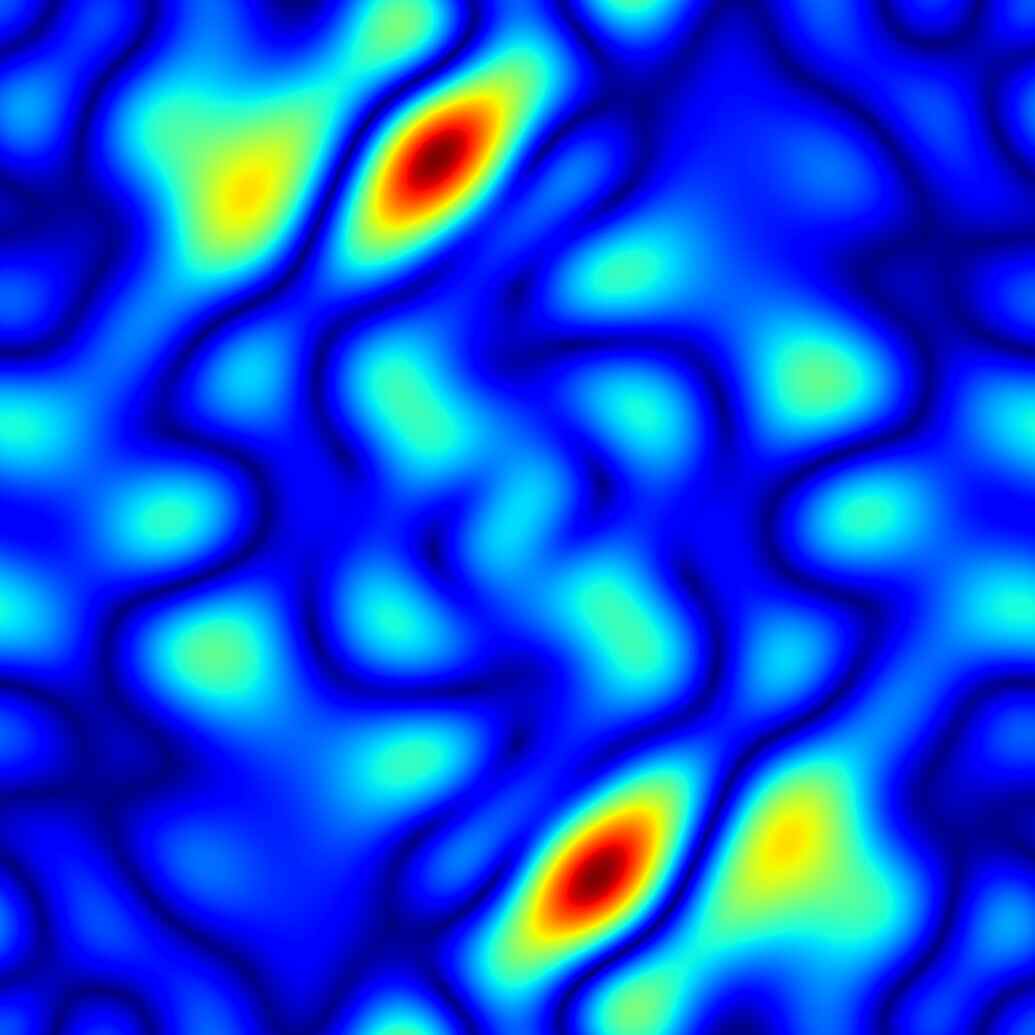}}
			& \fbox{\includegraphics[width=0.089\textwidth]{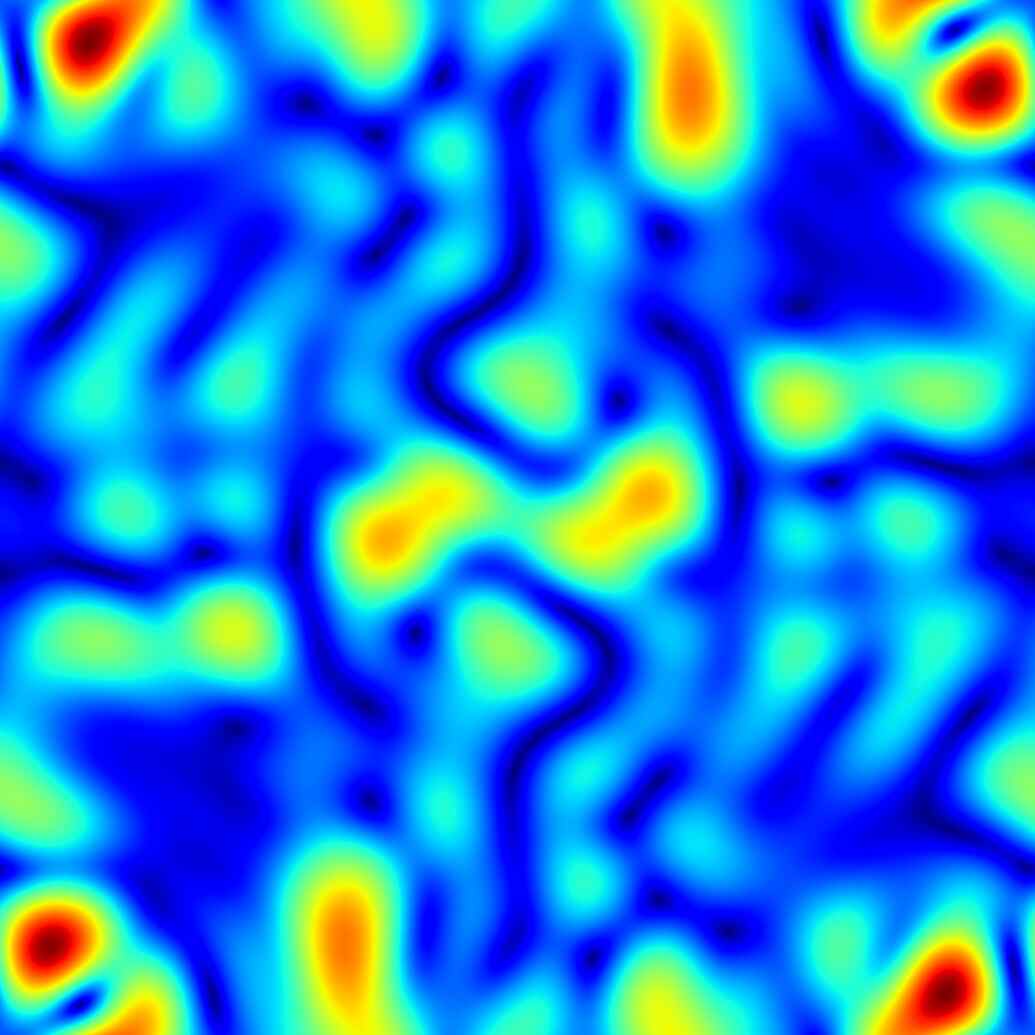}}
			& \fbox{\includegraphics[width=0.089\textwidth]{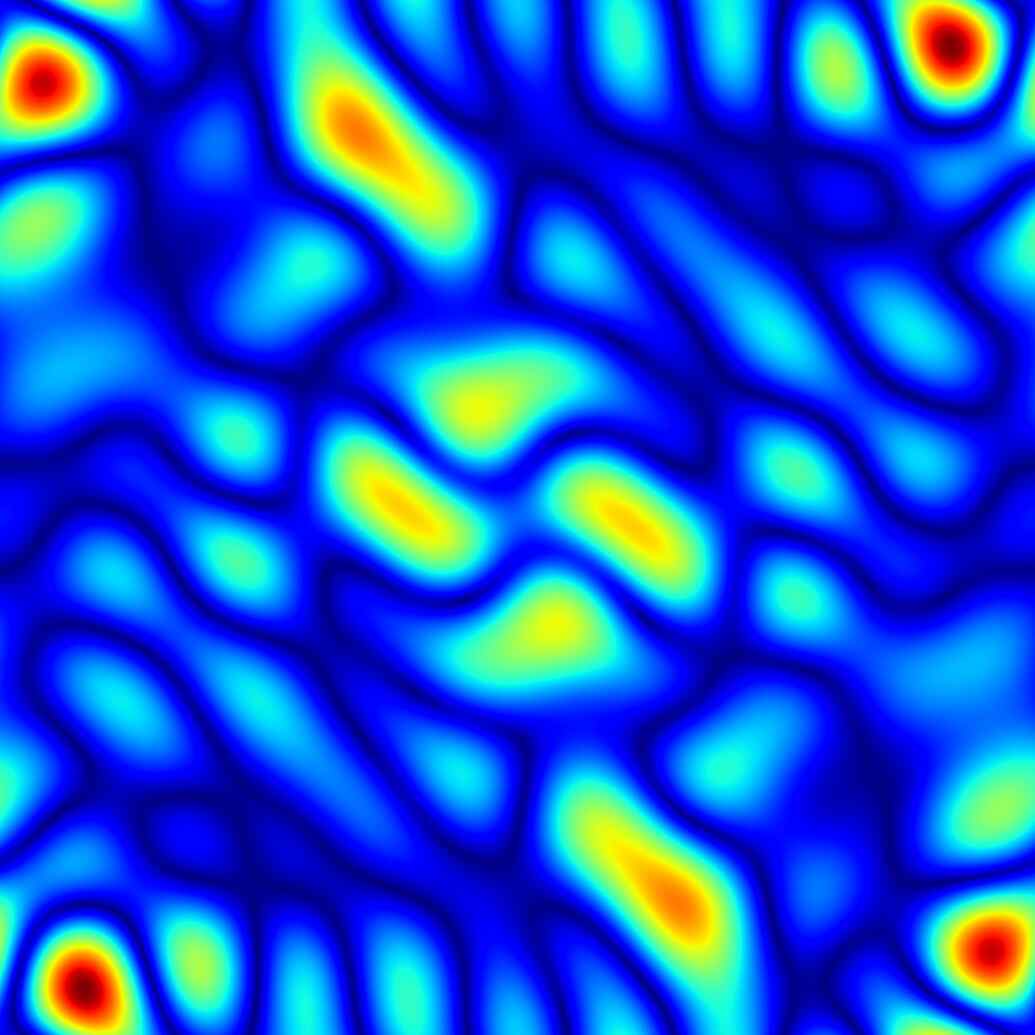}}
			& \fbox{\includegraphics[width=0.089\textwidth]{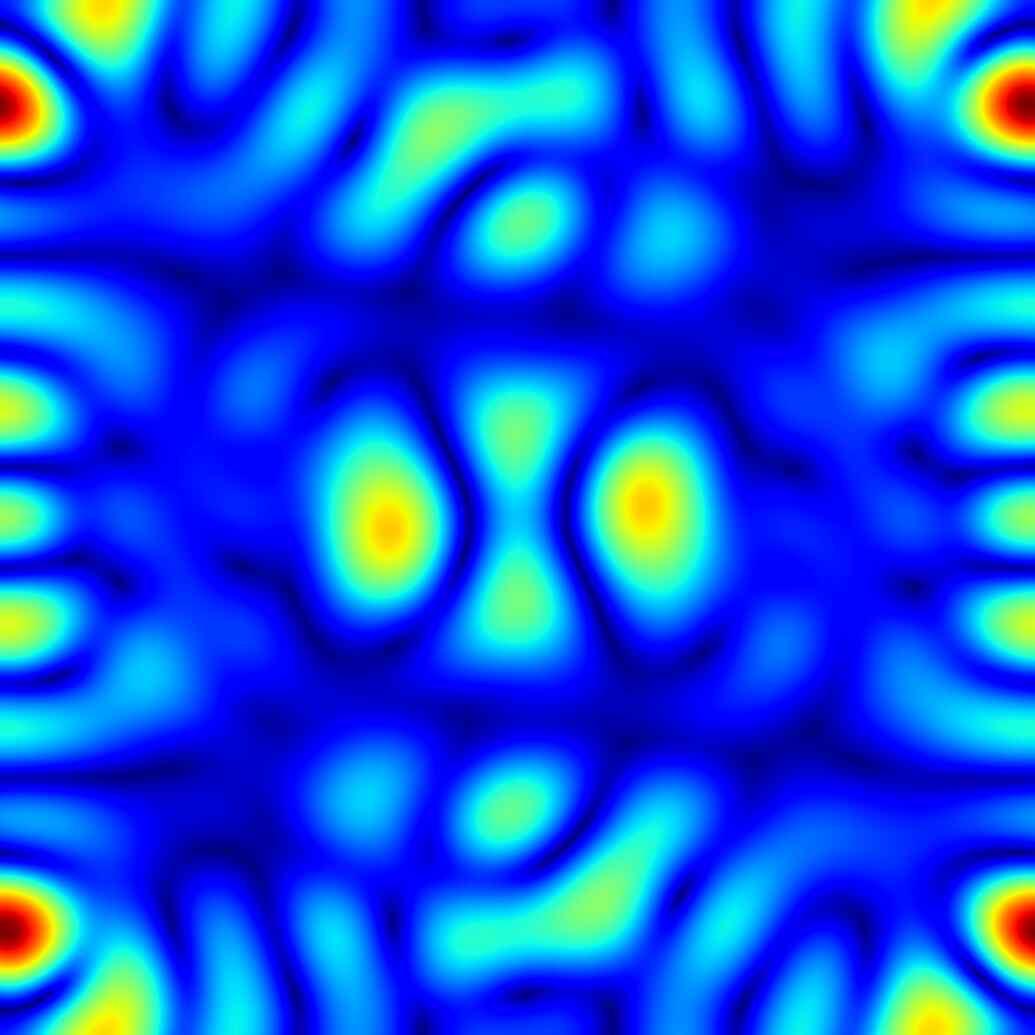}}
			& \fbox{\includegraphics[width=0.089\textwidth]{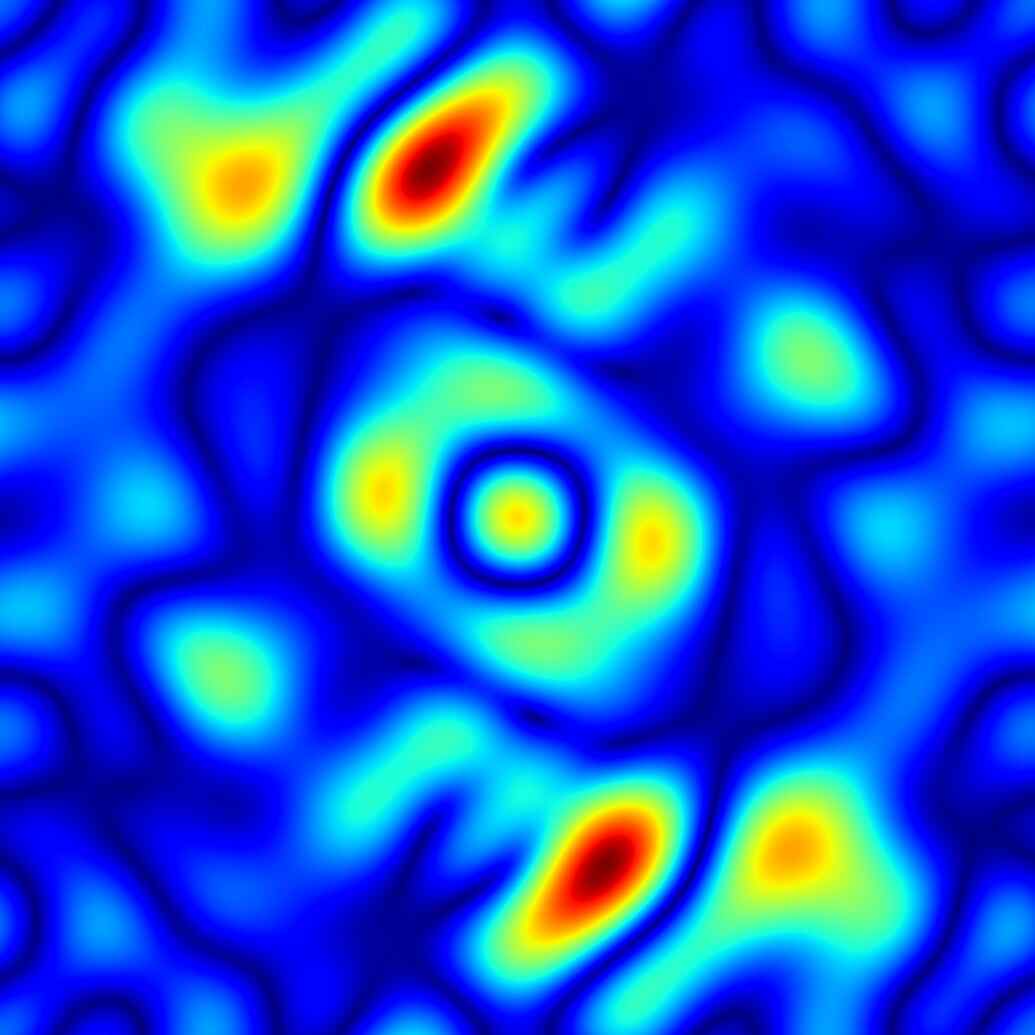}} \\

			\raisebox{19.5pt}{\fontsize{9pt}{10.5pt}\selectfont IN-1k\hspace{2pt}}
			& \fbox{\includegraphics[width=0.089\textwidth]{{figures/self_desc_viz/in1k_1}}}
			& \fbox{\includegraphics[width=0.089\textwidth]{{figures/self_desc_viz/in1k_2}}}
			& \fbox{\includegraphics[width=0.089\textwidth]{{figures/self_desc_viz/in1k_4}}}
			& \fbox{\includegraphics[width=0.089\textwidth]{{figures/self_desc_viz/in1k_7}}}
			& \fbox{\includegraphics[width=0.089\textwidth]{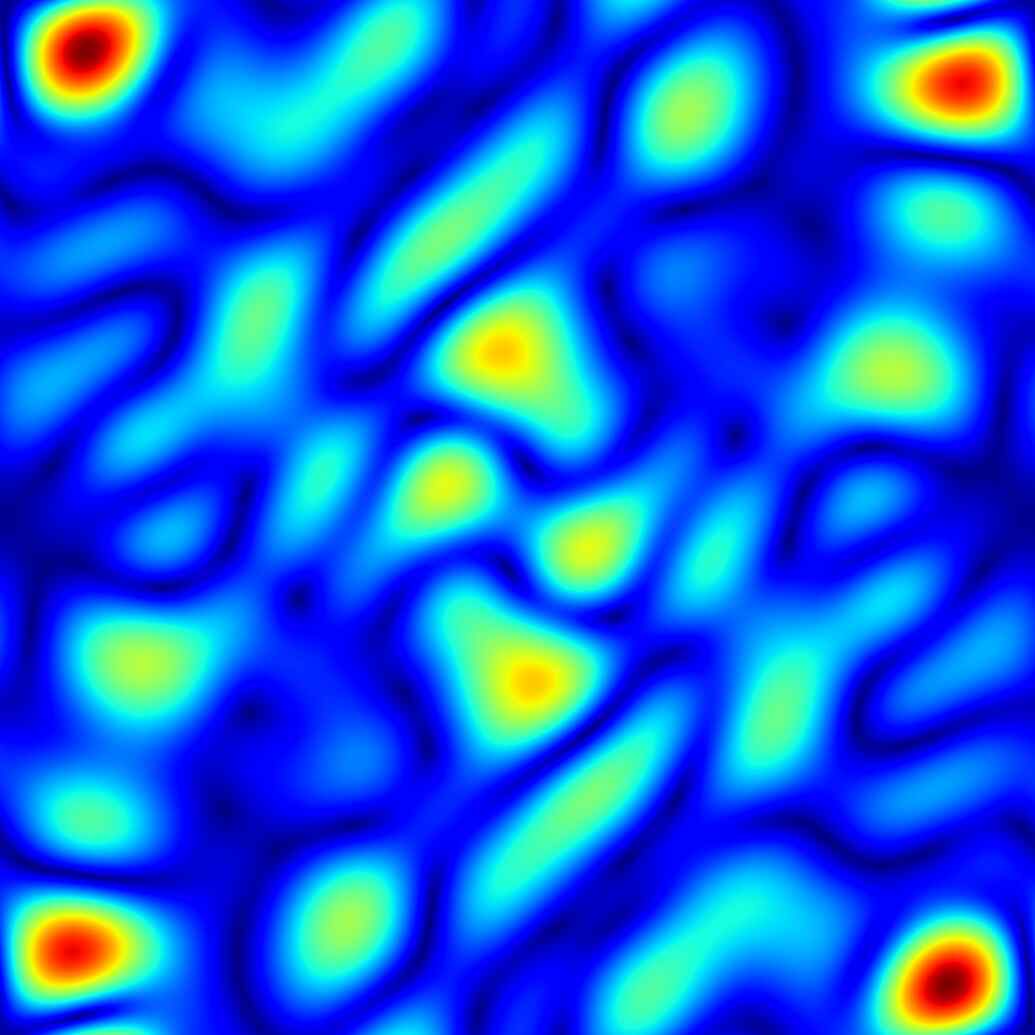}}
			& \fbox{\includegraphics[width=0.089\textwidth]{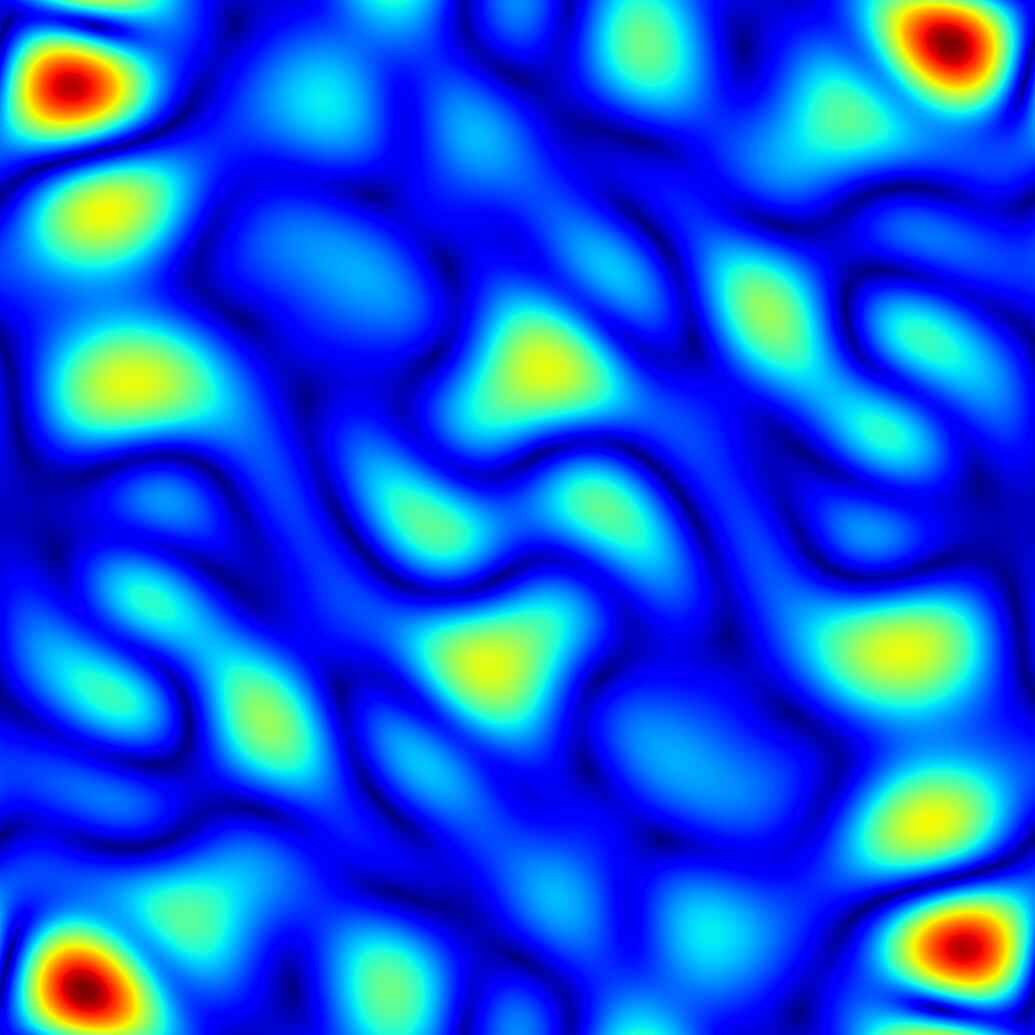}}
			& \fbox{\includegraphics[width=0.089\textwidth]{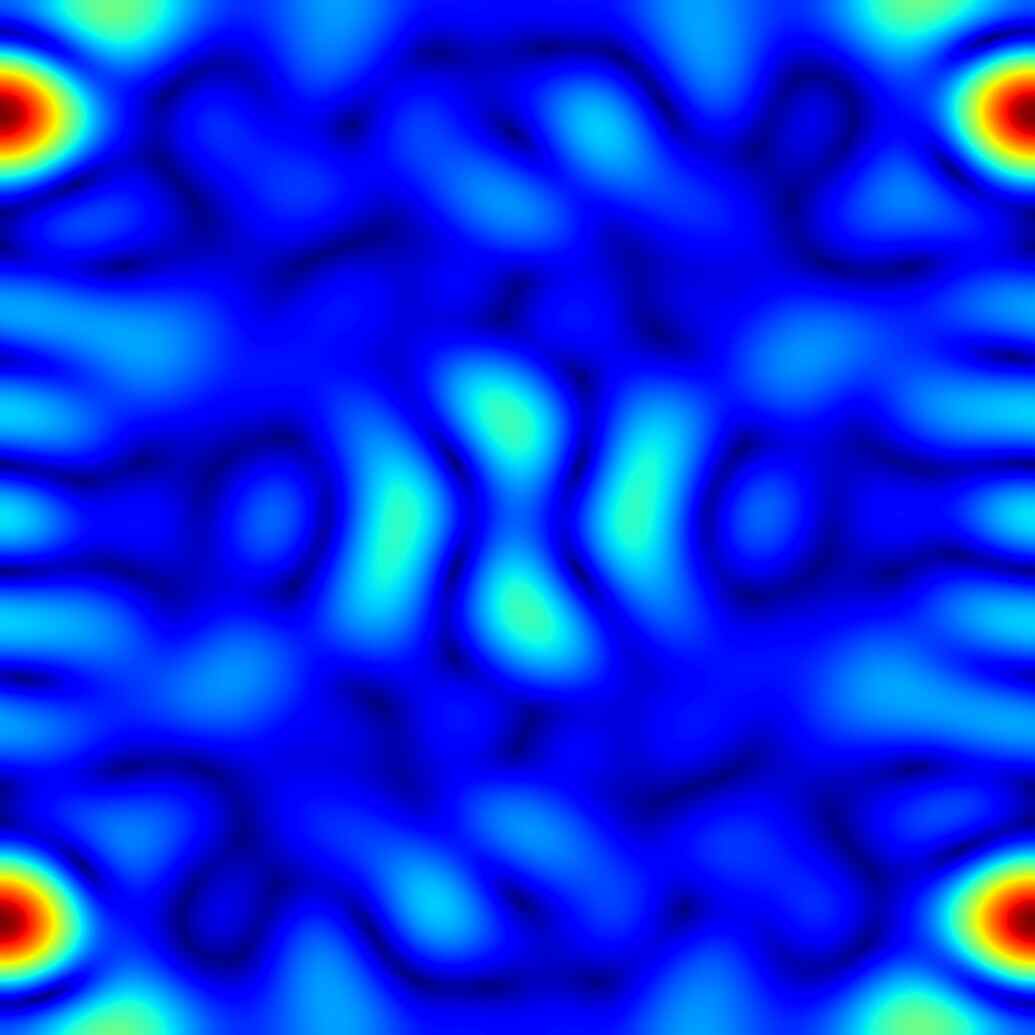}}
			& \fbox{\includegraphics[width=0.089\textwidth]{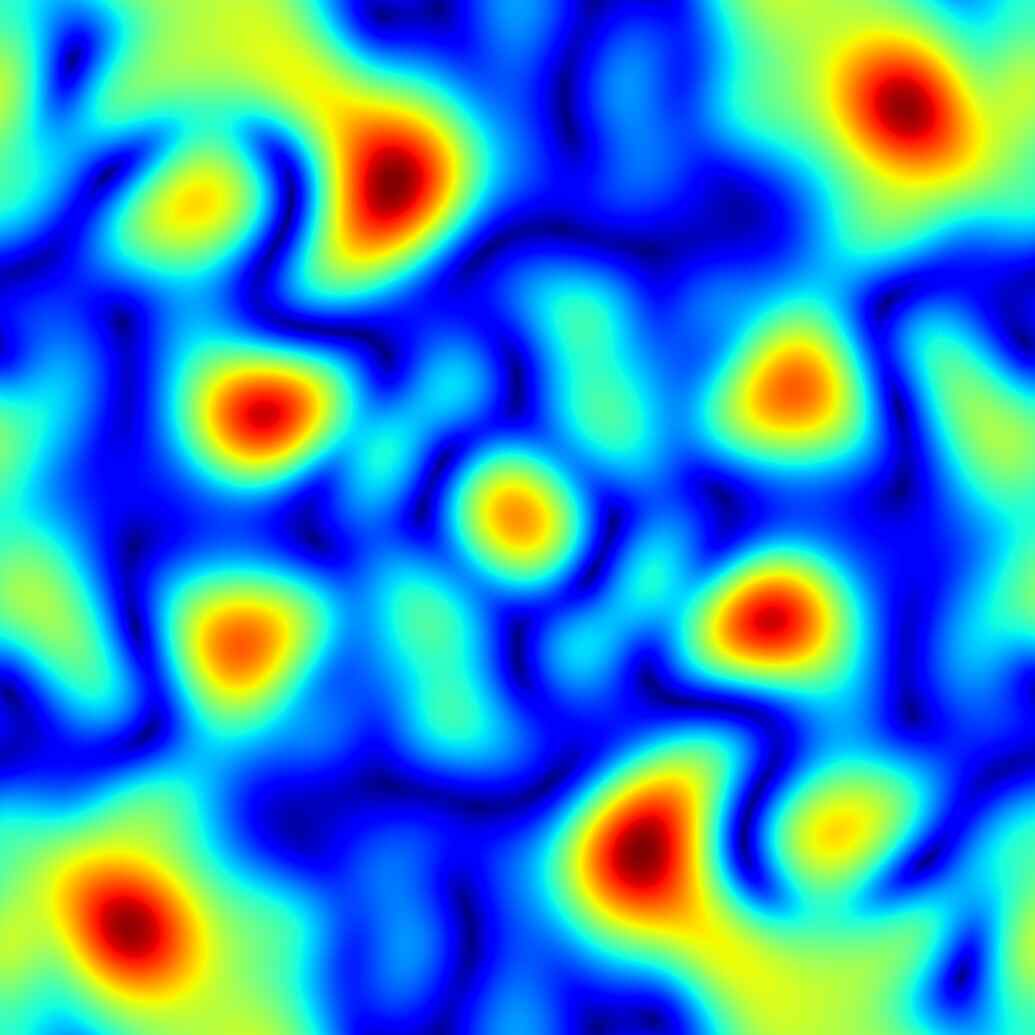}} \\

			\raisebox{19.5pt}{\fontsize{9pt}{10.5pt}\selectfont IN-22k\hspace{2pt}}
			& \fbox{\includegraphics[width=0.089\textwidth]{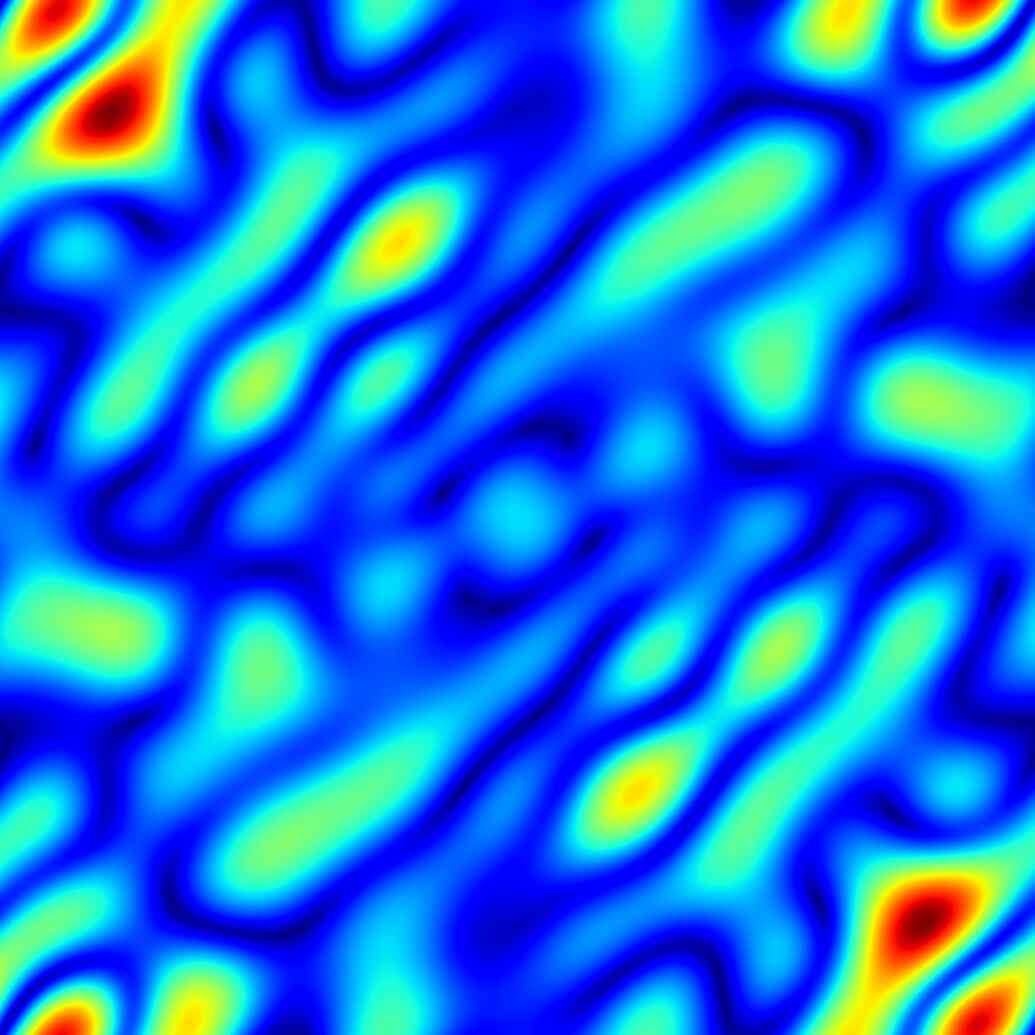}}
			& \fbox{\includegraphics[width=0.089\textwidth]{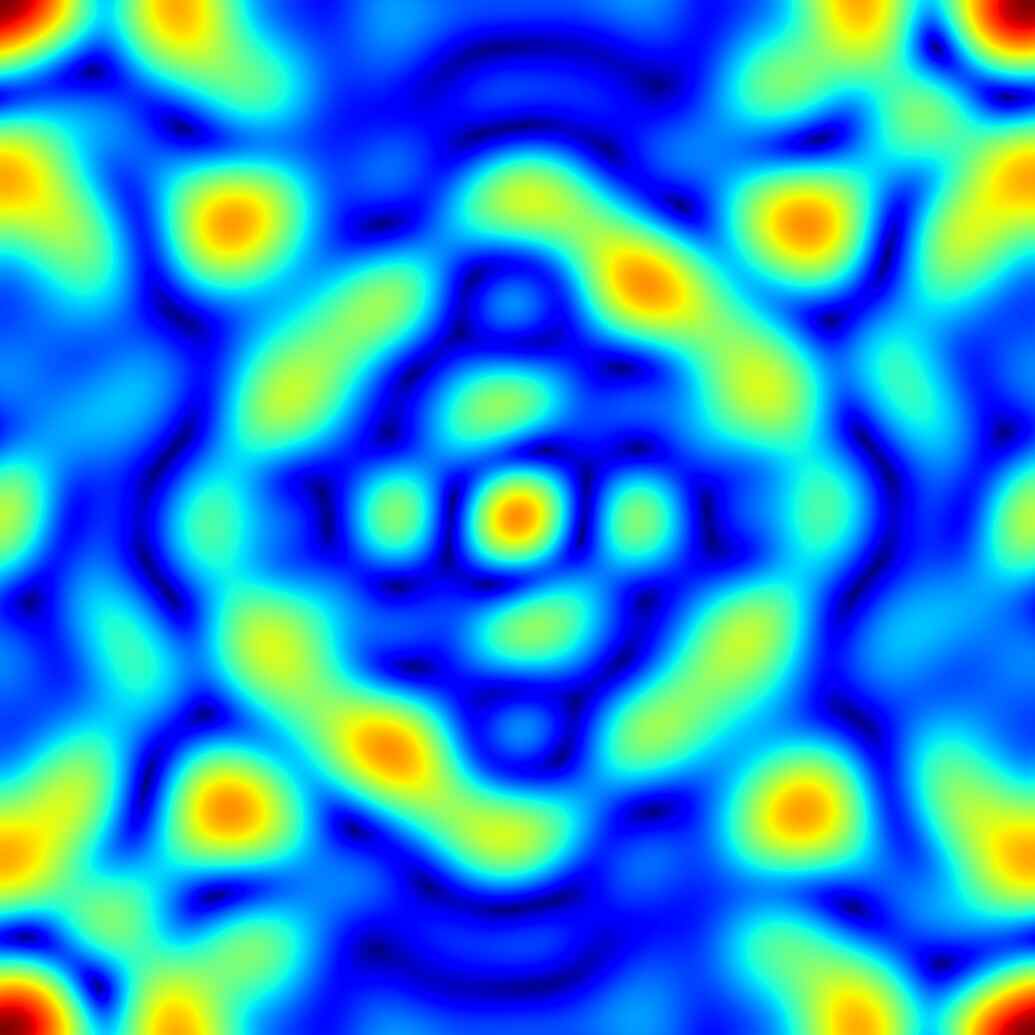}}
			& \fbox{\includegraphics[width=0.089\textwidth]{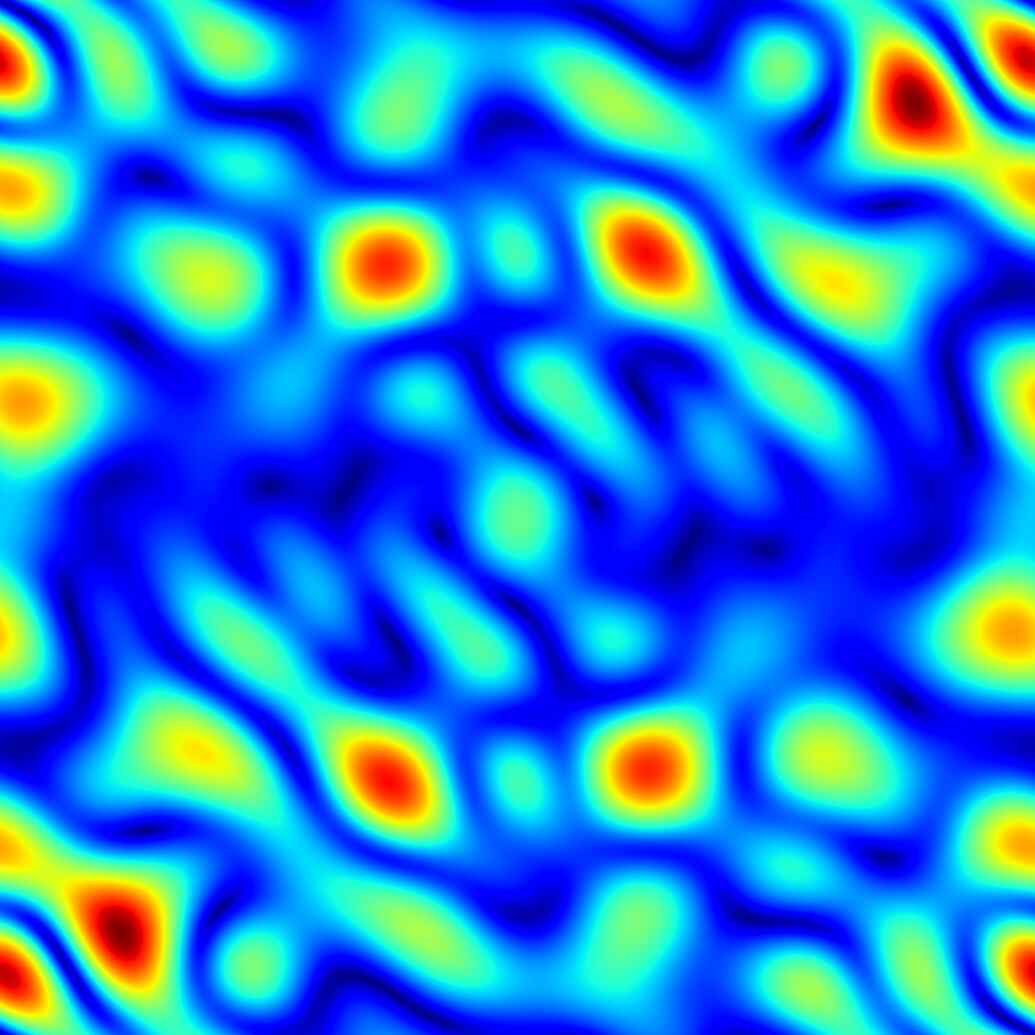}}
			& \fbox{\includegraphics[width=0.089\textwidth]{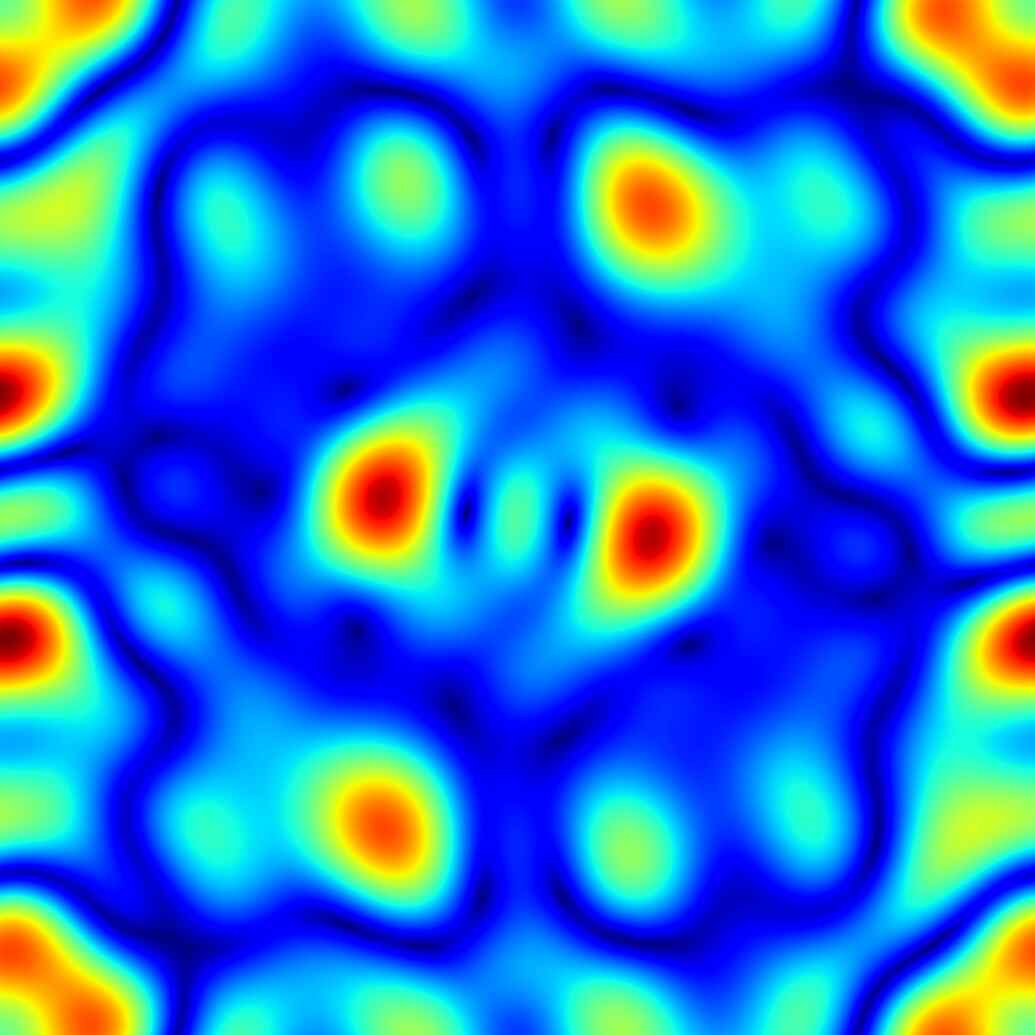}}
			& \fbox{\includegraphics[width=0.089\textwidth]{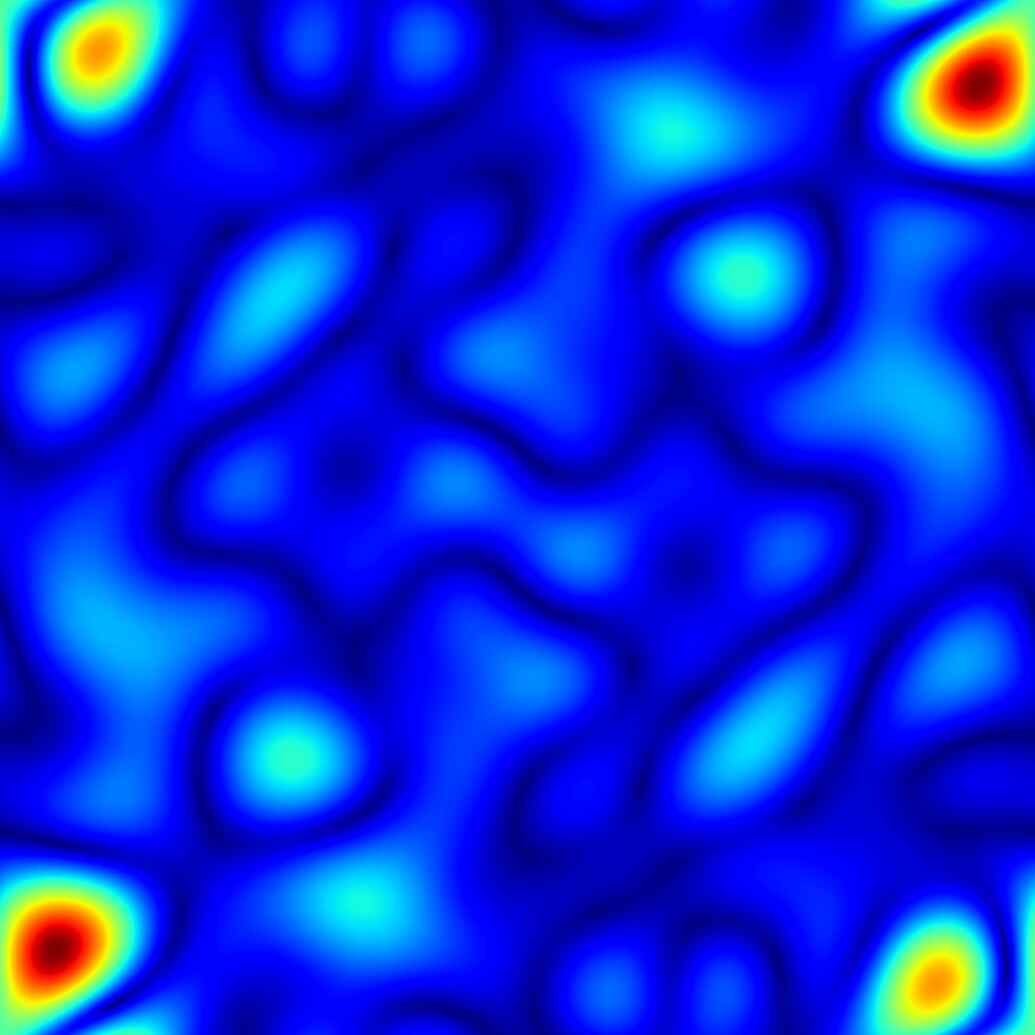}}
			& \fbox{\includegraphics[width=0.089\textwidth]{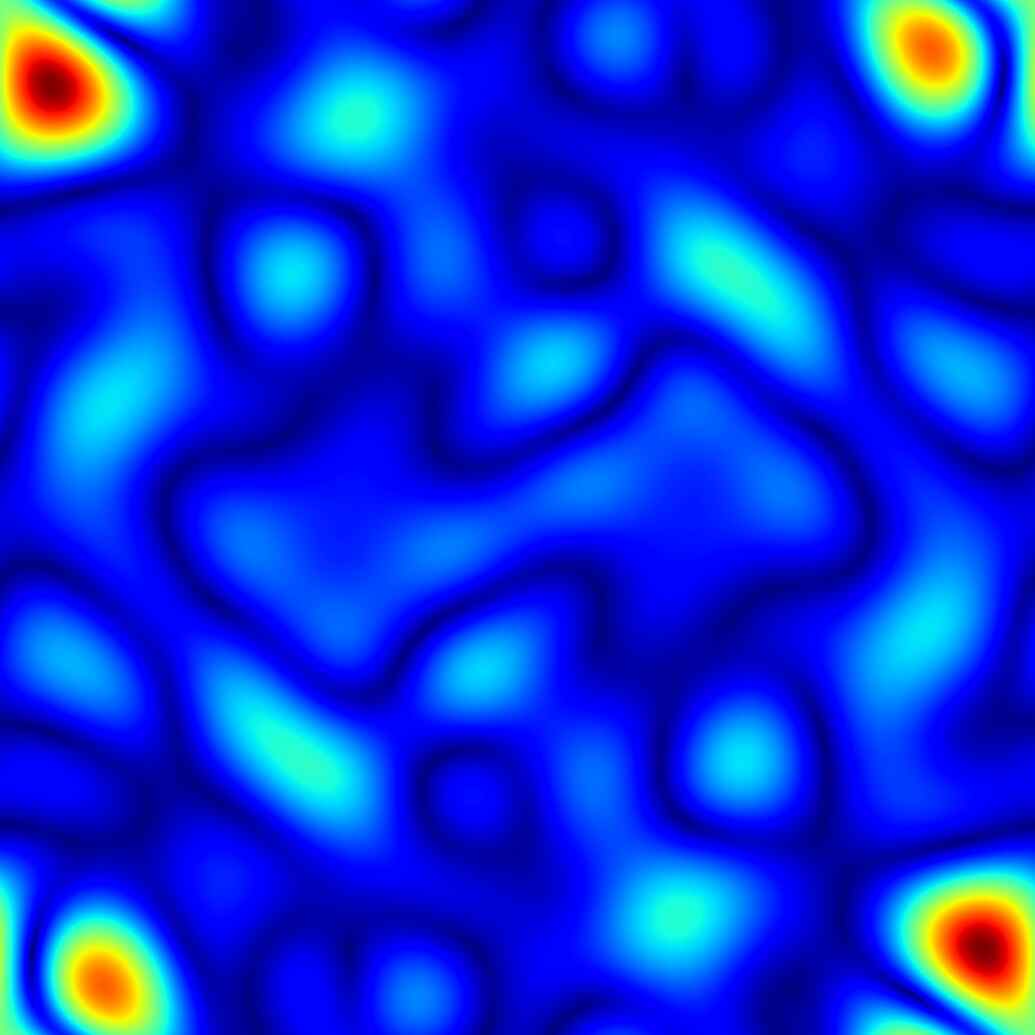}}
			& \fbox{\includegraphics[width=0.089\textwidth]{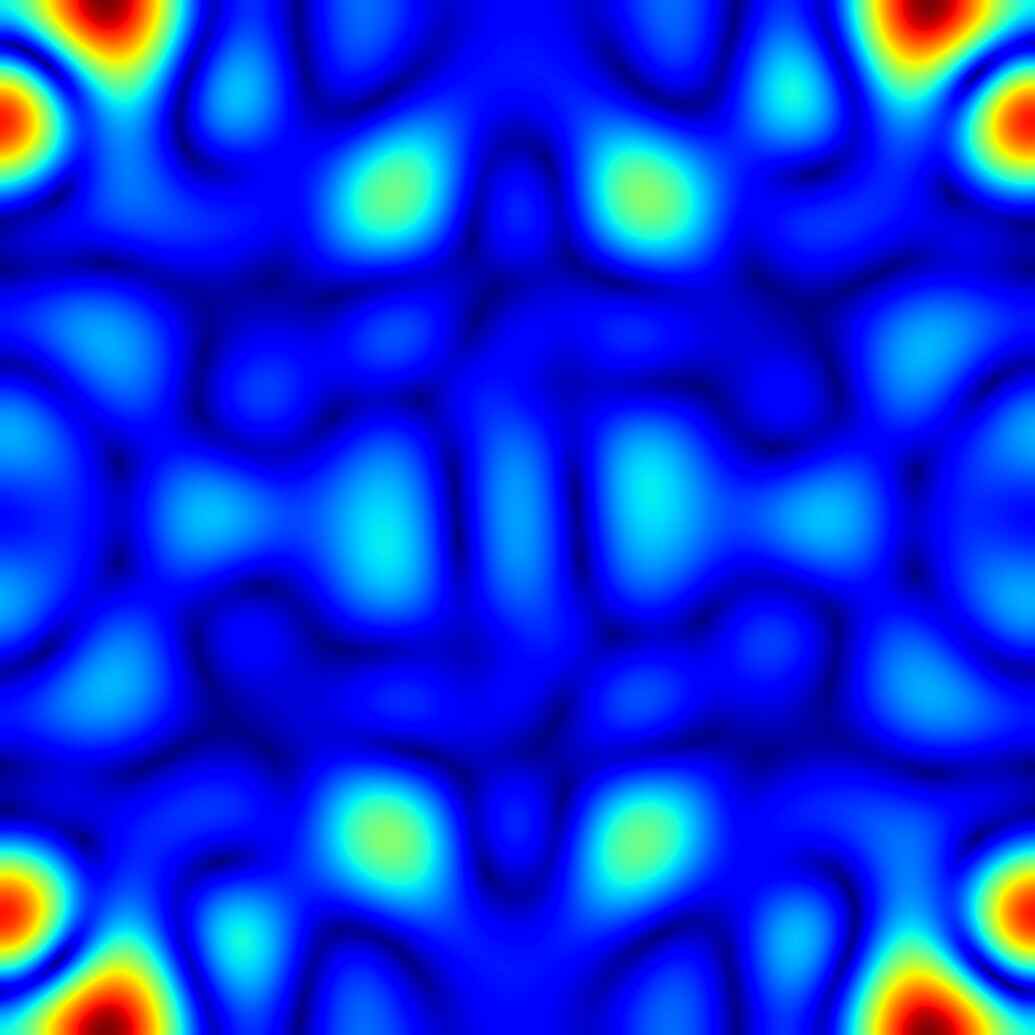}}
			& \fbox{\includegraphics[width=0.089\textwidth]{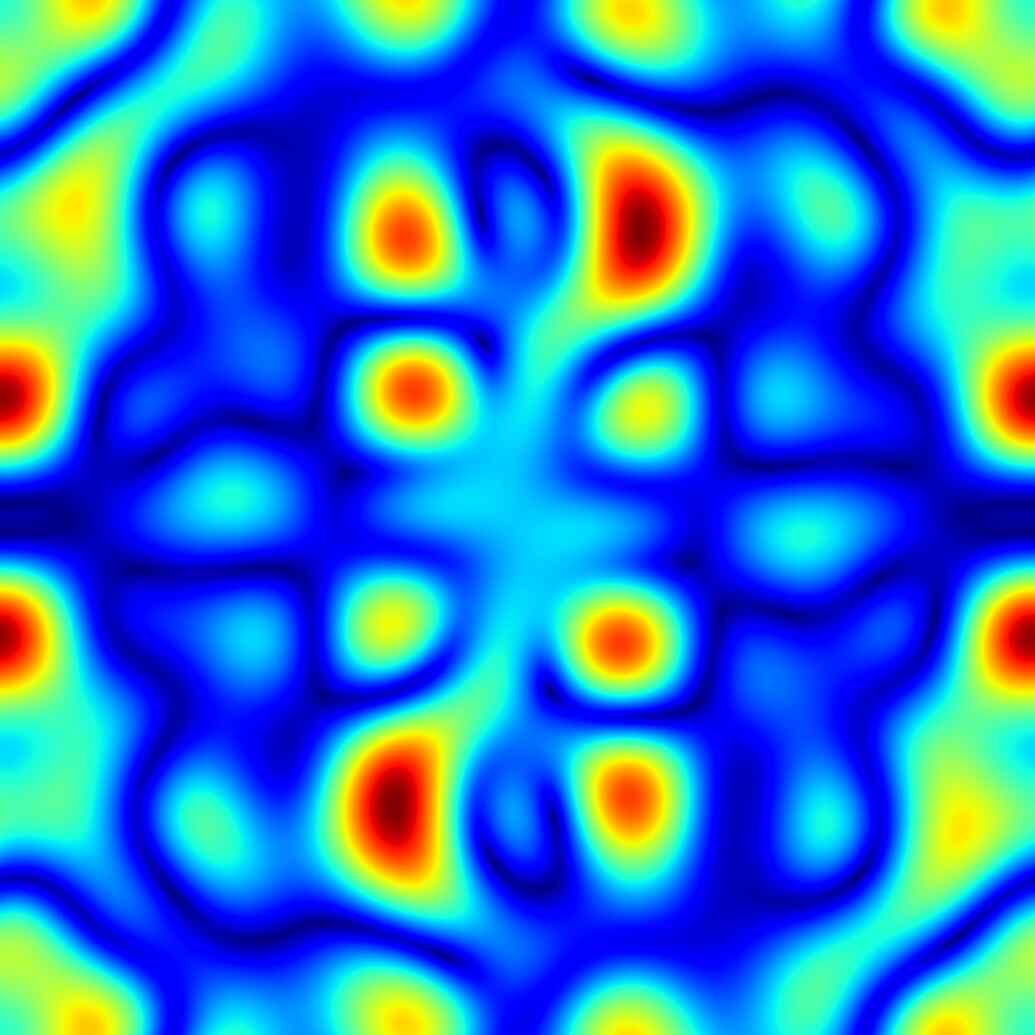}} \\

			\raisebox{19.5pt}{\fontsize{9pt}{10.5pt}\selectfont MIDB\hspace{2pt}}
			& \fbox{\includegraphics[width=0.089\textwidth]{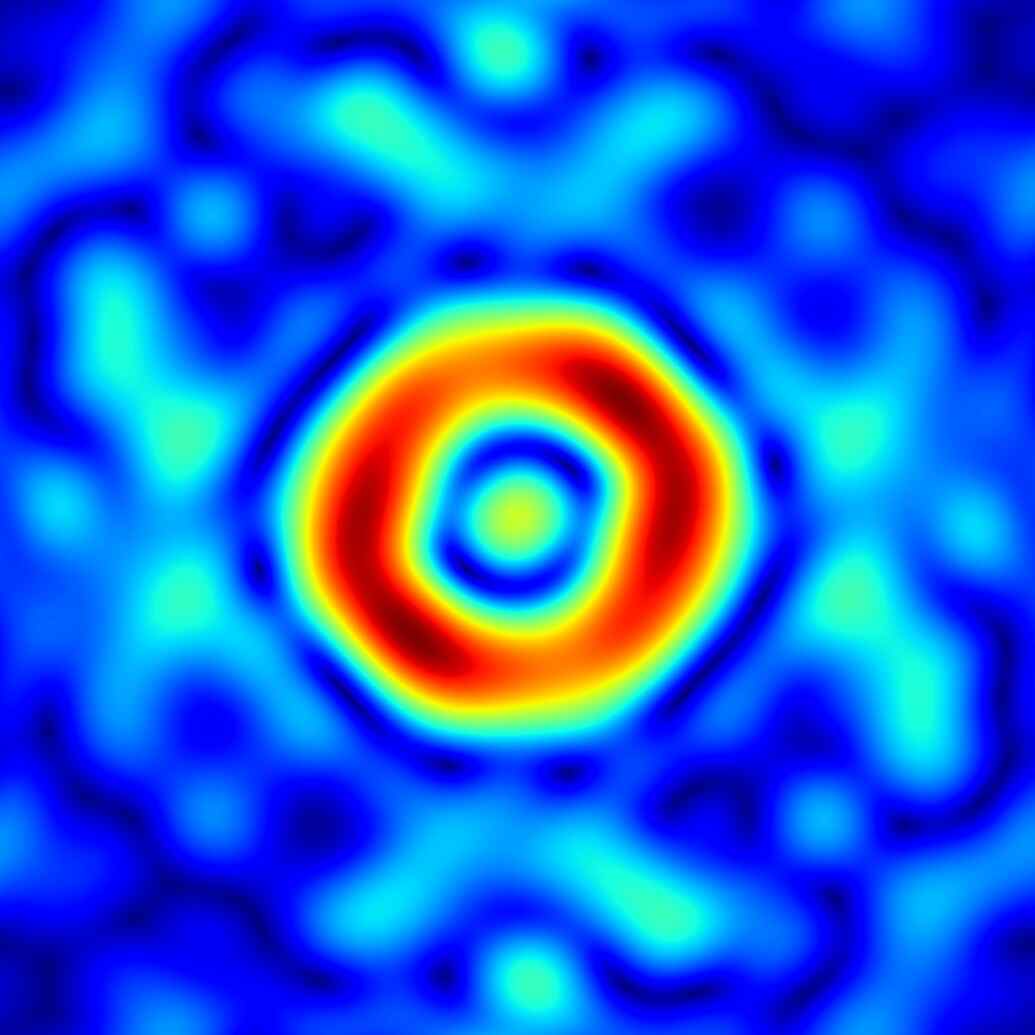}}
			& \fbox{\includegraphics[width=0.089\textwidth]{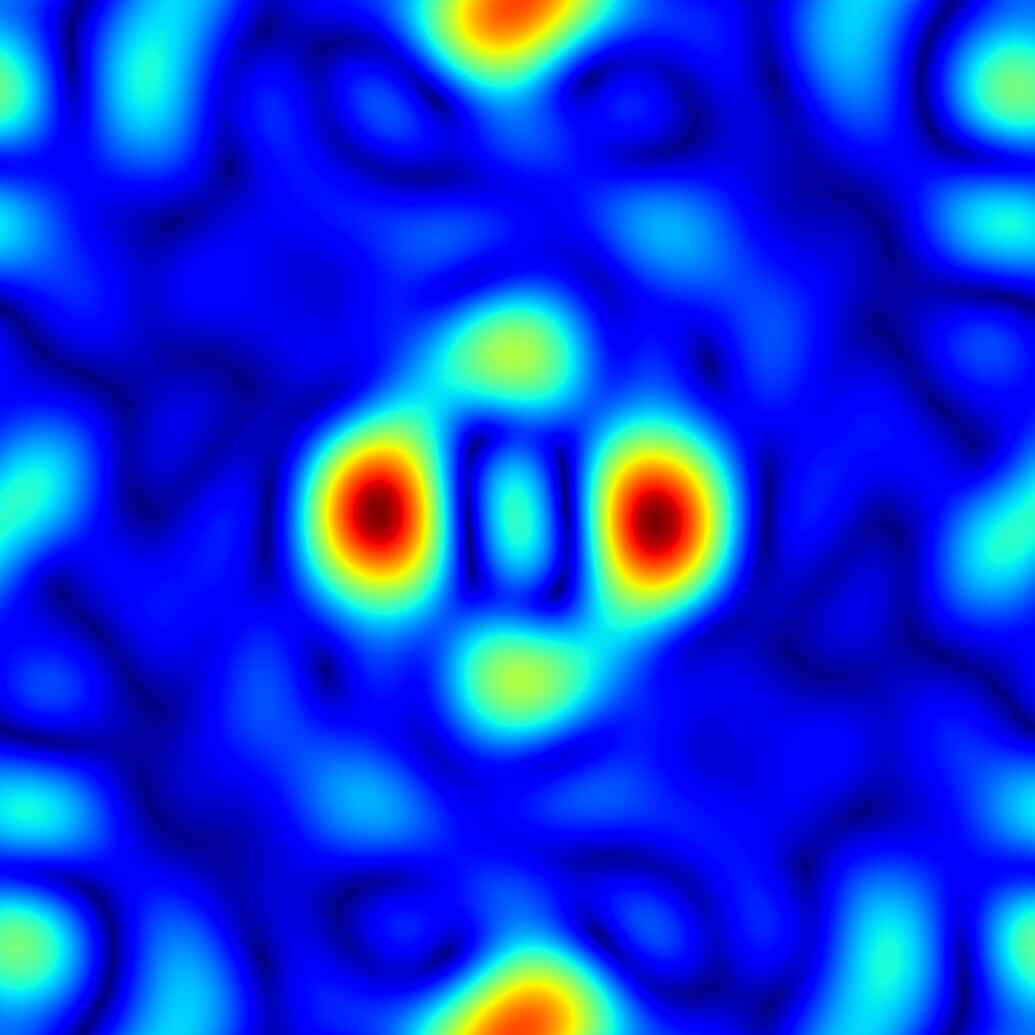}}
			& \fbox{\includegraphics[width=0.089\textwidth]{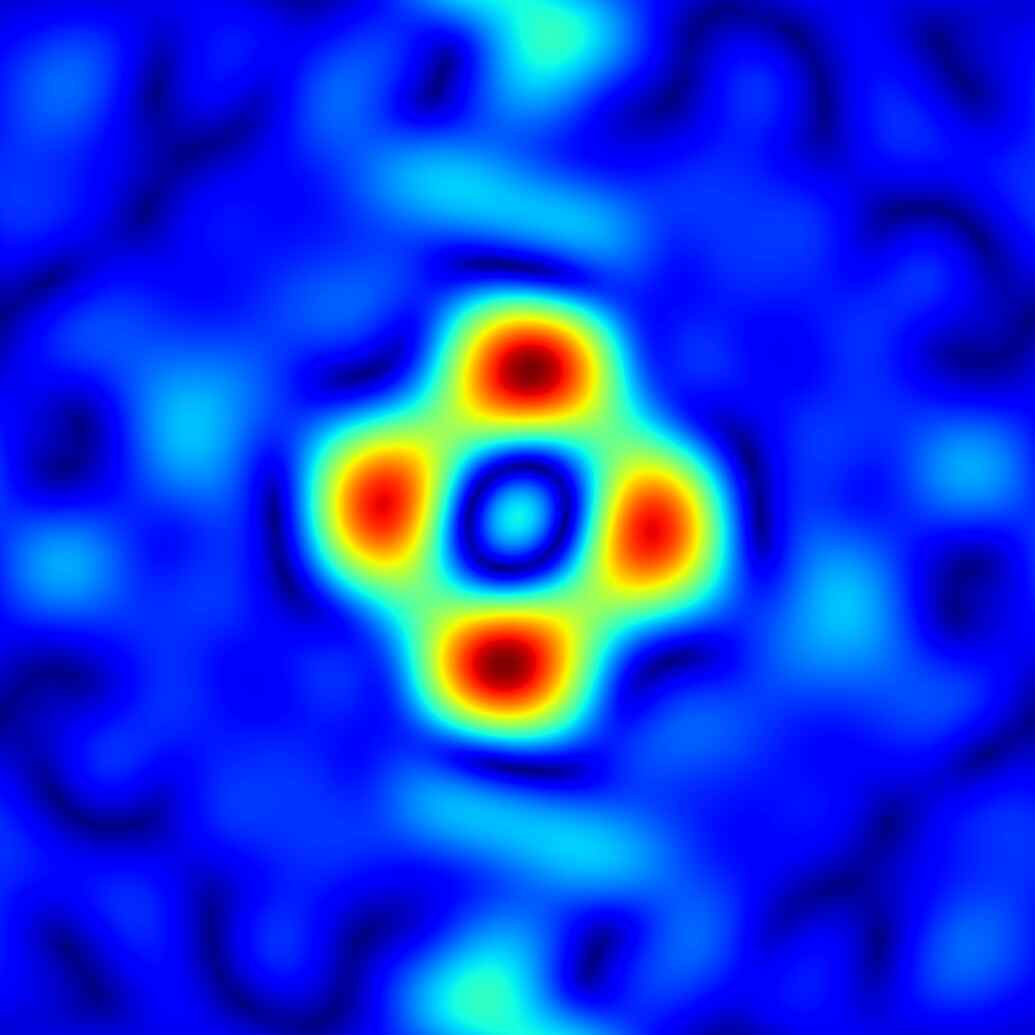}}
			& \fbox{\includegraphics[width=0.089\textwidth]{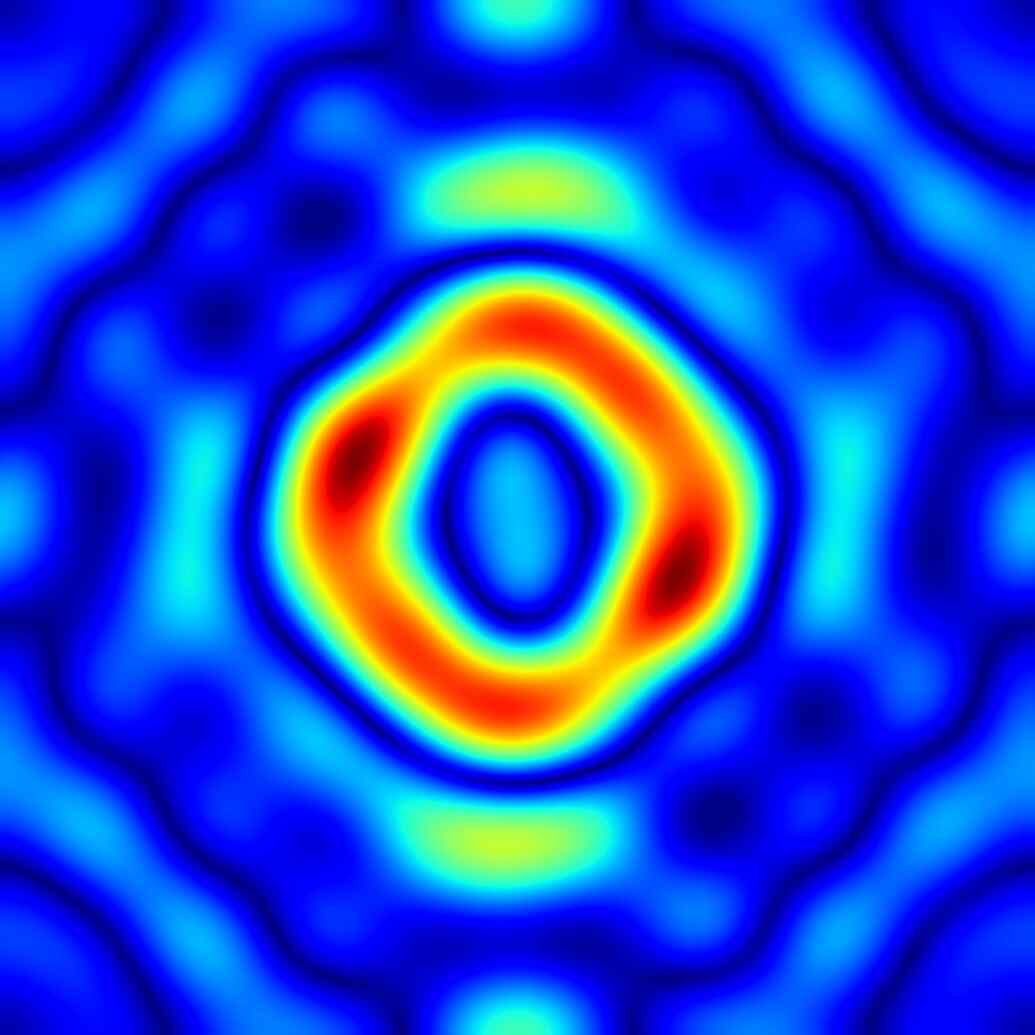}}
			& \fbox{\includegraphics[width=0.089\textwidth]{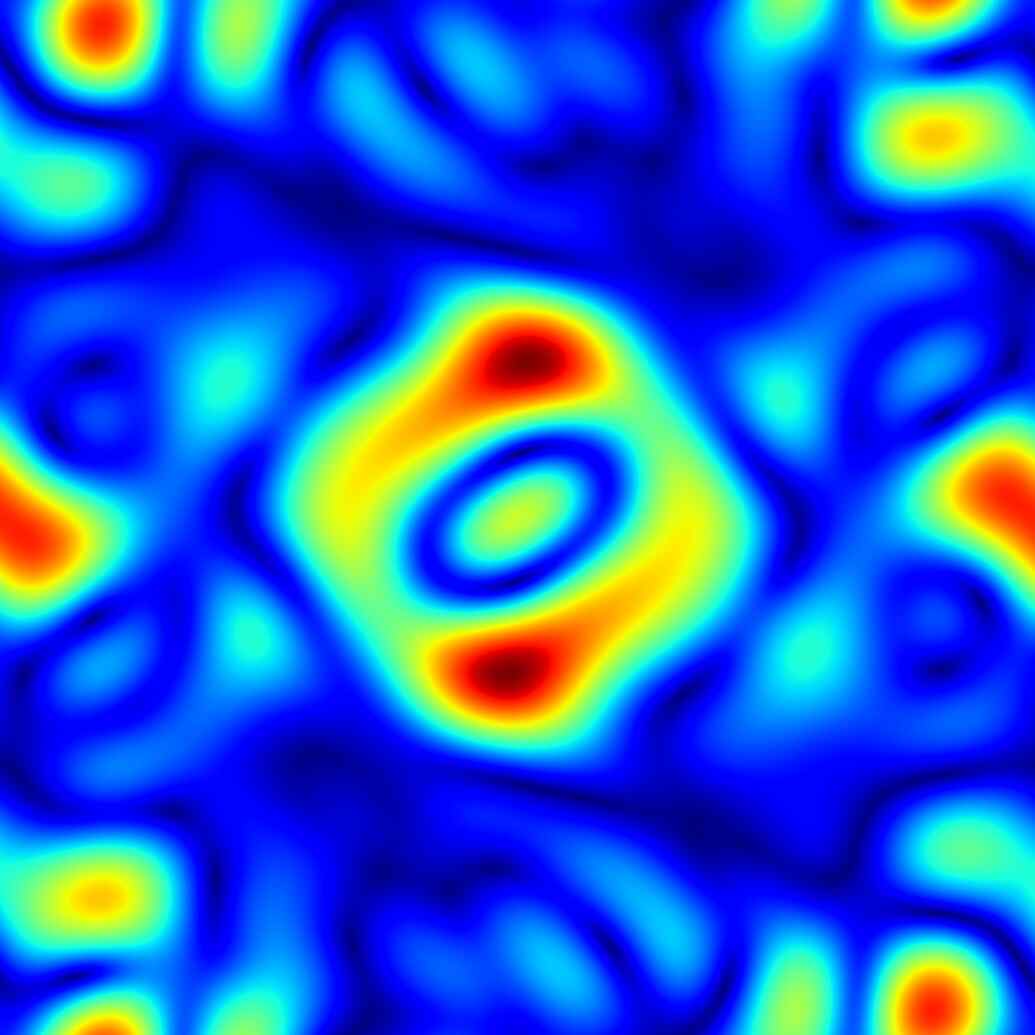}}
			& \fbox{\includegraphics[width=0.089\textwidth]{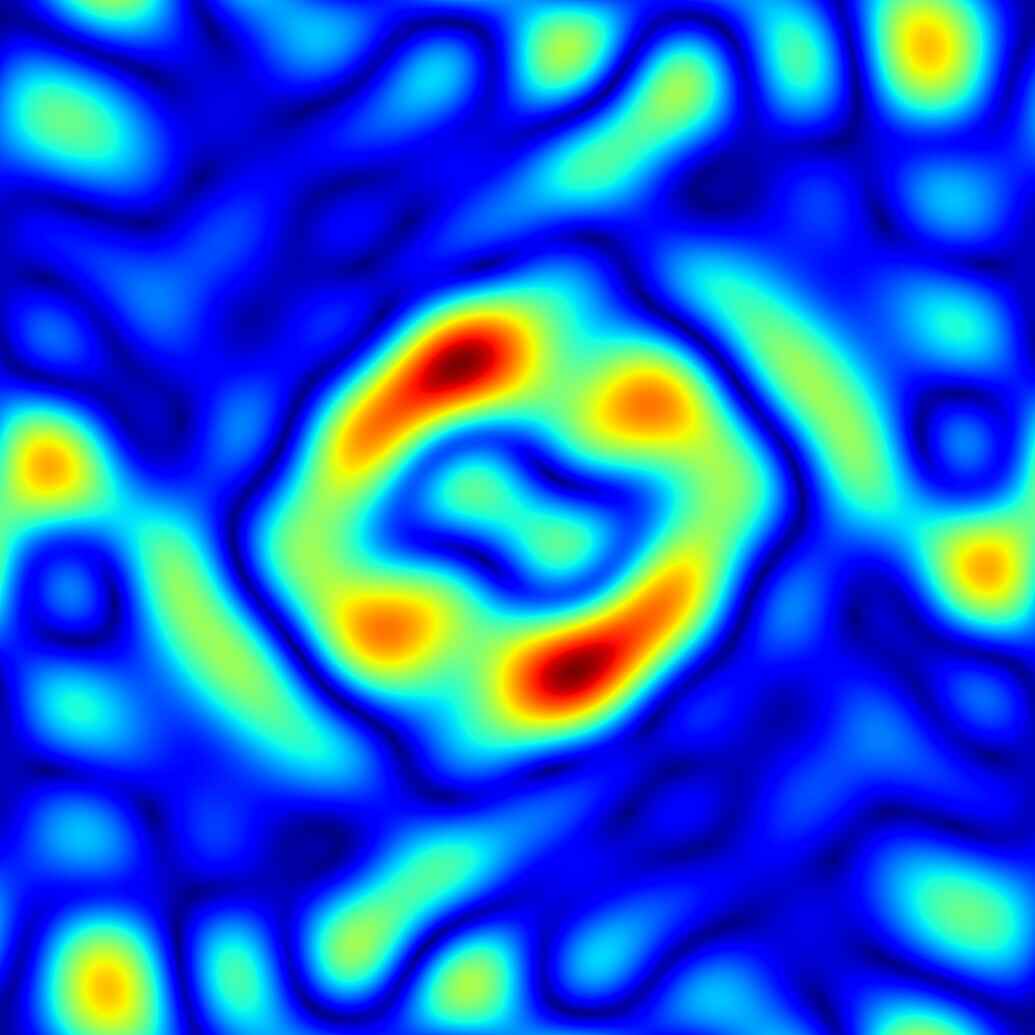}}
			& \fbox{\includegraphics[width=0.089\textwidth]{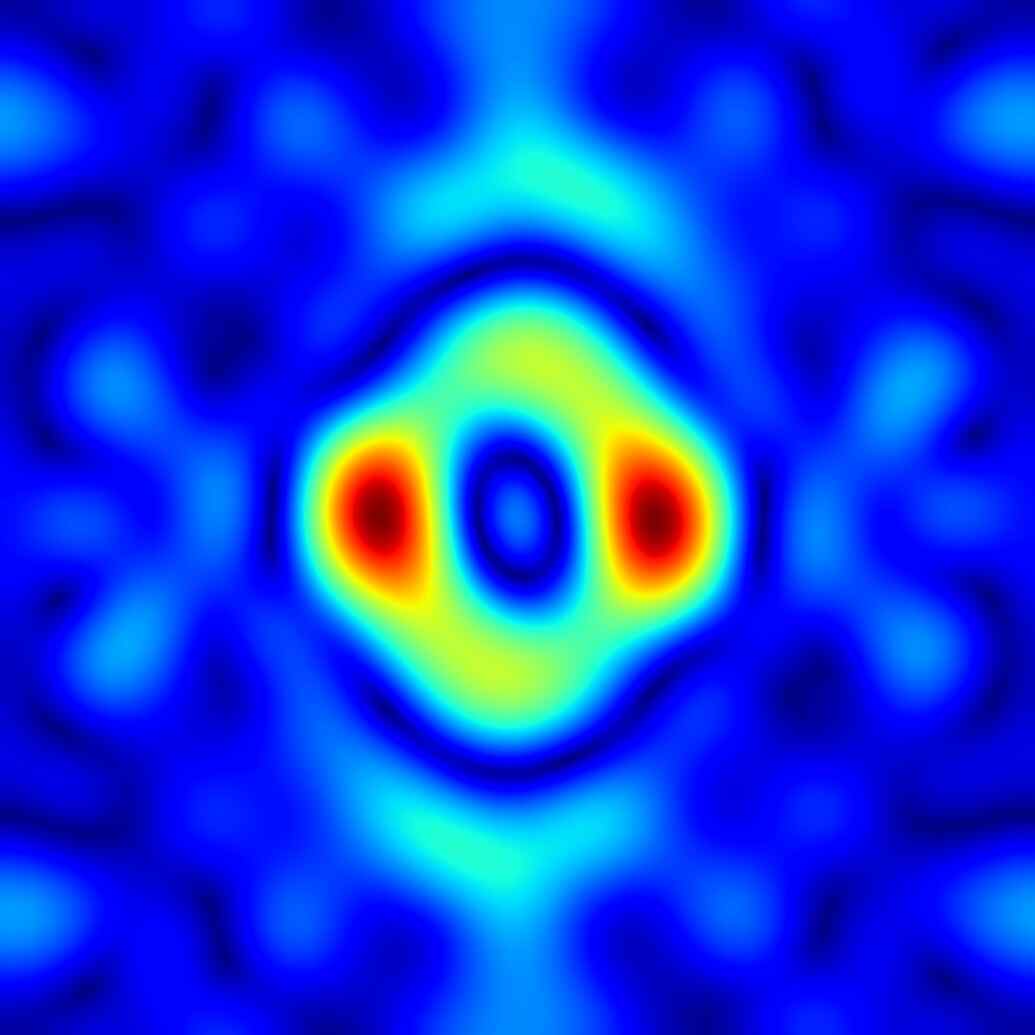}}
			& \fbox{\includegraphics[width=0.089\textwidth]{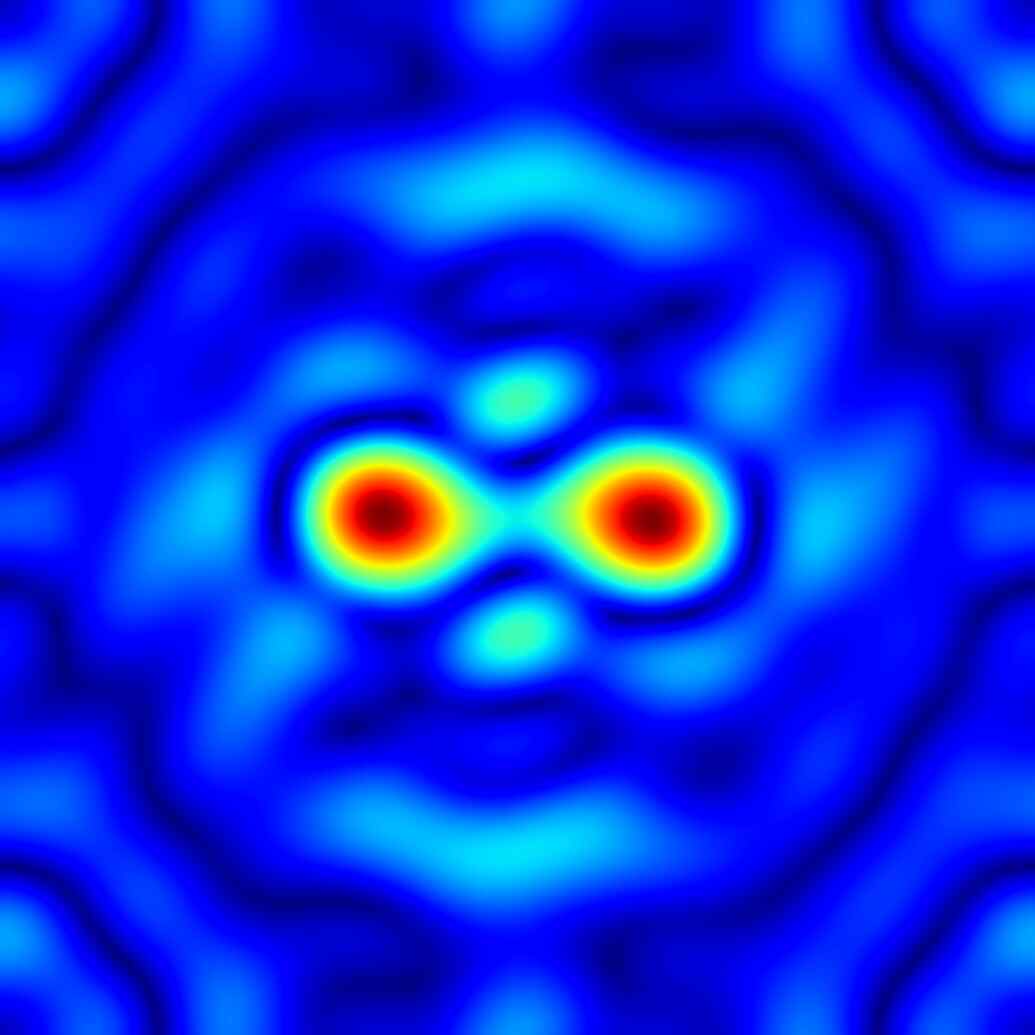}} \\

		\end{tabular}
	}

	\caption{Visualization of the average power spectrum of different filters in the forensic self-descriptions obtained from four real datasets.}
	\label{fig:spec_of_reals}
	\pulluppp
\end{figure*}

%% file: figures/ZeroShotAccVsThresholds.tex
\begin{figure}[!t]
	\centering
	\includegraphics[width=1.0\linewidth]{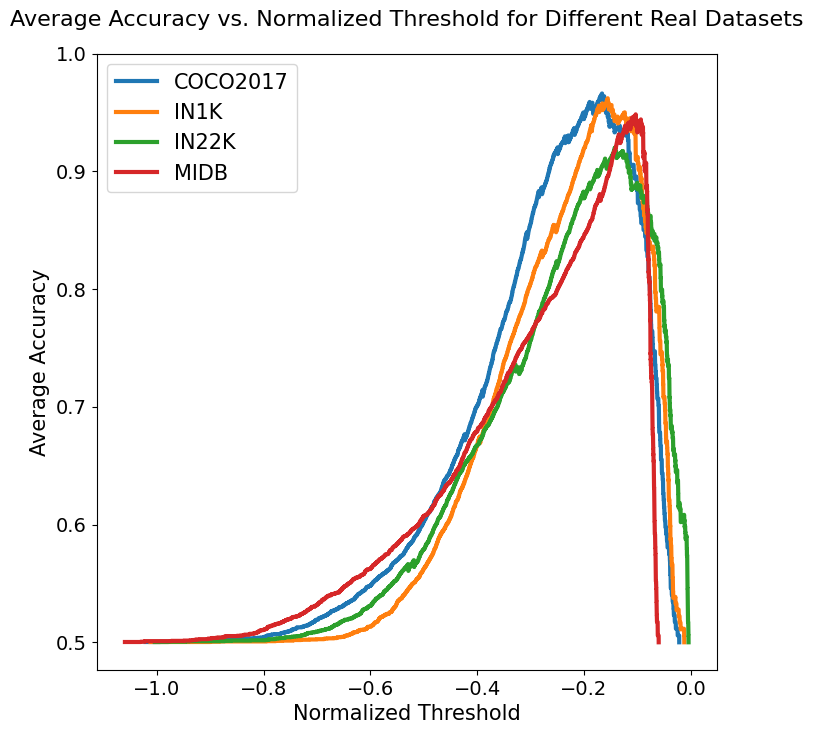}
	\pulluppp
	\caption{Zero-shot detection performance of our method versus different normalized thresholds.}
	\label{fig:acc_vs_thresholds}
	\pulluppp\pullup
\end{figure}

%% file: sections/suppl/ZeroShotPerfVsThresholds_v0.tex
\section{Zero-Shot Performance vs. Thresholds}
\label{supp:zeroshot_vs_thresholds}

In this section, we study the detection performance's impact as a result of varying the decision threshold. To do this, we vary a normalized threshold and measure the average accuracy over all real-vs-synthetic dataset pairs with respect to a real dataset. We note that the accuracy is balanced because the number of real and synthetic samples in each pair is identical. the results of this experiment is provided in Fig.~\ref{fig:acc_vs_thresholds}.

The results in Fig.~\ref{fig:acc_vs_thresholds} show that the average accuracy generally increases as the normalized threshold approaches an optimal range, peaking at a certain value before declining. This behavior is consistent across all datasets, though the precise peak accuracy and the threshold at which it occurs vary slightly between datasets. However, all peaks generally occur within the narrow range of thresholds between -0.10 and -0.14. This narrow range highlights the stability of our method's performance across different real datasets, indicating that forensic self-descriptions offer robust generalization to varying real-vs-synthetic scenarios.

This stability has practical implications: a system employing forensic self-descriptions for zero-shot detection may not require extensive threshold calibration for different datasets. Instead, it can rely on a pre-set threshold determined from a small validation set, simplifying deployment while maintaining consistently high performance across diverse datasets.

%% file: sections/suppl/EffectsOfRealDatasets_v0.tex
\section{Impact of Real Training Dataset Choice}
\label{supp:effects_of_diff_reals}

In this section, we examine the impact of the choice of the real dataset used for training to the overall zero-shot detection performance. We do this by evaluating the performance of forensic self-descriptions derived from residuals produced by scene content predictive models trained on one real dataset and tested on entirely different real datasets. The results of this experiment are provided in Fig.~\ref{fig:cross_val_diff_reals}.

The results in Fig.~\ref{fig:cross_val_diff_reals} illustrates the robustness and generalization capability of our proposed method when applied to unseen real datasets. Specifically, we achieve consistently high performance across all scenarios, with average AUC values typically remain around 0.94, regardless of the real dataset used for training or testing. This result highlights the fact that our method can maintain its strong performance even when the specific characteristics of real data available during training may differ from those encountered in the wild.

Notably, on MIDB where we observe a slight gap in performance when other datasets are used for training. This effect can be qualitatively explained by examining Fig.~\ref{fig:spec_of_reals} in Sec.~\ref{supp:spec_of_diff_reals}, where we observe that the self-descriptions obtained from real images in MIDB are significantly different from those in other datasets. This is because in constrast to other datasets
where images are often downloaded from the internet,
images in MIDB come directly from a camera without any subsequent post processing or compression.
Therefore, for practical applications, this finding shows that better performance may be achievable by training the scene content predictive models on a larger, combined set of real images from diverse sources.

%% file: sections/suppl/SpectralOfDiffReals_v0.tex
\section{Qualitative Study of Forensic Self-Descriptions of Different Real Datasets}
\label{supp:spec_of_diff_reals}

In this section, we explore the characteristics of the forensic self-descriptions of real images from different sources. In particular, we examine the power spectrum of different filters in the forensic self-descriptions across real image datasets (COCO2017, IN-1k, IN-22k, and MIDB). We show these visualizations in Fig.~\ref{fig:spec_of_reals}.

From Fig.~\ref{fig:spec_of_reals},
we can observe that the power spectra of the filters exhibit consistent patterns across the different datasets. For instance, similar spectral structures are observed in $FFT(\phi_2)$ and $FFT(\phi_3)$ of COCO2017, IN-1k, and IN-22k. While the spectral structures of other filters are slightly different across these three datasets, we observe that they are still significantly distinct from those produced by synthetic images (see Fig.~\ref{fig:self_desc_viz} in our main paper). This shows that our method of using forensic self-descriptions can accurately distinguish AI-generated images from real images. This is also supported by our experimental results in Sec.~\ref{sec:zeroshot_eval} of our main paper, where our average zero-shot detection performance is 0.960 with a standard deviation of only 0.01. In contrast, other methods have significantly more deviations between different real sources. For instance, NPR suffers big performance drops in IN-1k and IN-22k, ZED in IN-1k, and Aeroblade in IN-22k.

Notably, we see a much bigger difference in the spectral patterns of the self-descriptions of images in the MIDB dataset. This is because real images in this dataset come directly from a camera without subsequent post processing or compression. The fact that our forensic self-descriptions can capture these differences show that our method is highly generalizable and adaptable to many real-world image processing conditions.

%% file: sections/suppl/SpaceTimeAnalysis_v0.tex
\section{Space-Time Complexity Analysis}
\label{supp:space_time_analysis}

\input{tables/SpaceTimeComplexity}

In this section, we examine the runtime and memory cost in terms of the number of parameters of ours and competing methods. We record the average inference runtime per image by performing inference for each method using 1000 images from the ImageNet-1k dataset using a machine with an NVIDIA A6000 GPU.

The runtime and parameter comparison in Table~\ref{tab:space_time_comp} highlights a significant trade-off in our method. Our approach has the lowest number of parameters (2K), making it highly efficient in terms of model size and memory requirements. However, it takes the longest time per image (0.11 image/s), primarily due to the iterative residual modeling process, which requires optimization for each image to accurately capture forensic microstructures. In contrast, other methods such as Fang et al. achieve much faster runtimes (289.54 image/s) by leveraging pre-trained models or architectures optimized for inference speed, albeit at the cost of significantly larger parameter sizes. These results underscore that while our method is highly compact and lightweight, the computational complexity of its residual modeling process remains a bottleneck. In future work, we will address this issue by exploring faster optimization techniques or approximations to further enhance the practicality of our approach without sacrificing its accuracy and generalization capabilities.

%% file: tables/SpaceTimeComplexity.tex
\begin{table}[!t]
	\centering
	\caption{\label{tab:space_time_comp} Runtime as Images per second (im/s) and Number of Parameters for our method and competing methods in this paper.}
	\resizebox{0.8\linewidth}{!}{%
		\begin{tblr}{
				width = \linewidth,
				colspec = {m{25mm}m{20mm}m{20mm}},
				cell{1-20}{2-3} = {c},
				vlines,
				hline{1-3,11,16,18,21} = {-}{},
			}
			\textbf{Method} & \textbf{Time (im/s)} & \textbf{\# Params} \\
			\textbf{Ours}   & 0.11                 & \textbf{2K}        \\
			CnnDet          & 22.72                & 23M                \\
			PatchFor        & 22.93                & 191K               \\
			LGrad           & 19.53                & 46M                \\
			UFD             & 11.13                & 427M               \\
			DE-FAKE         & 4.90                 & 620M               \\
			Aeroblade       & 5.66                 & 14M                \\
			ZED             & 0.88                 & 809M               \\
			NPR             & 22.92                & 1.4M               \\
			DCTCNN          & 192.67               & 170K               \\
			RepMix          & 186.85               & 24M                \\
			Fang et al.     & \textbf{289.54}      & 1.2M               \\
			POSE            & 24.53                & 22M                \\
			Abady et al.    & 17.02                & 150M               \\
			FSM             & 24.06                & 437K               \\
			ExifNet         & 19.56                & 76M                \\
			CLIP-ViT-Base   & 159.31               & 151M               \\
			CLIP-ViT-Large	& 25.84                & 427M               \\
			ResNet-50       & 20.74                & 23M
		\end{tblr}
	}
\end{table}